\documentclass[lettersize,journal]{IEEEtran}

\usepackage[left=54pt,top=50pt,right=54pt,bottom=57pt]{geometry}

\usepackage{times}
\usepackage{tabularx}
\usepackage{subcaption}
\usepackage{multirow,multicol, array}
\usepackage[font={small}]{caption}   
\usepackage{graphicx,dblfloatfix}
\usepackage{wrapfig}
\usepackage{siunitx}

\let\labelindent\relax
\usepackage{enumitem}

\usepackage{amsmath,amsthm,amssymb,amsfonts}
\usepackage[titlenumbered,ruled]{algorithm2e}
\usepackage{textcomp}
\usepackage{acronym}
\usepackage{balance}
\usepackage{mdwmath}
\usepackage{bm}
\usepackage{bbm}
\usepackage{mdwtab}
\usepackage{array}
\usepackage{eqparbox}
\usepackage{cite}

\usepackage{color}
\usepackage{psfrag}
\usepackage{epsfig}
\usepackage{url}

\usepackage{epstopdf}
\usepackage{booktabs}
\usepackage{blindtext}
\usepackage[colorlinks,bookmarksopen,bookmarksnumbered,linkcolor=blue,citecolor=blue,urlcolor=blue]{hyperref}

\usepackage[colorinlistoftodos,prependcaption,textsize=tiny]{todonotes}

\usepackage[a-1b]{pdfx}

\renewcommand{\emph}{\textit}

\newtheorem*{lemma*}{Lemma}

\newtheorem*{problem*}{Problem}

\newcommand{\sgn}{\text{sign}}

\allowdisplaybreaks[4]

\makeatletter
\newcommand\fs@spaceruled{\def\@fs@cfont{\bfseries}\let\@fs@capt\floatc@ruled
    \def\@fs@pre{\vspace{5\baselineskip}\hrule height.8pt depth0pt \kern2pt}%
    \def\@fs@post{\kern2pt\hrule\relax}%
    \def\@fs@mid{\kern2pt\hrule\kern2pt}%
    \let\@fs@iftopcapt\iftrue}
\makeatother

\IEEEoverridecommandlockouts
\graphicspath{{images/}}

\begin{document}
	\acrodef{ISR}[\textsc{isr}]{Intelligence, Surveillance, and Reconnaissance}
	
	\title{Online Search-based Collision-inclusive\\ Motion Planning and Control for\\ Impact-resilient Mobile Robots}

	\author{Zhouyu Lu, Zhichao Liu, Merrick Campbell, and Konstantinos Karydis
		\thanks{The authors are with the Dept. of Electrical and Computer Engineering, University of California, Riverside. 
			Email: {\{zlu044, zliu017, mcamp077, karydis\}@ucr.edu}. 
		We gratefully acknowledge the support of NSF 
		\#IIS-1910087, ONR 
		\#N00014-18-1-2252 and \#N00014-19-1-2264, and ARL 
		\#W911NF-18-1-0266.  
		Any opinions, findings, and conclusions or recommendations expressed in this material are those of the authors and do not necessarily reflect the views of the funding agencies.}
	}
	
	\maketitle
	
\begin{abstract}
This paper focuses on the emerging paradigm shift of collision-inclusive motion planning and control for impact-resilient mobile robots, and develops a unified hierarchical framework for navigation in unknown and partially-observable cluttered spaces. At the lower-level, we develop a deformation recovery control and trajectory replanning strategy that handles collisions that may occur at run-time, locally.  The low-level system actively detects collisions (via embedded Hall effect sensors on a mobile robot built in-house), enables the robot to recover from them, and locally adjusts the post-impact trajectory. 
Then, at the higher-level, we propose a search-based planning algorithm to determine how to best utilize potential collisions to improve certain metrics, such as control energy and computational time. 
Our method builds upon A* with jump points. We generate a novel heuristic function, and a collision checking and adjustment technique, thus making the A* algorithm converge faster to reach the goal by exploiting and utilizing possible collisions. 
The overall hierarchical framework generated by combining the global A* algorithm and the local deformation recovery and replanning strategy, as well as individual components of this framework, are tested extensively both in simulation and experimentally. An ablation study draws links to related state-of-the-art search-based collision-avoidance planners (for the overall framework), as well as search-based collision-avoidance and sampling-based collision-inclusive global planners (for the higher level). 
Results demonstrate our method's efficacy for collision-inclusive motion planning and control in unknown environments with isolated obstacles for a class of impact-resilient robots operating in 2D. 
\end{abstract}
\vspace{-12pt}

\section{Introduction}
There has been an emerging paradigm shift in mobile robot motion planning and autonomous navigation whereby collisions with obstacles are not by default avoided but instead exploited to improve certain robot planning, control and navigation metrics~\cite{lu2019optimal,lu2020motion,mote2020collision,zha2021exploiting,lew2019contact}. Such \emph{collision-inclusive} planning and control strategies capitalize on results demonstrating how some forms of collisions can in fact be useful in terms of sensing, localization, control, and agility~\cite{schmickl2009get,karydis2014planning,haldane2016robotic,mulgaonkar2017robust,mayya2018localization,StagerICRA19,khedekar2019contact,mulgaonkar2020tiercel,liu2021toward}. 
Besides the benefits of embracing collisions, robot deployment in realistic (that is, dynamic, cluttered, and irregularly-shaped) environments may, at cases, make collision avoidance hard to achieve~\cite{hoy2015algorithms, campbell2012review}. For example, detecting all obstacles in the environment can be a challenge, especially when there exist translucent and/or transparent obstacles, such as glass walls, or reflective surfaces~\cite{mulgaonkar2020tiercel}. At the same time, using a conservative local planner may fail finding a feasible path to the goal even if one exists~\cite{oleynikova2018safe}. Collision-inclusive motion planners can help address the aforementioned challenges. 

Although research on collision-inclusive motion planning has already begun receiving attention, existing methods can be limited in their ways to apply in practical cases. On one hand, methods that evaluate the effect of collision within motion planning~\cite{mote2020collision, zha2021exploiting} do not apply to online problems. On the other hand, existing online collision-inclusive planning methods~\cite{mulgaonkar2017robust, StagerICRA19, de2020collision} cannot decide how to use collisions optimally, which could help guide the robot to the goal. Our previous online planning method~\cite{lu2020motion} can evaluate possible collisions in unknown space which lies outside the field-of-view (FoV) of the robot, but does not consider how to employ collisions optimally within the known (and/or visible) space. 

\begin{figure*}[!t]
	\vspace{0pt}
	\centering
	\begin{subfigure}[b]{0.30\textwidth}
		\includegraphics[trim={5cm 0.7cm 5cm 0.7cm},clip,width=0.9\linewidth]{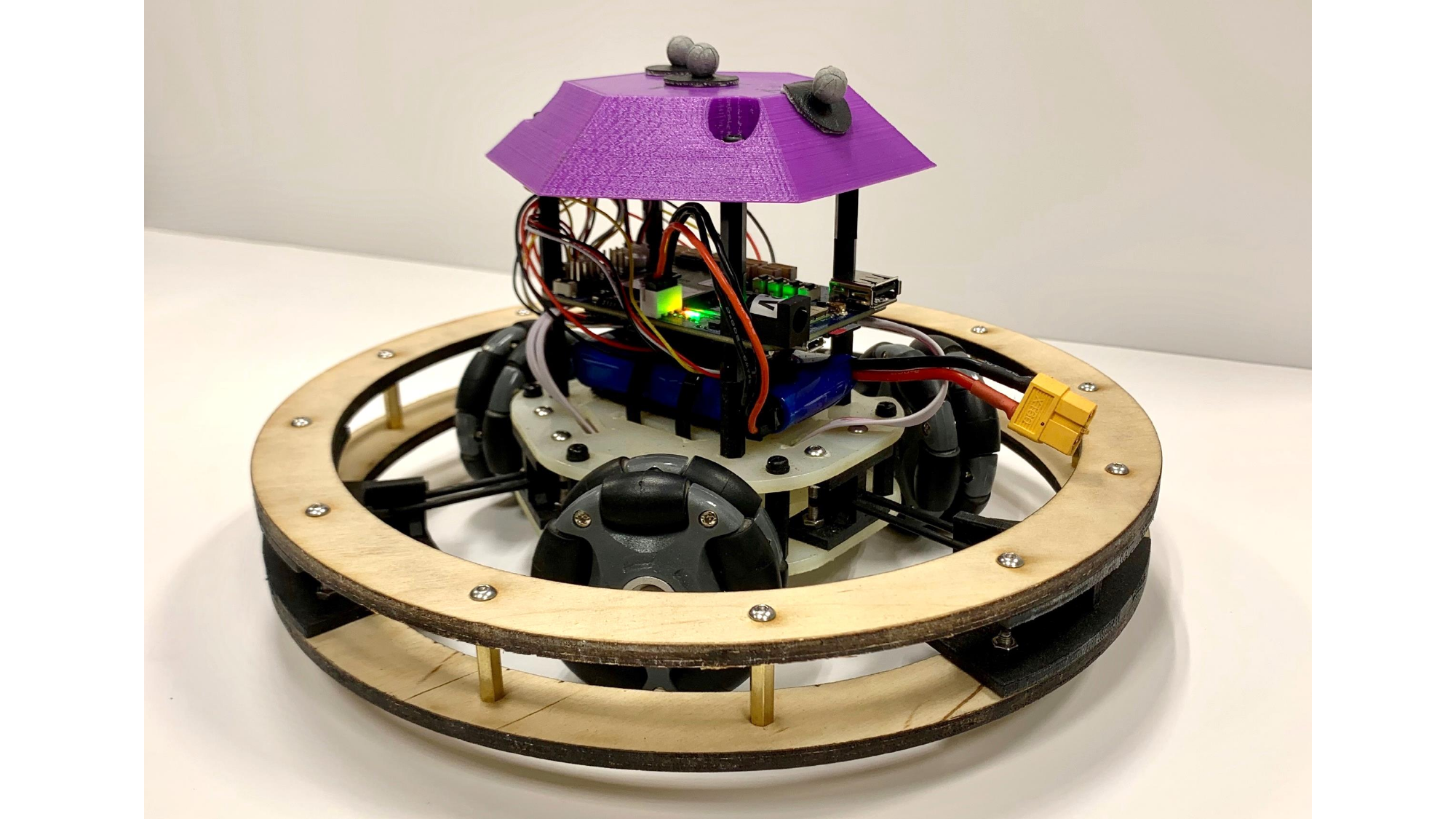}
		\vspace{-3pt}
		\caption{Passive impact-resilient robot.}
		\label{fig:omnipuck}
	\end{subfigure}
	\hspace{9pt}
	\begin{subfigure}[b]{0.30\textwidth}
		\includegraphics[trim={5cm 0.7cm 5cm 0.7cm},clip,width=0.9\linewidth]{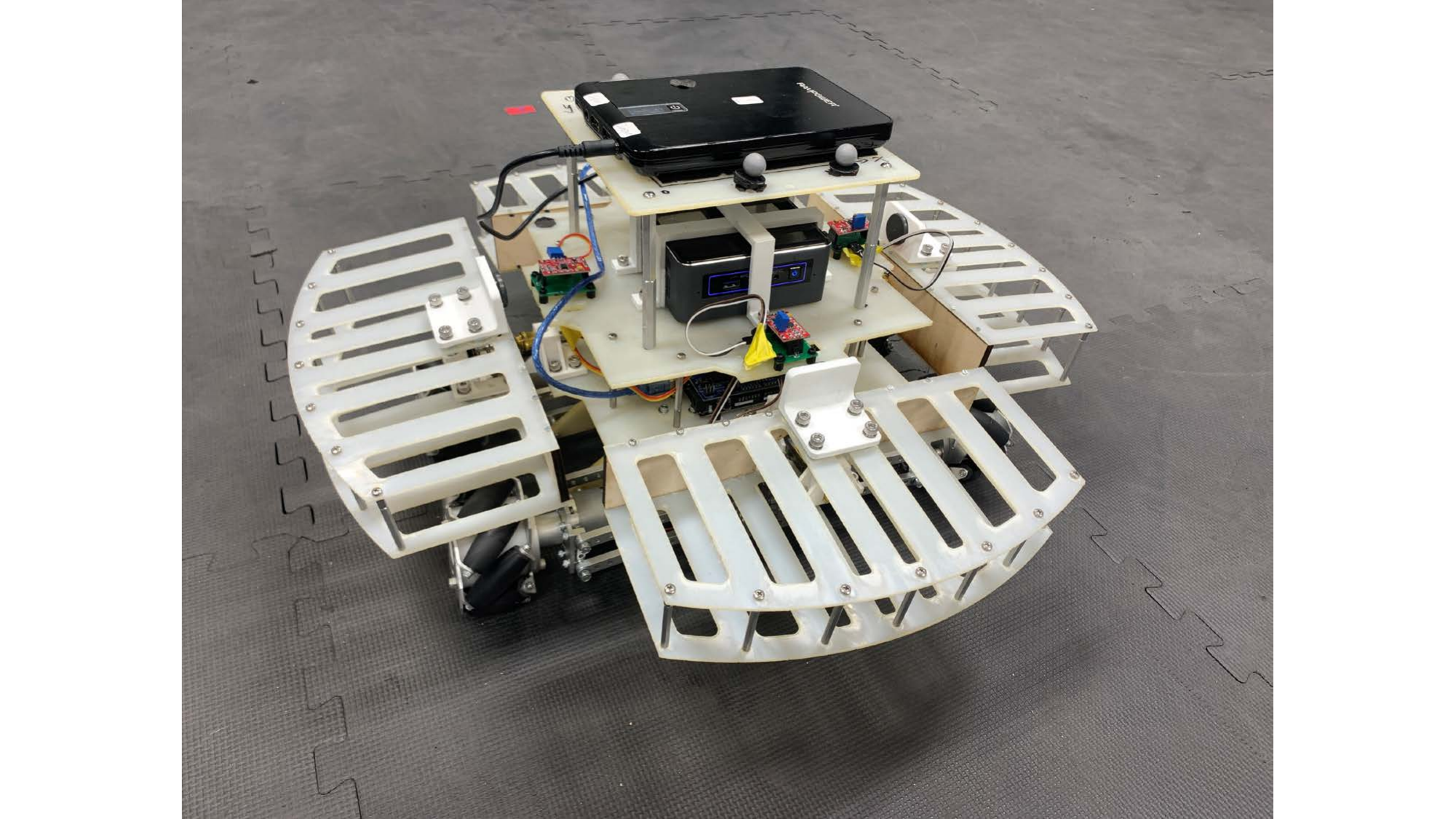}
		\vspace{-3pt}
		\caption{Active impact-resilient robot with $4$ arms.}
		\label{fig:4 arms}
	\end{subfigure}
	\hspace{9pt}
	\begin{subfigure}[b]{0.30\textwidth}
		\includegraphics[trim={5cm 0.7cm 5cm 0.7cm},clip,width=0.9\linewidth]{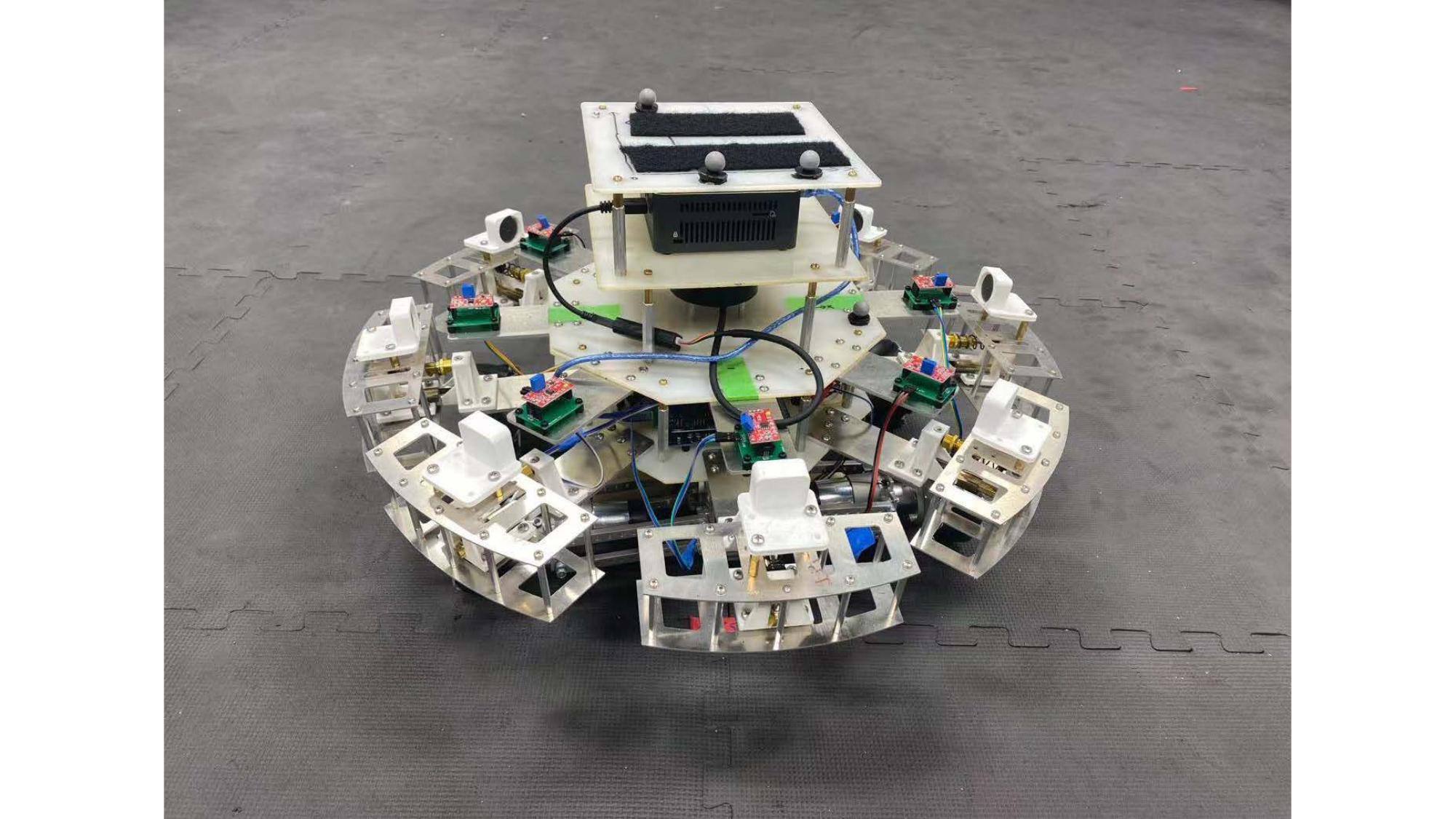}
		\vspace{-3pt}
		\caption{Active impact-resilient robot with $8$ arms.}
		\label{fig:8 arms}
	\end{subfigure}
	\vspace{-2pt}
	\caption{The evolution of our omni-directional holonomic robot prototypes used in our collision-inclusive motion planning and control research program. (a) The first iteration of the robot, inspired by the omnipuck robot~\cite{stager2016stochastic}, featured a passive collision ring and a single-board computer for motion control~\cite{lu2019optimal,lu2020motion}. This iteration has a radius of $0.12$\;m and weighs $0.6$\;kg. (b) The second iteration critically included an active collision ring-like structure that can sense collisions via embedded Hall effect sensors and a powerful onboard computer for online sensor data processing, decision making and motion control~\cite{lu2021deformation}. This iteration has a radius of $0.3$\;m and weighs $6$\;kg. A limitation of that prototype was the sparsity of its arms which reduced collision detection accuracy when the collision surface was not approximately perpendicular to any of the robot's arms. (c) The most recent prototype developed in this present paper builds upon successful features of the second version and has significantly improved collision detection accuracy due to a complete redesign of the arms' mechanical design and integration of more arms. This iteration has a radius of $0.3$\;m and weighs $8$\;kg.} 
	\vspace{-18pt}
	\label{fig:robots}
\end{figure*}

In this paper, we propose a unified online collision-inclusive motion planning and control framework that evaluates the effect of possible collisions and decides when it might be preferred to collide with an obstacle (or a surface more broadly) instead of avoiding it. Our framework applies to impact-resilient robots with three core capabilities: 1) collision resilience, 2) collision identification, and 3) post-impact characterization. We design and fabricate in-house a custom omni-directional holonomic wheeled robot equipped with a collision ring that integrates Hall effect sensors along the arms holding the ring in place (Fig.~\ref{fig:robots}); our robot satisfies all three core capabilities. The robot runs a \underline{D}eformation \underline{R}ecovery control and trajectory \underline{R}eplanning (DRR) strategy~\cite{lu2021deformation} that enables it to recover from a collision and rapidly replan its post-impact trajectory using the information provided by the Hall effect sensors. The DRR strategy acts as the local replanner of the unified framework developed herein. We also propose and develop a global search-based planning algorithm based on the collision model generated from the DRR strategy. Similar to~\cite{liu2017search}, our approach explores the space of trajectories using a set of short-duration motion primitives generated by solving an optimal control problem. Instead of pruning those primitives colliding with the obstacles in the global map, our proposed approach can adjust and evaluate them based on the collision model. 

Succinctly, the paper's contributions are as follows:
	\begin{itemize}
		\item We extend the DRR strategy to generate local trajectories when colliding with (non-)convex obstacles.
		\item We develop a search-based planner to generate global trajectories and evaluate it in different benchmarks.
		\item We propose and evaluate a unified online collision-inclusive motion planning and control framework integrating the DRR strategy and search-based collision-inclusive planning while considering the robot's FoV.
	\end{itemize}

Our method is systematically evaluated via both simulated and real-world experiments using planar holonomic wheeled robot kinematics in environments that contain isolated convex and non-convex obstacles. We first test the DRR strategy experimentally to ensure its feasibility and safety when applied to the physical robot. Data collected from this process help identify parameters for the collision model which is necessary to test the search-based collision-inclusive algorithm in simulation. Comprehensive benchmark comparisons against state-of-art collision-avoidance and collision-inclusive methods demonstrate the differences, similarities and the utility of specific components, as well as of the overall proposed framework. Moreover, experimentation with the physical robot in a single corridor environment is conducted to validate the performance of our unified online collision-inclusive motion planning and control framework. 

This paper builds upon and significantly extends previous results~\cite{lu2020motion,lu2021deformation}. 
The former~\cite{lu2020motion} focuses only at the global planning level and evaluates possible collisions in the unknown (not yet observed) space. 
The latter~\cite{lu2021deformation} focuses only at the local control and planning level that utilizes DRR based on a-priori given waypoints. 
This paper, in contrast, develops the unified framework that combines the global planning and local control and planning levels together. In this newly-developed approach, the global planner can evaluate possible collisions both within and outside the robot's FoV.
\footnote{~To make this present paper self-contained, important methods and results from the previous related papers~\cite{lu2020motion,lu2021deformation} are included herein.}

In what follows, we review related works in Sec.~\ref{sec:related} and introduce our overall system's structure in Sec.~\ref{sec:overview}. The deformation recovery control and post-impact trajectory replanning components are detailed in Sec.~\ref{sec:deformation} and Sec.~\ref{sec:replanner}, respectively. The global search-based planner is discussed in Sec.~\ref{sec:search-based}. Extensive benchmark (in simulation) and experimental results are given in Sec.~\ref{sec:experimental result}. Sec.~\ref{sec:conclusion} discusses key findings and current limitations, and elaborates on future directions of research enabled by the proposed framework.

\section{Related Works}\label{sec:related}
Collision-free motion planning algorithms handle obstacle avoidance in distinct ways (e.g.,~\cite{liu2017search, liu2017planning, zhou2021raptor,tordesillas2021faster,deits2015efficient}) to derive collision-free trajectories in real-time. Typically, such methods split the trajectory generation problem into two parts: 1) planning a collision-free geometric path or using motion primitives, and 2) optimizing the path locally to obtain a dynamically-feasible time-parameterized trajectory. When the environment is unknown (or partially-known), different strategies have been used based on those two-part framework. Many methods adopt the optimistic assumption~\cite{gao2018online, tordesillas2019real}, which treats the unknown space as collision-free. This strategy improves the speed of reaching goals but may not guarantee safety. In contrast, other methods treat the unknown space as obstacle-occupied~\cite{liu2016episodic} and only allow for motions within the already known free space or FoV-observed free space~\cite{lopez2017aggressive}. Although these restrictions can help ensure safety, they tend to lead to conservative motion. 

Tordesillas et al.~\cite{tordesillas2021faster} proposed a method that combines these two strategies by planning in both the known-free and unknown spaces. Instead of being overly optimistic about the unknown space, backup trajectories are also planned to enforce safety should the assumption about unknown space being free turns out to be wrong. While this method works well overall, it has put less emphasis on environment perception, lack of which may reduce safety or create over-conservative trajectories when the robot is tasked to operate at high speeds~\cite{mohta2018fast}. 
To this end, perception-aware strategies~\cite{richter2016learning, zhou2021raptor, heiden2017planning} have been proposed to predict unknown dangers and try to discover and avoid those dangers early on. However, prediction of unknown dangers does not necessarily ensure accuracy and usually requires additional computational effort which may limit online implementation. 

Different from collision-avoidance, there have been efforts on designing impact-resilient robots that can withstand collisions instead (e.g.,~\cite{briod2014collision,haldane2015integrated,stager2016stochastic,li2019agile,StagerICRA19,de2020collision,mulgaonkar2020tiercel, zha2021exploiting,lu2020motion,liu2021toward,lu2021deformation}). With such robots as hand, one research direction has been to design characterization methods that can make the robot sense the collision and recover from the collision state. Most of such characterization methods have mainly focused on utilizing data from an onboard inertial measurement unit (IMU)~\cite{battiston2019attitude}. However, IMUs are usually unable to distinguish collisions during aggressive maneuvers and to detect static contacts, resulting in low accuracy in collision detection. Sensors that could detect deformation of the robot during the collision process have been used in the past to provide more accurate collision detection~\cite{briod2013contact,de2020collision}. In related yet distinct previous work~\cite{liu2021toward}, we have implemented a passive quadrotor arm design with Hall effect sensors, making the robot able to detect and characterize collisions. The ability to sense and characterize collisions has led to various different methods to replan the local trajectory once the collision is detected~\cite{mulgaonkar2017robust, de2020collision, StagerICRA19}.
Planning methods using motion patterns, e.g., to move forward in straight lines until collision with environment boundaries and then rotate in place and move forward again, have also been proposed~\cite{nilles2021visibility, alam2017minimalist, lewis2013planning}. Such methods can run online in environments with non-convex, polygon-shaped obstacles. A different trajectory generation method can be achieved by assuming the robot maintains contact with the obstacle~\cite{khedekar2019contact}. Although these methods increase robustness and safety of post-impact trajectories, they cannot determine where the robot should collide with the environment to help it redirect toward the globally-planned goal. 

Related works~\cite{mote2020collision, zha2021exploiting} propose methods to evaluate and design possible collision spots of the global trajectory. Mote et al.~\cite{mote2020collision} have introduced an empirical algebraic collision model by directly relating pre- and post-impact velocities with no thrust commanded. Then, a mixed-integer planning method based on that model is used to compute collision-inclusive trajectories in a known environment. However, integer constraints are hard to create, and solving a mixed-integer programming problem is usually time-consuming, making it impossible to run online for planning in an unknown environments. Further, the approach~\cite{mote2020collision} has been demonstrated with a specific pair of objects over a relatively limited range of conditions (obstacles need to be line segments). Zha and Mueller~\cite{zha2021exploiting} have proposed a rapidly exploring random tree (RRT) based planning method to plan global trajectories with collisions. The impulsive model is used to create the post-collision state once the pre-collision state is generated. 
Findings from~\cite{zha2021exploiting} suggest that a collision-inclusive sampling-based planner is likely to find better trajectories in cluttered environments (such as narrow tunnels) as compared to environments that contain isolated obstacles. In addition, such algorithms remain limited in their use for online planning in unknown (or partially-known) environments.

Compared to our previous work~\cite{lu2020motion}, which developed a global path planner that explicitly trades-off between risk and collision exploitation only in unknown space, this paper proposes a new search-based global planner using a set of short-duration motion primitives which exploit possible collisions in the environment. The planner treats the unknown space as collision-free. Our planner can generate waypoints with explicit information about possible collisions and more reasonable time allocation for the local trajectory generator. 
Contrary to collision avoidance methods with hard constraints that generate trajectories only in conservative local space~\cite{liu2017planning, tordesillas2021faster}, our method utilizes gradient-based trajectory optimization (GTO)~\cite{zhou2021raptor}, which typically formulates trajectory generation as a nonlinear optimization problem and incorporates the artificial potential field (APF) to ensure safety. However, since GTO does not guarantee the robot will avoid all possible collisions, especially in unknown environments, we utilize the DRR strategy~\cite{lu2021deformation} for local trajectory generation; once a collision is sensed and characterized, DRR can ensure that the robot will recover from the collision and keep progressing toward its global goal. 

\vspace{-6pt}
\section{System Overview}\label{sec:overview}

\subsection{Overall Framework}
\begin{figure}[!h]
\vspace{-15pt}
\centering
\includegraphics[trim=0.5cm 0.1cm 1.0cm 0.1cm, clip,width=0.95\linewidth]{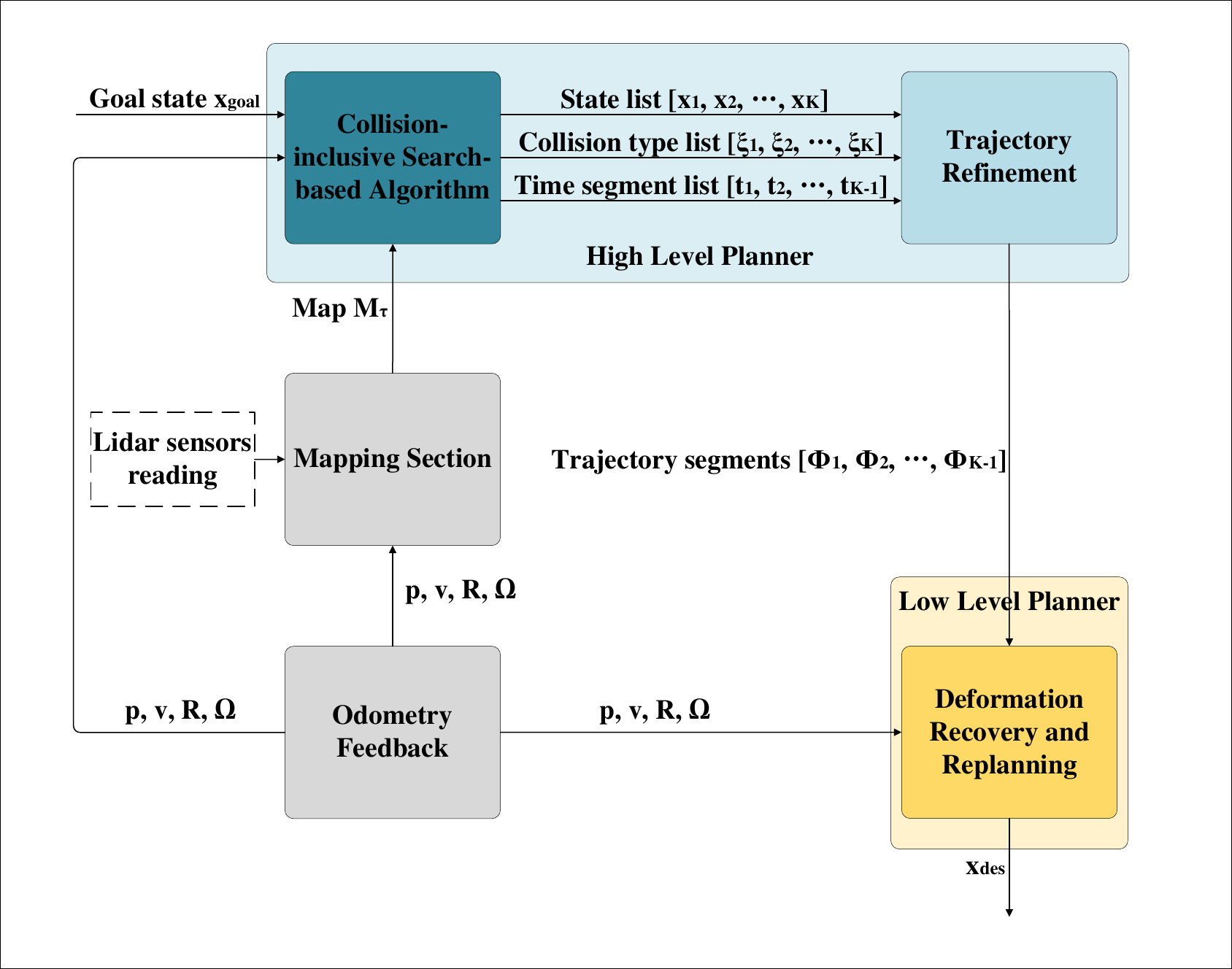}
\vspace{-12pt}
\caption{Overview of our unified framework for collision-inclusive motion planning and control. The method builds upon two novel components developed in this work; a global collision-inclusive planner and a local trajectory generator.}%
\label{fig:Navigation-software-architecture}
\end{figure}

Our overall system architecture is shown in Fig.~\ref{fig:Navigation-software-architecture}. Novel contributions relate to the low-level planner (Sec.~\ref{sec:deformation} and~\ref{sec:replanner}) and the high-level planner (Sec.~\ref{sec:search-based}) in partially-known environments. The robot may collide with obstacles that were not detected at any time instant that the map (e.g., provided via LiDAR scans) refreshes. Instead of stopping when sensing the collision, the robot locally refines the trajectory and continues to explore the unknown space. To avoid repeated collisions with an obstacle (reminiscent of stacking into local minima) if another collision occurs while the robot follows the locally-revised trajectory part, the robot will then stop and invoke the high-level global planner to make more substantial refinements to the trajectory. Both processes run online.

\begin{figure}[!t]
\vspace{0pt}
\centering
\includegraphics[trim=0.5cm 0.1cm 1.0cm 0.1cm, clip,width=0.8\linewidth]{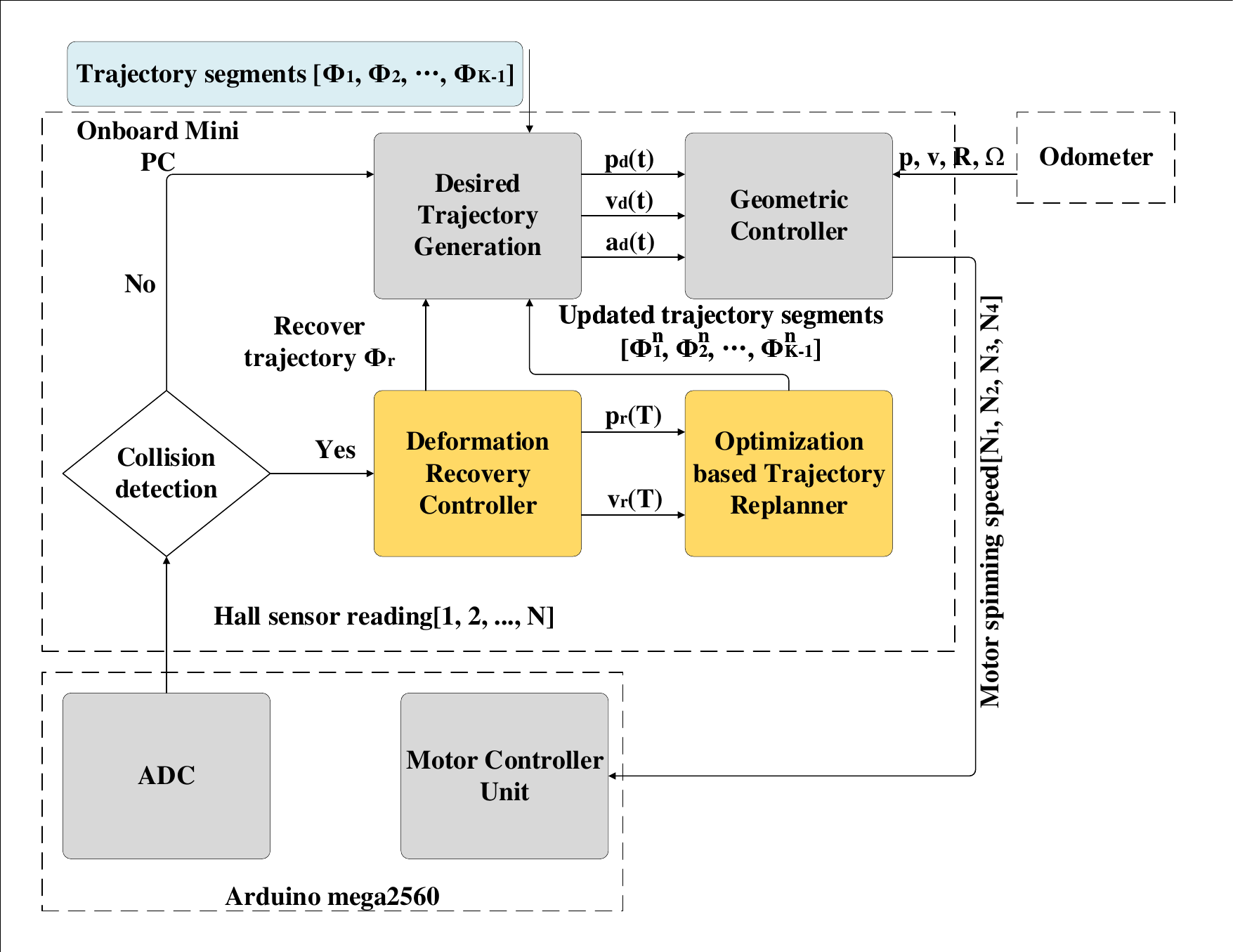}
\vspace{-6pt}
\caption{The DRR strategy for this work's low-level planner. 
}%
\label{fig:Software-architecture}
\vspace{-18pt}
\end{figure}

Contrary to collision avoidance algorithms, we do not impose any obstacle-related constraints in trajectory generation, nor we run a geometric collision check once a trajectory is generated at the low level. Instead, we directly generate a trajectory based on given waypoints.\footnote{~The list of waypoints can be computed via any path planning method. It is independent from our proposed collision-inclusive planning algorithm.} If a collision occurs, the robot receives a signal that a collision has occurred from any of the Hall effect sensors embedded between the main chassis and its deflection surfaces and activates a deformation recovery controller. The controller (Sec.~\ref{sec:deformation}) makes the robot detach from the collision surface by recovering from the deformation, and determines a post-collision state for the robot so as to facilitate post-impact trajectory replanning. The replanner (Sec.~\ref{sec:replanner}) refines the initial trajectory since collisions change the second-order continuity of the trajectory followed before collision. To do so, the replanner uses the post-collision state computed by the recovery controller as initial state for refined trajectory generation.  
The procedure repeats as new collisions occur in the future, in a reactive and online manner. 
Figure~\ref{fig:Software-architecture} shows the DRR strategy, along with specific implementation components for experimentation. 

We select GTO for post-impact recovery and global trajectory refinement based on~\cite{zhou2021raptor} that revealed that GTO-based methods are particularly effective for local replanning, which is key for high-speed online motion planning in unknown environments. One drawback of GTO is the presence of local minima that may lead to undesirable solutions. Specifically, GTO may yield a trajectory that intersects with the obstacles in the environment~\cite{zhou2021raptor}. Our DRR strategy can resolve this issue by offering a way to run a quick replan locally after the collision happens to ensure post-impact consistency.

\vspace{-6pt}
\subsection{Problem Assumptions and Notation}
The proposed approach applies under the following:
\begin{itemize}
    	\item The boundary of the environment is known. 
		\item Operating environments attain the form of confined corridors with isolated convex and non-convex obstacles, and only planar collisions obstacles are considered. 
		\item During deformation and until a collided arm recovers its initial length, the tip of the arm remains in contact with the collision surface but does not rotate about the $z$ axis, and the wheels of the robot contact the ground. 
		\item The Hall effect sensor can return the information of collision state timely. 
\end{itemize}
Key notation used in this paper is shown in Table~\ref{tab:notations overall}.

\begin{table}[h!]
\vspace{-4pt}
\caption{Key notation.}
\label{tab:notations overall}
\vspace{-12pt}
\begin{center}
\begin{tabular}{l l}
\toprule
$\bm{l}_0$&  neutral length vector of the spring\\
$\bm{l}_s$&  pre-tensioned spring length (arm not compressed)\\
$\bm{l}_e$&  length at maximum spring load following Hooke's law\\
$\bm{l}$  &  current spring length (deformation vector)\\
$\bm{p}$  &  position vector of the robot\\
$\bm{v}$  &  velocity vector of the robot\\
$\bm{a}$  &  acceleration vector of the robot\\
$\bm{s}^{-}$  &  state vector prior to the collision and recovery\\
$\bm{s}^{+}$  &  state vector post to the collision and recovery\\
$\bm{s}_{d}$  &  state vector of the robot (point of mass model)\\
\midrule
$k_{e}$& spring constant of the arm.\\
$k_{d}$& damping coefficient of the arm.\\
\midrule
$\tau$ & time interval\\
$t_{c}$ & time instance when the sensing collision\\
$T_{r}$ & time horizon of deformation recovery\\
$T_{rep}$ & time horizon of replanning\\
\midrule
$~^{w}_{b}{R}$ & Rotation matrix from body frame to world frame\vspace{2pt}\\
$~^{w}_{c}{R}$ & Rotation matrix from collision frame $\mathcal{F}_c$ to world frame\\
\midrule
${F}$ & state transition matrix of deformation controller\\
${G}$ & input to state matrix of deformation controller\\
${C}$ & mapping matrix of polynomial coefficient $\bm{\eta}$ \\
${Q}$ & cost matrix of smoothness term\\
${A}_{f}$ & state transition matrix in free space\\
${B}_{f}$ & input to state matrix related to $\bm{s}_{d}$ in free space\\
\midrule
${J}_{s}$ & smoothness term objective function \\
${J}_{o}$ & objective function of the clearance \\
${J}_{v}$ & penalty on velocity \\
${J}_{a}$ & penalty on acceleration \\
\bottomrule
\end{tabular}
\end{center}
\vspace{-14pt}
\end{table}

\section{Deformation Recovery Control} \label{sec:deformation}

The purpose of our proposed deformation controller is to make the robot recover from a collision and reach a post-impact state that can facilitate recovery trajectory replanning (which we discuss in the next section). 
    
\subsection{Problem Setting}\label{sec:deformation problem}
\begin{figure}[!t]
\vspace{6pt}
  \centering\
  \begin{subfigure}{0.235\textwidth}
   \includegraphics[trim={0.6cm 0.3cm 0.6cm 0.3cm, width=0.75\linewidth}, clip, width=0.95\textwidth]{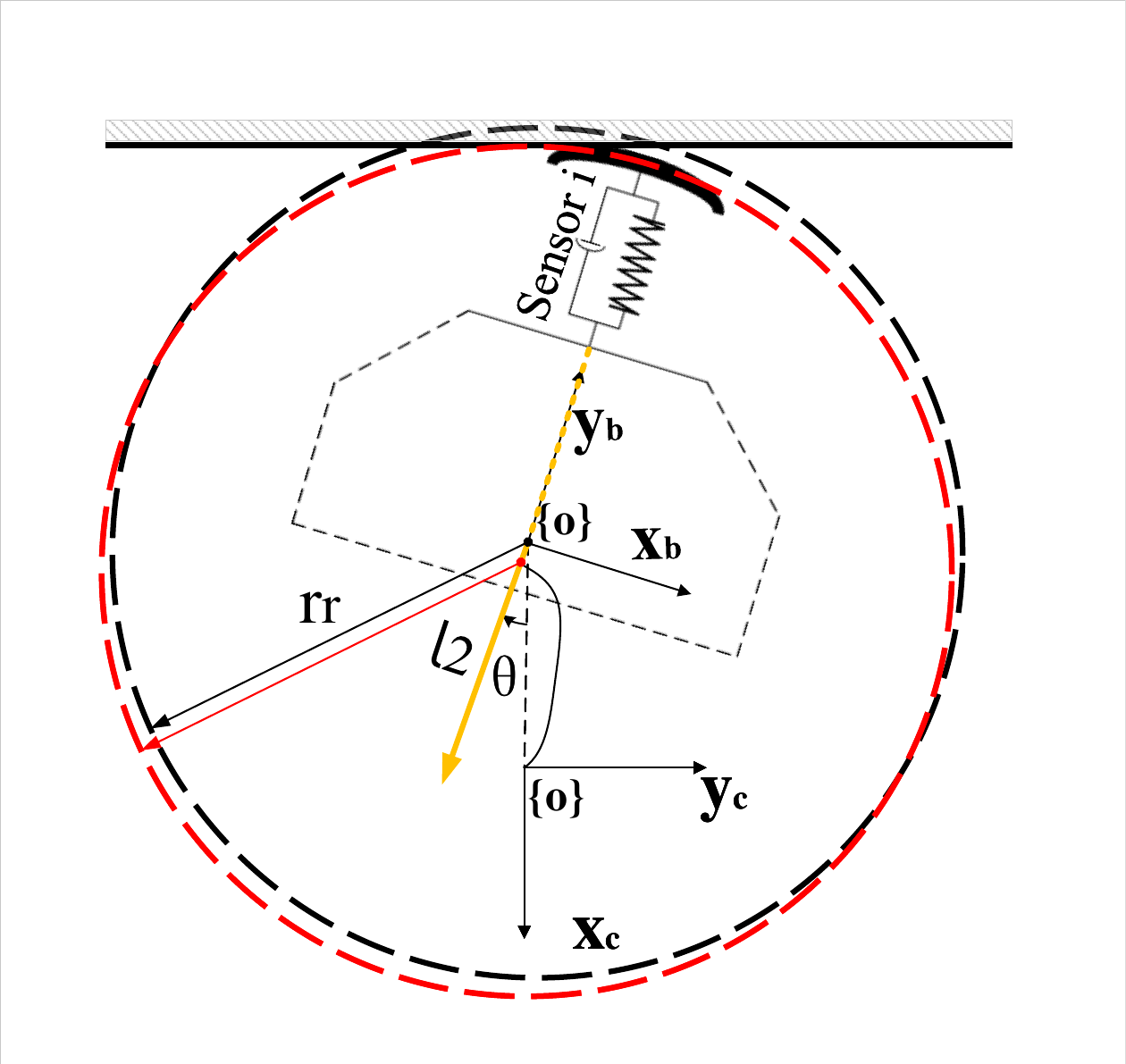}
   \end{subfigure}
    \begin{subfigure}{0.23\textwidth}
        \includegraphics[trim={6cm 0.6cm 7.8cm 0.6cm, width=0.75\linewidth}, clip, width=0.95\textwidth]{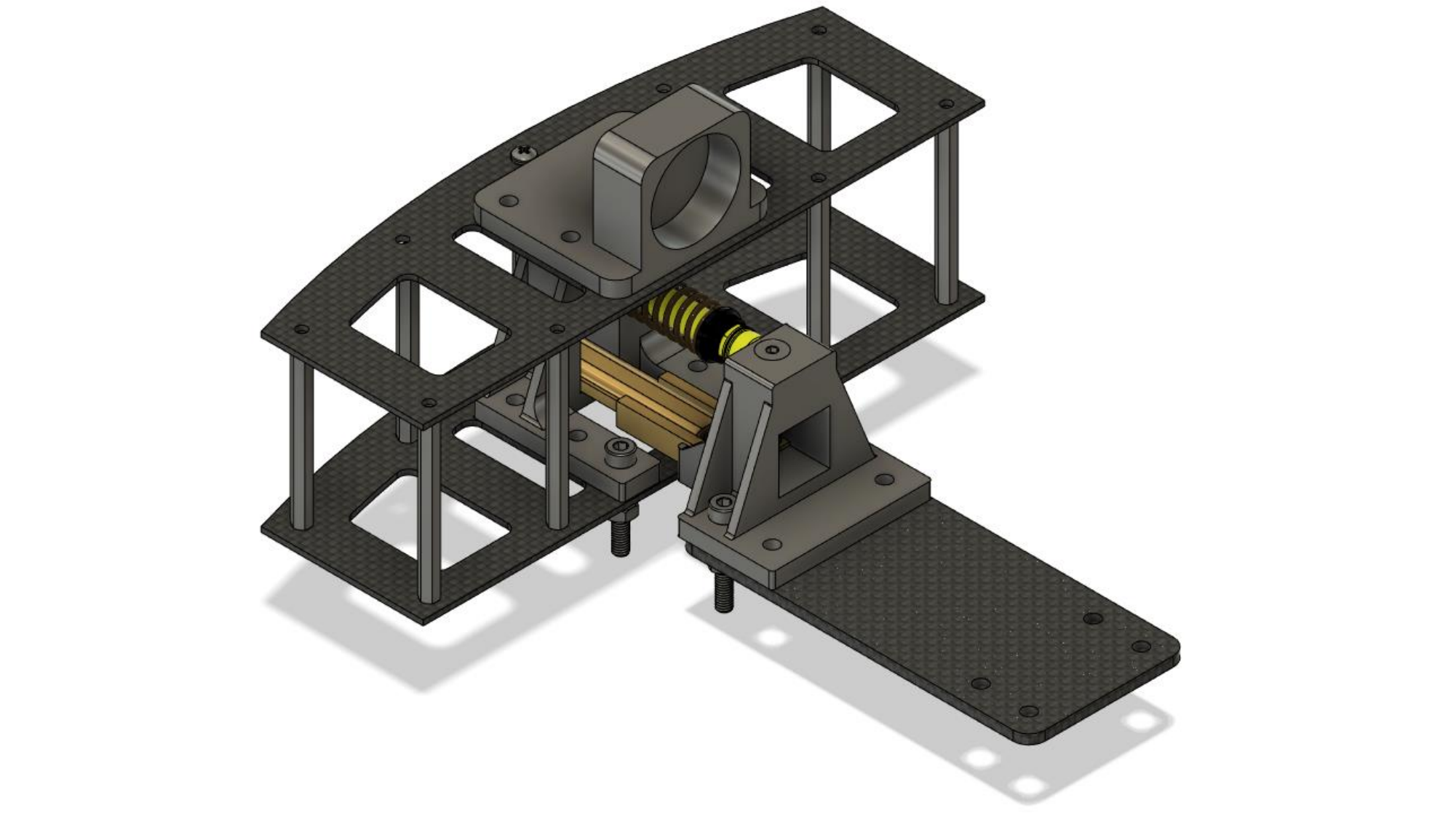}
      \end{subfigure}
      \vspace{-1pt}
\caption{(Left) Model of our wheeled robot equipped with compliant arms. (Right) Close-up view of the assembly of the visco-elastic prismatic joint and Hall effect sensor.}
\label{fig:Voigt-Collision}
\vspace{-9pt}
\end{figure}

Consider a holonomic mobile robot (Fig.~\ref{fig:8 arms}), modeled as a point mass $m$.
The robot's main chassis is connected to its deflection surfaces via visco-elastic prismatic joints (Fig.~\ref{fig:Voigt-Collision}). Note that the springs inside each joint are pre-tensioned.
The robot's compliant arms can both protect the robot from collision damage, and generate an external force driving it away from obstacles. 
External forces along each arm are caused via visco-elastic deformations assumed to follow the Voigt model; $k_{e}$ and $k_{d}$ denote the spring constant and damping coefficient, respectively. 
Hall effect sensors are used to measure the amount of deformation along each arm, and to signal collision detection when a user-tuned arm compression threshold is exceeded.\footnote{~The threshold is tuned based on the sensitivity of the Hall effect sensors.}
The arm design with the bump sensor mechanism is similar to the button mechanism~\cite{sandin2003robot} and helps protect the robot from damage caused by collision as well as sense the collision in real time. Collision detection accuracy is related to the number of arms on the robot.

We consider four key quantities related to spring lengths: neutral $\bm{l}_{0}$, pre-tensioned $\bm{l}_{s}$, maximum-load $\bm{l}_{e}$, and current $\bm{l}$ (also referred to as deformation vector). These quantities play a significant role in the deformation recovery controller; they are also summarized in Table~\ref{tab:notations overall}, along with other key notation. 
In single-arm collisions, current spring length vector $\bm{l}$ is aligned with the unit vector along the colliding arm, pointing from the tip of the arm to the center of robot along the compliant prismatic joint. For clarity of presentation, we consider in the following single-arm collisions. In multi-arm collisions we compute individual contributions from each colliding arm's spring and then consider their vector sum as the compound deformation vector used in lieu of $\bm{l}$.

We use three coordinate systems. The world and body frames ($^{w}_{b}{R}$ denotes the rotation matrix from body to world frames while $^b \bm{l}$ denotes the deformation vector expressed in the body frame), and a (local) collision frame $\mathcal{F}_{c}$. This frame is defined at the time instant a collision occurs, $t_c$, and remains fixed for throughout the collision recovery process, $T_{r}$. Its origin coincides with the origin of the robot when a collision is detected. Basis vector $\{\bm{n}, \bm{t}, \bm{k}\}$ of $\mathcal{F}_{c}$ are defined normal, tangent and upwards with respect to the deformation vector $\bm{l}$. Let $\theta$ be the angle of deformation vector $\bm{l}$ in $\mathcal{F}_{c}$.\footnote{~Note that inability to define the deformation vector may make the collision frame ill-defined. There are three special cases for this to happen. 
One is when two opposite arms deform exactly equally. In this case, there are two possible solutions to define the direction of the deformation vector along the line connecting the two arms. However, our algorithm still works as it prioritizes motion along the tangent to the collision vector (this would be the case of going through a very narrow straight corridor).
The second case contains asymmetric collisions with three or more arms such that the vector sum is still zero. Then, one can define the collision frame based on the most dominant (in terms of magnitude) individual collision vector. Our algorithm can still work, though it is possible that more collisions will occur as the robot tries to navigate through (this would be the case of going through a very narrow curvy corridor).
The last case is when there is an even (four or greater) number of symmetric collisions of exactly the same magnitude. The collision frame can no longer be defined. However, this case can only happen if the robot is radially pressed (entrapped) so that motion is completely restricted, which is not expected to occur during normal operation.
}

The (frame-agnostic) robot collision dynamics is given by $m \ddot{{\bm{l}}} + k_{d}\dot{{\bm{l}}} +k_{e}({\bm{l}} - {\bm{l}_{0}}) = m{\bm{a}_{in}}$,
where ${\bm{a}_{in}}$ is the robot's body acceleration input as provided by the robot's motors.

\subsection{Deformation Controller}\label{subsec:deformation controller}
The deformation recovery controller's task is to steer the post-impact state of the robot to a desired one within a time period of $[t_{c}, t_{c} + T_{r}]$. The time horizon $T_{r}$ is an important hyper-parameter tuned by the user. Typically, longer $T_{r}$ means the robot will recover from collision with longer time and smoother motion pattern. Through a preliminary calibration phase we selected $T_{r} = 0.5s$.

The deformation controller operates with respect to the local, collision frame $\mathcal{F}_{c}$. Let the state variable be $~^{c}{\bm{s}} = [\hspace{-3pt}~^{c}p_{x} ~^{c}p_{y} ~^{c}\theta ~^{c}v_{x} ~^{c}v_{y}]^{\top}$. 
The control input is $\bm{u} = [u_{x}\ u_{y}\  u_{\theta}]^{\top}$, where
$u_{x} = (^c\bm{a}_{in} - \frac{k_{e}}{m}(^c\bm{l}_{s} - \hspace{-3pt}~^c\bm{l}_{0}))\cdot\hspace{-3pt}~^c\bm{n}$, 
$u_{y} = (^c\bm{a}_{in} - \frac{k_{e}}{m}(^c\bm{l}_{s} - \hspace{-3pt}~^c\bm{l}_{0}))\cdot\hspace{-3pt}~^c\bm{t}$, and 
$u_{\theta} =\hspace{1pt} ^c\hspace{-0pt}\bm{\omega}\cdot\hspace{-3pt}~^c\bm{k}$ with $^c\bm{\omega}$ being the angular velocity of the robot in the collision frame. Note that position control terms include compensation for the force caused by the spring being pre-tensioned when the robot's arm is at its rest length.
Then, the state space model of the robot recovering from collision can be expressed as
\begin{equation}\label{state space model}
    \hspace{-3pt}\left\{\hspace{-6pt}
    \begin{array}{lr} 
        \dot{p}_{x} = v_{x}& \\
        \dot{p}_{y} = v_{y}& \\
        \dot{\theta} = u_{\theta}& \\
        \dot{v}_{x} = -\frac{k_{e}}{m}{p}_{x} - \frac{k_{d}}{m}{v_{x}} + u_{x}& \\
        \dot{v}_{y}=-\frac{k_{e}(\mu \sgn(v_{y}) + \tan \theta){p}_{x} + f_{0}}{m} -\frac{k_{d}(\mu \sgn(v_{y}) + \tan \theta)v_{x}}{m} + u_{y}
    \end{array} \right.
\end{equation}
where $f_{0} = \mu k_{e} \sgn(v_{y}) (^c\bm{l}_{s} - \hspace{-3pt}~^c\bm{l}_{0})\cdot\hspace{-3pt}~^c\bm{n}$. 

Since the robot is holonomic, we can decouple orientation from position control.\footnote{~In our approach we seek to make the robot keep the same orientation it has at the instant it collides throughout the collision recovery process. We follow this approach because it can simplify the overall deformation recovery control problem without sacrificing optimality.}
The orientation and angular velocity errors during recovery time $t\in[t_c,t_c+T_{r}]$ are $\bm{e}_{R}(t) = \frac{1}{2}({R}^{\top}_{d}{R}-{R}^{\top}{R}_{d})^{\vee}$ and $\bm{e}_{\dot{R}}(t) = \bm{\omega} - {R}^{\top}{R}_{d}\bm{\omega}_{d}$, respectively.\footnote{~The vee map $\vee$ is the inverse of a skew-symmetric mapping.} 
Index $d$ denotes desired quantities; these are ${R}_{d} = {R}(t_{c})$ and $\bm{\omega}_{d} = [0\ 0\ 0]^{\top}$. (All terms are with respect to collision frame $\mathcal{F}_c$.)
Then,  
\begin{equation}\label{eq:rotational}
u_{\theta} = -K_{r}e_{R, z}(t)-K_{\omega}e_{\dot{R}, z}(t)\enspace.
\end{equation}
Note that since this is a planar collision problem, the collision recovery orientation controller considers only the $z-$components of orientation and angular velocity errors. 

Regarding collision recovery position control, note that the translation-only motion in \eqref{state space model} is affine. Thus, we can apply feedback linearization. 
The linearized system matrix ${F}$ is
\begin{align*}
 {F} = \begin{bmatrix}
0 & 0 & 1 & 0 \\
0 & 0 & 0 & 1\\
-\frac{k_{e}}{m} & 0 & -\frac{k_{d}}{m} & 0\\
0  & 0 & 0 & 0\\
\end{bmatrix}
\end{align*}
with state vector $\bm{s}_{d} = [p_{x}\ p_{y}\ v_{x}\ v_{y}]^{\top}$. The control input matrix is ${G} = {I_{2\times2}}$ with control input vector $\bm{\nu} = [\nu_{x}\ \nu_{y}]$ given by 
\begin{equation} \label{eq:feedback linearization}
\left\{
    \begin{array}{lr} 
        \nu_{x} = u_{x} & \\
        \nu_{y} = u_{y} - \frac{k_{e}(\mu \sgn(v_{y}) + \tan \theta)x + f_{0}}{m} - \frac{k_{d}(\mu \sgn(v_{y}) + \tan \theta)v_{x}}{m}
\end{array} 
    \right.\hspace{-16pt}
\end{equation}

We formulate an optimal control problem with fixed time horizon $T$ based on the linearized system $\dot{\bm{x}} = {F}\bm{x} + {G}\bm{\nu}$. Using the change of variable $\tau=t-t_c$,\footnote{~We employ this change of variable for clarity. Problem (4) resets every time a new collision occurs; this gives rise to an LTI system, hence the change of variable can apply.} we seek to solve
    \begin{subequations}\label{optimal control}
    \begin{alignat}{2}
    &\!\min_{\bm{s}_{d}}        &\qquad& \int\limits_{0}^{  T_{r}}{(\bm{s}_{d}(\tau)^{\top}{\Gamma}\bm{s}_{d}(\tau)+\bm{\nu}^{\top}(\tau){H}\bm{\nu}(\tau))}d\tau\label{eq:optctrl}\\
    &\text{subject to} &      & \dot{\bm{s}}_{d} = {F}\bm{s}_{d} + {G} \bm{\nu},\label{eq:constraint1}\\
    &                  &      & -\lVert \bm{l}_{e} - \bm{l}_{s} \rVert \cos{\theta}\leq {p}_{x} \leq 0.\label{eq:constraint2}\\
    &                  &      & \bm{s}_{d}(0) = [p_{0, x}\ 0\ v_{0,x}\
    v_{0,y}].\label{eq:constraint3}\\
    &                  &      & \bm{s}_{d}(T_{r}) = [0\ p_{T, y}\ v_{T,x}\
    v_{T,y}].\label{eq:constraint4}
    \end{alignat}
    \end{subequations}

Matrices $\Gamma = \gamma\begin{bmatrix}
    {I_{2\times2}} & {0} \\
    {0} & {0_{2\times2}} \\
\end{bmatrix}$ and $H = h{I_{2\times2}}$ penalize the displacement during the recovery process and the control input, respectively. There is a trade-off between the displacement and the control input of the robot. Tuning parameters $\gamma$ and $h$ balance this trade-off to select the controller with minimal control energy and displacement.

Constraint~\eqref{eq:constraint2} dictates that the robot should be in contact with the collision surface until the colliding arm's spring has recovered its original, pre-tensioned length $\bm{l}_s$ (i.e. the arm is no longer compressed) without compressing beyond its linear region $\bm{l}_e$. 
Constraints~\eqref{eq:constraint3} and~\eqref{eq:constraint4} enforce initial and terminal position and velocity conditions, respectively. 
In detail, $p_{0, x}$ is determined by the colliding arm's Hall effector sensor reading. Since the vector form of the sensor's reading (that is, $~^{b}\bm{l}-~^{b}\bm{l}_s$) is expressed in the body frame, we need transform it to the collision frame $\mathcal{F}_{c}$ as per
\begin{equation}\label{eq: x0}
    p_{0, x} = -[1\;0] 
    ~^{w}_{c}{R}^\top 
    ~^{w}_{b}{R}
    ~(~\hspace{-3pt}^{b}\bm{l} - \hspace{-3pt} ~^{b}\bm{l}_{s})\enspace.
\end{equation}

The velocity components at the collision instant $v_{0,x}$ and $v_{0,y}$ are expressed in frame $\mathcal{F}_{c}$ and are estimated at run-time.\footnote{~In the experiments conducted in this work, velocity measurements are provided via a motion capture camera system, but the method applies as long as velocity estimates are available, e.g., via optical flow.} Post-impact, the arm needs to be uncompressed (hence $p_{T, x}$ is set to $0$), but $p_{T, y}$ is treated as an unconstrained free variable. 
Post-impact terminal velocity components $v_{T,x}$ and $v_{T,y}$ are also expressed in $\mathcal{F}_{c}$ and can be set freely. In Sec.~\ref{sec:replanner}, we discuss how to generate $v_{T,x}$ and $v_{T,y}$ based on the preplanned trajectory. 
We discretize the linearized system in \eqref{eq:constraint1} with sampling frequency $f=10$\;Hz using the Euler method, and solve the corresponding quadratic program with CVXOPT. The process is summarized in Alg.~\ref{recovery controller}.

Computed control inputs~\eqref{optimal control} and~\eqref{eq:rotational} make the robot detach from the collision surface and help bring it to a temporary post-collision state which can be used as the initial condition for post-impact trajectory generation. We discuss this next.

\begin{algorithm}[h!]
	\caption{Deformation recovery controller}
	\label{recovery controller}
	\LinesNumbered
	\SetKwInOut{Input}{input}
	\SetKwInOut{Output}{output}
	\SetKwInOut{Parameter}{parameter}
	\SetKwFunction{FMain}{\textsc{RecoveryController}}
	\Input{Displacement in body frame $~^{b}\bm{l} - ~^{b}\bm{l}_{s}$ via Hall effect sensors readings; 
	collision time instant $\tau_{c} \in [0, \Delta t_{i_{c}})$; position in world frame at collision instant, $~^{w}\bm{p}_{\tau_{c}}$; velocity in world frame at collision instant, $~^{w}\bm{v}_{\tau_{c}}$; rotation matrix 
	$~^{w}_{b}{R}$; 
	rotation matrix 
	$~^{w}_{c}{R}$; next waypoint point in world frame, $~^{w}\bm{p}_{next}$.}
	\Output{Control input $\bm{u}$}
	\Parameter{Maximum velocity of the robot $v_{max}$}
	\SetKwProg{Fn}{Function}{:}{\KwRet $\bm{u}$}
    \Fn{\FMain{$~^{b}\bm{l} - ~^{b}\bm{l}_{s}$, $\tau_{c}$, $\Delta t_{i_{c}}$, $~^{w}\bm{p}_{\tau_{c}}$, $~^{w}\bm{p}_{next}$, $~^{w}\bm{v}_{\tau_{c}}$, $~^{w}_{b}{R}$, $~^{w}_{c}{R}$}}{
	    $~^{w}\bm{v}_{T} \leftarrow \frac{~^{w}\bm{p}_{next} - ~^{w}\bm{p}_{\tau_{c}}} {\Delta t_{i_{c}} - \tau_{c}}$\\
	     $~^{c}\bm{v}_{T} \leftarrow ~^{w}_{c}{R}^{\top}~^{w}\bm{v}_{T}$\\
	    \If{$~^{c}{v}_{T, x} < 0$}
	    {$~^{c}{v}_{T, x} \leftarrow 0$\\
	    }
	    \If{$\lVert ~^{c}\bm{v}_{T} \rVert \geq v_{max}$}
	    {$~^{c}\bm{v}_{T} \leftarrow v_{max}normalize(~^{c}\bm{v}_{T})$
	    }
	    Calculate $p_{0, x}$ based on \eqref{eq: x0} with $\bm{l}_{b}$\\
	    $p_{0, y} \leftarrow 0$ \\
	    Calculate ${u}_{x}$ and $u_{y}$ based on \eqref{optimal control} and \eqref{eq:feedback linearization} with given $\bm{v}_{T}$ and $\bm{p}_{0}$\\
	    Calculate $u_{\theta}$ based on \eqref{eq:rotational}\\
	    $\bm{u} \leftarrow [u_{x}\ u_{y}\  u_{\theta}]^{\top}$
	 }
\end{algorithm}

\section{Post-impact Trajectory Replanning}\label{sec:replanner}
\subsection{Problem Formulation}\label{subsec:replanner formulation}
We formulate the post-impact trajectory generation problem as a quadratic program with equality constraints, i.e.
\vspace{-6pt}
\begin{subequations} \label{Trajectory}
\begin{alignat}{2}
&\!\min_{{\eta}}   &\qquad& 
   J_{s}(\bm{\eta}) = \sum\limits_{i = i_{c}}^{N_{I}}\int\limits_{0}^{\Delta t_{i}}{\left\lVert ~^{w}\bm{p}_{i}^{(q)}(t)\right\rVert}dt
\label{eq:minimal_effort}\\
&\text{subject to} &      & {C}^{(0)}_{0, i_{c}, \beta}\bm{\eta}_{i_{c}, \beta} = ~^{w}\bm{p}_{r, \beta}, 
\label{eq:me_constraint1}\\
&                  &      & {C}^{(1)}_{0, i_{c}, \beta}\bm{\eta}_{i_{c}, \beta} = ~^{w}\bm{v}_{r, \beta}, 
\label{eq:me_constraint2}\\
&                  &      & {C}^{(\alpha)}_{\Delta t_{N_{I}}, N_{I}, \beta}\bm{\eta}_{N_{I}, \beta} = ~^{w}\bm{d}_{\Delta t_{N_{I}}, N_{I}, \beta}^{(\alpha)}, \notag\\
&                  &      & \alpha = \{ 0, 1 \dots q-1\}, 
\label{eq:me_constraint3}\\
&                  &      & {C}^{0}_{\Delta t_{i}, i, \beta}\bm{\eta}_{i, \beta} = ~^{w}\bm{p}_{i+1, \beta}, 
\label{eq:me_constraint4}\\
&                  &      & {C}_{\Delta t_{i},i+1}^{(\alpha)}\bm{\eta}_{i, \beta} = {C}_{0,i+1, \beta}^{(\alpha)}\bm{\eta}_{i+1, \beta}, \notag\\
&                  &      & \alpha = \{1, 2 \dots q-1\}. \label{eq:me_constraint5}
\end{alignat}
\end{subequations}

For polynomial segments, we can rewrite $J_{s}$ as 
\begin{align*}
    J_{s}=\sum\limits_{\beta \in \{x, y\}}\sum\limits_{i = i_{c}}^{N_{I}}\bm{\eta}_{i, \beta}^{\top}Q^{q}_{\beta}(\Delta t_{i})\bm{\eta}_{i, \beta} \enspace,
\end{align*}
where $i_{c}$ is the segment where the collision happens and $N_{I}$ is the number of trajectory segments.
Superscript $q$ denotes the derivative order; for example, $q = \{1, 2, 3, 4\}$ correspond to min-velocity, min-acceleration, min-jerk and min-snap trajectories, respectively. Subscript $\beta \in \{x, y\}$ indicates the $x$ and $y$ component of the trajectory, and %
$\Delta t_{i}$ is the time duration for $i^{th}$ polynomial segment. 
Parameter ${\eta}_{i, \beta}$ is the vector of coefficients of $i^{th}$ polynomial. 
${C}_{0, i, \beta}^{(\alpha)}$ maps the coefficients to $\alpha^{th}$ order derivative of the start point in segment $i$, while ${C}_{\Delta t_{i}, i, \beta}^{(\alpha)}$ maps the coefficients to $\alpha^{th}$ order derivative of the end point in segment $i$. 

Constraints~\eqref{eq:me_constraint1} and~\eqref{eq:me_constraint2} impose the initial values for the $0^{th}$ and the $1^{st}$ order derivatives to match the position and velocity values attained via the collision recovery controller, respectively. 
Constraint~\eqref{eq:me_constraint3} imposes that the $\alpha^{th}$ order derivatives of the end position are fixed. 
Constraint~\eqref{eq:me_constraint4} imposes that the trajectory will pass through desired waypoints after $i_{c}$. 
Constraint~\eqref{eq:me_constraint5} is imposed to ensure $\alpha^{th}$ continuity among polynomial segments. 

We solve this quadratic programming (QP) problem given initial (post-collision) and end states, and intermediate waypoints. Then, we perform time scaling as in~\cite{liu2017planning} to reduce the maximum values for planned velocities, accelerations and higher-order derivatives as appropriate, and thus improve dynamic feasibility of the refined post-impact trajectory.

The solution of the QP problem serves as the initial value for GTO~\cite{gao2017gradient}, where we change the objective function to
\begin{equation}
   \min   \qquad 
   \lambda_{s} J_{s} + \lambda_{o} J_{o} + \lambda_{d} (J_{v}+J_{a})\enspace,
\end{equation}
where $J_{o}$ is the cost to avoid collisions, and $J_{v}$ and $J_{a}$ are the penalties when candidate velocity and acceleration solutions exceed the dynamic feasibility limit, respectively. Weight parameters $\lambda_{s}$, $\lambda_{o}$ and $\lambda_{d}$ trade off between smoothness, trajectory clearance and dynamical feasibility, respectively.

Similar to~\cite{gao2017gradient}, we use an exponential cost function. At a position with distance $d$ to the closest obstacle, the cost $c_{o}(d)$ is written as 
\begin{equation}\label{eq:cost_d}
   c_{o}(d) = \alpha_{o}\exp{(-d - d_{o})/\gamma_{o}}\enspace,
\end{equation}
where $\alpha_{o}$ is the magnitude of the cost function, $d_{o}$ is the threshold where the cost starts to rapidly rise, and $\gamma_{o}$ controls the rate of the function’s rise.
Then, $J_{o}$ can be computed as
\begin{equation} \label{eq:J_o}
\begin{aligned}
   J_{o} & = \sum\limits_{i = i_{c}}^{N_{I}}\int\limits_{0}^{\Delta t_{i}}c_{o}(\bm{p}(t))\lVert \bm{v}(t) \rVert dt \\
   & = \sum\limits_{i = i_{c}}^{N_{I}}\sum\limits_{k = 0}^{N}c_{o}(\bm{p}(t_{k}))\lVert \bm{v}(t_{k}) \rVert \delta t \enspace.
\end{aligned}
\end{equation}
$J_{v}$ can be computed in a similar manner, whereby $c_{v}(v)$ is the cost function applied on the velocity and attains the same form as in \eqref{eq:cost_d}. We can then obtain  
\begin{equation}\label{eq:J_v}
\begin{aligned}
   J_{v} & = \sum\limits_{\beta \in \{x, y\}}\sum\limits_{i = i_{c}}^{N_{I}}\int\limits_{0}^{\Delta t_{i}}c_{v}({v}_{\beta}(t))\lVert {a}_{\beta}(t) \rVert dt \\
   & = \sum\limits_{\beta \in \{x, y\}} \sum\limits_{i = i_{c}}^{N_{I}}\sum\limits_{k = 0}^{N}c_{v}({v}_{\beta}(t_{k}))\lVert {a}_{\beta}(t_{k}) \rVert \delta t 
\end{aligned}
\end{equation}

The formulation of $J_{a}$ is similar to \eqref{eq:J_v}. The cost function of the acceleration constraint $c_{a}(a)$ is also an exponential function similar to $c_{o}(d)$ and $c_{v}(v)$, since it is can penalize when close to or beyond acceleration bounds while staying flat when away from the bounds. We apply a similar Newton trust region method as in~\cite{gao2017gradient} to optimize the objective.

\begin{figure}[!t]
\vspace{3pt}
 \begin{subfigure}{.235\textwidth}
  \centering
  \includegraphics[trim={0.1cm 0.2cm 0.1cm 0.2cm}, clip, width=0.9\linewidth]{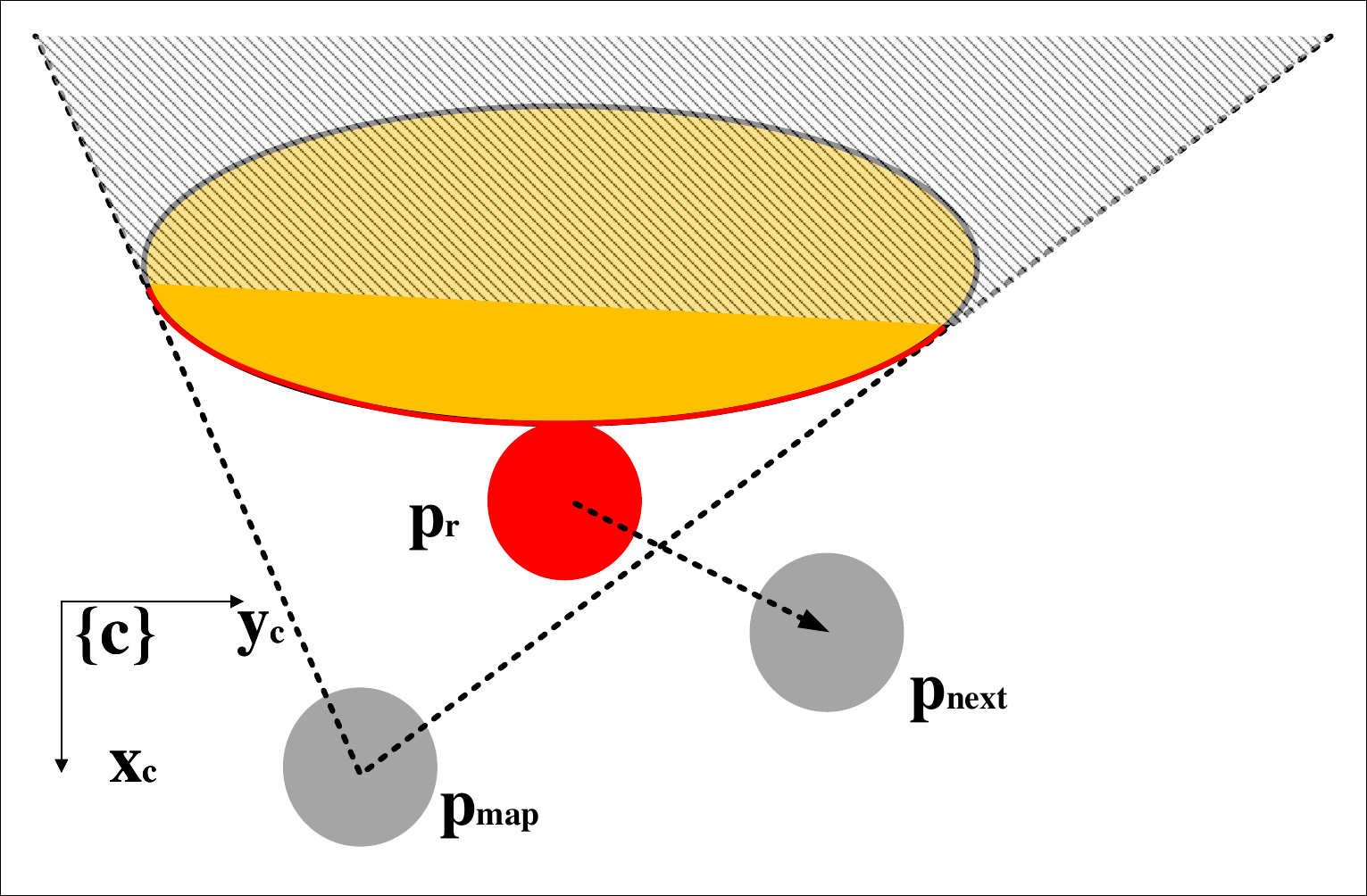}
  \vspace{-4pt}
  \caption{No adjustment made.}
  \label{fig:no adjustment}
 \end{subfigure}%
 \begin{subfigure}{.235\textwidth}
  \centering
  \includegraphics[trim={0.1cm 0.2cm 0.1cm 0.2cm}, clip,width=0.9\linewidth]{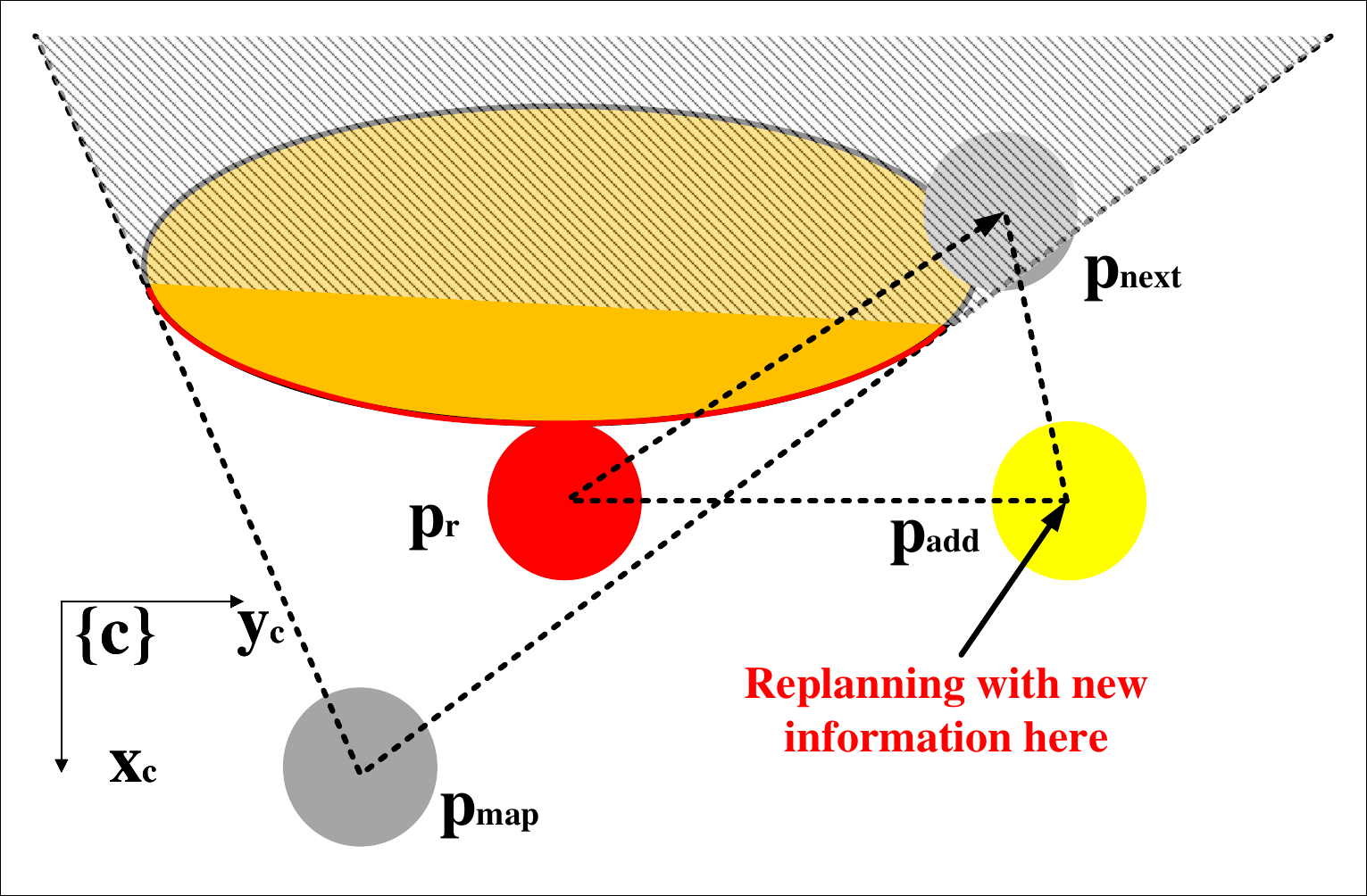}
  \vspace{-4pt}
  \caption{Add new waypoint $\bm{p}_{add}$.}
  \label{fig:adjustment next}
 \end{subfigure}
 \vspace{-3pt}
 \caption{Waypoint adjustment process when the collision is sensed.}
 \label{fig:cases of waypoint adjustment}
 \vspace{-18pt}
\end{figure}

\subsection{Waypoint Adjustment}
In some cases, it may be necessary to adjust the waypoints given by a preplanned trajectory with the information obtained from the collision, and then solve the aforementioned problem in Sec.~\ref{subsec:replanner formulation} with the adjusted waypoints. 
Such cases occur when there is no direct line of sight between the collision state and the waypoint at the end of the immediately next trajectory segment following collision recovery. By enabling such waypoint adjustment, the algorithm promotes exploration and in certain cases prevents the robot from being trapped in a local minima in which repeated collisions at the same (or very close-by) place could otherwise occur. 

With reference to Alg.~\ref{pseudo replanner}, we express in the local collision frame $\mathcal{F}_{c}$ the next waypoint $\mathbf{waypoint\_list}[i_{c} + 1]$ (lines 2--4). In line 5, we adjust $\mathbf{waypoint\_list}$ with the information we get from collision. Details of this process are shown in Fig.~\ref{fig:cases of waypoint adjustment}. We add an additional waypoint $\bm{p}_{add}$ to create a path detouring the collided obstacle. Then we select the shortest path among all the possible paths toward the next waypoint $\bm{p}_{next}$ that was originally in the list before collision. Possible $\bm{p}_{add}$ waypoints are generated by either using a path generation algorithm (e.g., jump point search) when the complete collision surface can be perceived, or by searching along the $y$-axis of collision frame $\mathcal{F}_{c}$ by a (user-defined) exploration distance $\epsilon_{explore}$ when the complete collision surface cannot be reliably perceived (e.g., via LiDAR measurements).

\begin{algorithm}[h!]
	\caption{Post-impact waypoint adjustment}
	\label{pseudo replanner}
	\LinesNumbered
	\SetKwInOut{Input}{input}
	\SetKwInOut{Output}{output}
	\SetKwInOut{Parameter}{parameter}
	\SetKwFunction{FMain}{\textsc{WaypointAdjustmentLine}}
	\Input{Position after the collision recovery in world frame, $~^{w}\bm{p}_{r}$; waypoint list of preplanned trajectory;  $\mathbf{waypoint\_list}$;  $~^{w}_{c}{R}$; trajectory segment $i_{c}$ where the collision happens.}
	\Output{waypoint list after adjustment $\mathbf{waypoint\_list}$}
	\Parameter{Robot radius $r_{rob}$}
	\SetKwProg{Fn}{Function}{:}{\KwRet $\mathbf{waypoint\_list}$}
    \Fn{\FMain{$~^{w}\bm{p}_{r}$, $\mathbf{waypoint\_list}$, $~^{w}_{c}{R}$, $i_{c}$}}{
	$~^{w}\bm{p}_{next} \leftarrow \mathbf{waypoint\_list}[i_{c} + 1]$ \\
	Transfer $~^{w}\bm{p}_{next}$ into $\mathcal{F}_{c}$ frame to get $~^{c}\bm{p}_{next}$\\
	Transfer $~^{w}\bm{p}_{r}$ into $\mathcal{F}_{c}$ frame to get $~^{c}\bm{p}_{r}$\\
	Adjust $\mathbf{waypoint\_list}$ as Fig.~\ref{fig:cases of waypoint adjustment} 
	}
\end{algorithm}

As the robot progresses and reaches the additional waypoint $p_{add}$ that was added following the collision, then it replans based on latest information provided from the perception module. This happens when the robot either reaches the added waypoint (to ensure that the next waypoint is in free space) or it senses another collision from the deformation sensor. In this case, the original $\bm{p}_{next}$ will change as well. Note that this process runs online. In the case that the robot senses a collision before reaching $\bm{p}_{add}$, then it will recover and stop (instead of running the fast replanning approach listed above) and call the global planner to revise larger parts of the trajectory.
If a new waypoint is inserted in the list, we map the path generated by $^{w}{p}_{r}$ and waypoints in the list after $i_{c} + 1$ into time domain using a trapezoidal velocity profile. If no new waypoint is inserted, we set the time duration of $i_{c}$ segment in~\eqref{Trajectory} as $\Delta t_{i_{c}} = t_{i_{c}+1} - t_{c}$, where $t_{i_{c}+1}$ is the time reaching next waypoint ${p}_{i_{c}+1}$ in the preplanned trajectory.

\section{Search-based Collision-inclusive Planning}\label{sec:search-based}
In this section, we propose the main algorithm to generate the waypoint list and trajectory segments that serve as the input to the DRR strategy.

\subsection{Problem Formulation} \label{subsec:problem formulation}


Let the system state $\bm{s}_{d}(t) \in \mathcal{S} \subset \mathbb{R}^{n\times q}$ contain the configuration and the $(q - 1)^{th}$-order derivatives in 2D (i.e. $n = 2$). The free state space, $\mathcal{S}^{free} \subset \mathcal{S}$, contains both obstacle-free configurations, $\mathcal{P}^{free}$, as well as the system's dynamical constraints,  $\mathcal{D}^{free}$, which include minimum and maximum bounds on velocity $[v_{min}$, $v_{max}]$, acceleration $[a_{min}$, $a_{max}]$, jerk $[j_{min}$, $j_{max}]$, and other higher-order derivatives. We can then write $\mathcal{S}^{free} = \mathcal{P}^{free} \times \mathcal{D}^{free} = \mathcal{P}^{free} \times [v_{min},v_{max}] \times [a_{min},a_{max}] \times \cdots$. $\mathcal{P}^{obs} = \mathcal{P} \setminus \mathcal{P}^{free}$ and $\mathcal{S}^{obs} = \mathcal{P}^{obs} \times \mathcal{D}^{free}$ defines the obstacle region.

The differential flatness of some mobile robot systems (e.g.,~\cite{liu2017search}) helps design control inputs from 1D time-parameterized polynomial trajectories independently for each of the $n$ positions. Hence, $\bm{s}_{d}(t) = [\bm{p}_{D}(t)^{\top}, \dot{\bm{p}}_{D}(t)^{\top}, \cdots, \bm{p}^{(q - 1)}_{D}(t)^{\top}]^{\top}$ where $\bm{p}_{D}(t) = \sum\limits_{i=0}^{q} \bm{d}_{i} \frac{t^{i}}{i!}$ and $\bm{D} = [\bm{d}_{0}, \cdots, \bm{d}_{q}] \in \mathbb{R}^{n\times (q+1)}$, and $\bm{d}_{i}=[{d}_{i, x}\ {d}_{i, y}]^{\top}$ in~\eqref{Trajectory}. To simplify the notation, we re-express the derivatives as $\bm{v}(t) = \dot{\bm{p}}_{D}^{\top}(t)$, $\bm{a}(t) = \ddot{\bm{p}}_{D}^{\top}(t)$, $\bm{j}(t) = \dddot{\bm{p}}_{D}^{\top}(t)$, etc., and drop subscript ${D}$. 

We can construct the polynomial trajectories
via $\bm{p}^{(q)}_{D}(t) = \bm{u}(t)$ with controls $\bm{u}(t) \in \mathcal{U} = [-u_{max}, u_{max}]^{n} \subset \mathbb{R}^{n}$.
In state space form this yields 
$\dot{\bm{s}}_{d}(t)={A}_{f}\bm{s}_{d}(t) + {B}_{f}\bm{u}(t)$, with 
\begin{equation} \label{system free}
{A}_{f} = \begin{bmatrix}
{0} & {I}_{n} & {0} & \cdots &{0} \\
{0} & {0} & {I}_{n} & \cdots &{0} \\
\vdots & \ddots & \ddots & \ddots & \vdots \\
{0} & \cdots & \cdots & {0} & {I}_{n} \\
{0} & \cdots & \cdots & {0} &{0} 
\end{bmatrix}, {B}_{f} = \begin{bmatrix}
{0} \\
{0} \\
\vdots \\
{0} \\
{I}_{n}
\end{bmatrix}\enspace.
\end{equation}

In collision-inclusive planning, we consider a smoothness cost $J_{s}(\bm{D}) = \sum\limits_{k=1}^{K} \int\limits_{0}^{T_{k}} \lVert \bm{u}(t) \rVert^{2}dt = \sum\limits_{k=1}^{K} \int\limits_{0}^{T_{k}} \lVert \bm{p}^{(q)}_{D, k}(t) \rVert^{2}dt $. The trajectory is not $q$th order differentiable as it would be in collision avoidance. The smoothness of the entire trajectory is the sum of its $q$th order differentiable segments. We consider two additional costs. First, $T_{g} = \sum\limits_{k=1}^{K} T_{k} + (K-1) T_{r}$ penalizes the overall trajectory duration. 
Then, 
\begin{equation}\label{collsion cost}
J_{c}= \begin{cases} \frac{(\left| ~^{c}{v}_{x}^{+}\right| -  \left| ~^{c}{v}_{x}^{-}\right|)^{2} + \sum\limits_{\beta \in \{y,z\}}\Delta E_{\beta}}{T_{r}} & \forall \zeta(t) = 1\\
0 & otherwise
\end{cases}
\end{equation}
evaluates the effect of a collision in changing the direction of motion of the robot.  $\Delta E_{\beta} = (~^{c}{v}_{\beta}^{+} - ~^{c}{v}_{\beta}^{-})^{2}$, where $~^{c}{v}_{\beta}^{+} = ~^{c}{v}_{\beta}(t + T_{r})$ and $~^{c}{v}_{\beta}^{-} = ~^{c}{v}_{\beta}(t)$. $~^{c}{v}_{\beta}^{+}$ can be approximated via Alg.~\ref{post collision state} if $\bm{p}_{goal}$ is known. (We discuss Alg.~\ref{post collision state} in detail in Sec.~\ref{subsec:collision check}.) 
We also define an indicator function $\zeta(t)= \{0,1\}$ that signals if the robot is colliding at time $t$.

We can then define the optimization problem 
\vspace{-6pt}
\begin{subequations} \label{motion planning inclusive}
\begin{alignat}{2}
&\!\min_{\bm{D}, T_{g}}   &\qquad& 
   J_{s}(\bm{D}) + \rho_{t} T_{g} + \rho_{c}J_{c}(t)
\label{eq:minimal_effort_mpci}\\
&\text{subject to} &      &\dot{\bm{s}}_{d}(t)={A}_{f}\bm{s}_{d}(t) + {B}_{f}\bm{u}(t), {\forall} \zeta(t) = 0, 
\notag\\
&                  &      &{\forall} t \in [0, T_{g}]
\label{eq:mpci_constraint1}\\
&                  &      &\bm{s}_{d}(t+T_{r})=F_{DRR}(\bm{s}_{d}(t)), {\forall} \zeta(t) = 1, 
\notag\\
&                  &      &{\forall} t \in [0, T_{g}-T_{r}]
\label{eq:mpci_constraint2}\\
&                  &      &\zeta(t) \in \{0, 1\}, {\forall} t \in [0, T_{g}]
\label{eq:mpci_constraint3}\\
&                  &      &\zeta(t) = 0, if\ \bm{s}_{d}(t + \delta t) \in \mathcal{S}^{free}, \delta t \rightarrow 0, 
\notag\\
&                  &      &{\forall} t \in [0, T_{g}]
\label{eq:mpci_constraint4}\\
&                  &      &\zeta(t) = 1, if\ \bm{s}_{d}(t + \delta t) \in \mathcal{S}^{obs}, \delta t \rightarrow 0, 
\notag\\
&                  &      &{\forall} t \in [0, T_{g} - T_{r}]
\label{eq:mpci_constraint5}\\
&                  &      &\bm{s}_{d}(0) = \bm{s}_{d0}, \bm{s}_{d}(T_{g}) \in \mathcal{S}^{goal},
\notag\\
&                  &      &\zeta(0) = 0, \zeta(T_{g}) = 0 
\label{eq:mpci_constraint6}\\
&                  &      &\bm{s}_{d}(t) \in \mathcal{S}^{free}, \bm{u}(t) \in \mathcal{U}, {\forall} t \in [0, T_{g}]
\label{eq:mpci_constraint7}\\
&                  &      &^{c}{v}_{x}(t) \in \mathcal{V}^{c}, if\ \zeta(t) = 1\enspace.
\label{eq:mpci_constraint8}
\end{alignat}
\end{subequations}

Parameters $\rho_{t} > 0$ and $\rho_{c} > 0$ regulate the relative importance of trajectory smoothness, duration, and amount of collisions that switch the direction of motion. Conditions~\eqref{eq:mpci_constraint4} and~\eqref{eq:mpci_constraint5} determine how the value for $\zeta(t)$ is being set. In~\eqref{eq:mpci_constraint8}, $\mathcal{V}^{c} = [-v_{max, c}, v_{max, c}]$; $v_{max, c}$ indicates the maximum collision velocity which, if exceeded, will lead to the robot flipping over. Thus, we set the pre-collision velocity component along $x$ axis of $\mathcal{F}_{c}$ as ${v}_{x}(t) \in \mathcal{V}^{c}$. 

Herein we show that, similar to the collision avoidance motion planning problem~\cite{liu2017search}, safety constraints may be addressed by reformulating problem~\eqref{motion planning inclusive} into a deterministic shortest path one with $(n \times q)$ state $\mathcal{S}$ and $n$ control $\mathcal{U}$. Since the dimensionality of $\mathcal{U}$ remains at $n$, search-based planning (e.g., A*~\cite{likhachev2003ara}) that discretizes $\mathcal{U}$ using motion primitives can be an effective way to determine in finite-time resolution-complete and optimal (in the discretized space) trajectories.

\subsection{Motion Primitives}\label{subsec:motion primitives}
Choosing a number of samples $r \in \mathbb{Z}^{+}$ along each axis $[-u_{max}, u_{max}]$, which defines a discretization step: $du = \frac{u_{max}}{r}$ and results in $M = (2r + 1)^{n}$ motion primitives, is one technique to acquire the discretization $\mathcal{U}_{M}$.
Given initial state $\bm{s}_{d0} = [\bm{p}_{0}^{\top}, \bm{v}_{0}^{\top}, \cdots]^{\top}$, we generate a motion primitive of duration $\tau > 0$ that applies piece-wise constant control 
\begin{equation} \label{calculate input}
\tilde{\bm{u}}_{m}(t) = \begin{cases}
\bm{u}_{m}  & \bm{p}_{D}^{(q-1)}(t) \in [\bm{p}_{D, min}^{(q-1)}, \bm{p}_{D, max}^{(q-1)}], \\
\bm{0}  & \bm{p}_{D}^{(q-1)}(t) \not \in [\bm{p}_{D, min}^{(q-1)}, \bm{p}_{D, max}^{(q-1)}]
\end{cases}
\end{equation}
where ${u}_{m} \in \mathcal{U}_{M}$ for $t \in [0, \tau]$. 
Given initial conditions, 
\begin{equation} \label{motion primitive PD}
\bm{p}_{D}(t) = \tilde{\bm{u}}_{m}(t)\frac{t^{q}}{q!} + \cdots + \bm{v}_{0} t + \bm{p}_{0}
\end{equation}
is a piece-wise function. The resulting trajectory of~\eqref{system free} is
\begin{equation} \label{system free discretize}
\begin{matrix} \bm{s}_{d}(t) = \\
~ \end{matrix}
\begin{matrix} \underbrace{e^{{A}_{f} t}} \\ {A}_{df}(t) \end{matrix} 
\begin{matrix} \bm{s}_{d0} + \\
~ \end{matrix} 
\begin{matrix} \underbrace{\int_{0}^{t} e^{{A}_{f} (t-\sigma)}B_{f} \tilde{\bm{u}}_{m}(\sigma) d\sigma}\enspace. \\ {B}_{df}(t){\bm{u}_{m}} \end{matrix}
\end{equation}

By beginning at $\bm{s}_{d0}$ and applying all primitives to acquire the $M$ possible states after $\tau \in [0, \tau_{f}]$ (Alg.~\ref{motion primitive}), we can create a graph representation of the attainable system states. There will be $M^{2}$ potential states at time $2\tau$ if all primitives are applied to each of the $M$ states once again. The set of reachable states $\mathcal{S}^{r}$ is finite given the free space $\mathcal{S}^{free}$ is bounded. These enable the construction of a graph the states of which are connected by a motion primitive $\bm{e} = (\tilde{\bm{u}}_{m}, \tau, \xi)$ with $\xi$ being an integer (discussed in Sec.~\ref{subsec:collision check}).

\begin{algorithm}[!h]
	\caption{ Collision-inclusive motion primitive generation}
	\label{motion primitive}
	\LinesNumbered
	\SetKwInOut{Input}{input}
	\SetKwInOut{Output}{output}
	\SetKwInOut{Parameter}{parameter}
	\SetKwFunction{FMain}{\textsc{GetMotionPrimitive}}
	\Input{Initial state $\bm{s}_{d} \in \mathcal{S}^{r} \subset \mathcal{S}^{free}$; motion primitive set $\mathcal{U}_{M}$, upper-bound of duration $\tau_{f}$}
	\Output{Reachable set $\mathcal{R}(\bm{s}_{d})$ from $\bm{s}_{d}$ in one step; costs set $\mathcal{C}(\bm{s}_{d})$; duration set $\mathcal{T}(\bm{s}_{d})$; collision states set $\mathcal{Z}(\bm{s}_{d})$}
	\Parameter{Time interval $\delta t$; recover time $T_{r}$ in DRR}
	\SetKwProg{Fn}{Function}{:}{\KwRet $\mathcal{R}(\bm{s}_{d}),\ \mathcal{C}(\bm{s}_{d}),\ \mathcal{T}(\bm{s}_{d}),\ \mathcal{Z}(\bm{s}_{d})$}
    \Fn{\FMain{$\bm{s}_{d}$, $\mathcal{U}_{M}$, $\tau_{f}$}}{
        $\mathcal{R}({\bm{s}_{d}}) \leftarrow \emptyset$, $\mathcal{C}({\bm{s}_{d}}) \leftarrow \emptyset$, $\mathcal{T}(\bm{s}_{d}) \leftarrow \emptyset$, $\mathcal{Z}({\bm{s}_{d}}) \leftarrow \emptyset$\\
        \For{\textup{\textbf{all}} $\bm{u}_{m} \in \mathcal{U}_{M}$}
        {
            Calculate edge $e_{m}(t)$ according to \eqref{system free discretize} for $t \in [0, \tau_{f}]$\\
            \eIf{$e(t) \in \mathcal{S}^{free}\ \textup{\textbf{for all}}\ t \in [0, \tau_{f}]$}
            {
            $\zeta_{m} \leftarrow 0$\\
            $\tau_{m} \leftarrow \tau_{f}$\\
            $\bm{s}_{d,m} \leftarrow e_{m}(\tau_{f})$\\
            $\mathcal{R}({s}_{d}) \leftarrow \mathcal{R}({s}_{d}) \bigcup \{ {s}_{dm} \}$\\
            $J_{D} \leftarrow \int_{0}^{\tau}\lVert \tilde{\bm{u}}_{m}(t) \rVert^{2} dt$\\
            $\mathcal{C}(\bm{s}_{d}) \leftarrow \mathcal{C}(\bm{s}_{d}) \bigcup \{J_{D} + \rho_{t} \tau_{f}\}$\\
            $\mathcal{T}(\bm{s}_{d}) \leftarrow \mathcal{T}(\bm{s}_{d}) \bigcup \{ \tau_{m} \}$\\
            $\zeta(\bm{s}_{d}) \leftarrow \mathcal{Z}(\bm{s}_{d}) \bigcup \{ \zeta_{m} \}$\\
            }
            {
            $\zeta_{m} \leftarrow 1$\\
            Generate $\bm{s}_{d,m}$, $\tau$ and calculate $J_{c}$ or prune this primitive (discussed in \ref{subsec:collision check} and \ref{subsec:jump point}). \\
            $\mathcal{R}(\bm{s}_{d}) \leftarrow \mathcal{R}(\bm{s}_{d}) \bigcup \{ \bm{s}_{d,m} \}$\\
            $J_{D} \leftarrow \int_{0}^{\tau}\lVert \tilde{\bm{u}}_{m}(t) \rVert^{2} dt$\\
            $\mathcal{C}(\bm{s}_{d}) \leftarrow \mathcal{C}(\bm{s}_{d}) \bigcup \{J_{D} + \rho_{t} (\tau + T_{r}) + J_{c}\}$\\
            $\mathcal{T}(\bm{s}_{d}) \leftarrow \mathcal{T}(\bm{s}_{d}) \bigcup \{ \tau_{m} \}$\\
            $\zeta(\bm{s}_{d}) \leftarrow \mathcal{Z}(\bm{s}_{d}) \bigcup \{ \zeta_{m} \}$\\
            }
        }
  } 
\end{algorithm}

We construct the graph to explore the free state space $\mathcal{S}^{free}$ using Alg.~\ref{motion primitive}. Given the constant time upper-bound $\tau_{f}$ and the fully specified state $\bm{s}$, the primitive is derived in line 4 using the control input $\bm{u}_{m}$;
lines 5--23 check whether the primitive intersects with the obstacles and then modify those primitives intersecting with the obstacles. This step will be further discussed Sec.~\ref{subsec:collision check}. In lines 6--13, we evaluate the end state of a valid primitive not intersecting with the obstacles and we add it to the set of successors of the current node; meanwhile, we estimate the edge cost from the corresponding primitive. In lines 16--21, we modify the end state of the primitive and add it to the set of successors of the current node; meanwhile, we estimate the edge cost related to the corresponding modified primitive. Line 19 shows that we consider $T_{r}$ for the robot recovering from the collision using DRR in the cost function. Further modification of the cost function about estimating the cost related to $J_{c}$ part will be discussed in Sec.~\ref{subsec:collision check}. 
The nodes in the successor set $\mathcal{R}(\bm{s}_{d})$ are added to the graph after we have checked all the primitives in the finite control input set. Finally, the graph keeps growing until we reach the goal is reached.

\subsection{Deterministic Shortest Trajectory} \label{subsec:shortest trajectory}
We can re-formulate~\eqref{motion planning inclusive} as a graph-search problem using the set of motion primitives $\mathcal{U}_{M}$ and the induced discretization. To do so, we introduce additional constraints for the control input $\bm{u}(t)$ in~\eqref{motion planning inclusive} to be piecewise-constant.
We introduce an additional variable $N \in \mathbb{Z}^{+}$, so that $T_{g} = \sum\limits_{k=0}^{N-1} (\tau_{k} + \zeta_{k+1} T_{r})$, and $\tilde{\bm{u}}_{k}$ is computed by~\eqref{calculate input} with $\bm{u}_{k} \in \mathcal{U}_{M}$ for $k = 0, \cdots, N-1$ and a constraint in \eqref{eq:mpci_constraint7}: 
\begin{align*}
    \bm{u}(t) = \sum\limits_{k=0}^{N-1} \tilde{\bm{u}}_{k} \mathbbm{1}_{t \in [T_{k}, T_{k + 1}]}\enspace.
\end{align*}
By letting $T_{i} = \sum\limits_{k=0}^{i-1} \tau_{k}$ we can force the control trajectory to be a composition of the motion primitives in $\mathcal{U}_{M}$. 
Given an initial state $\bm{s}_{d0} \in \mathcal{S}^{free}$, a goal area $\mathcal{S}^{goal}$ and a finite set of motion primitives $\mathcal{U}_{M}$ with duration $\tau > 0$, we seek to select a series of motion primitives $\bm{u}_{0:N-1}$ of length $N$, such that
\begin{subequations} \label{mpci serach based planning}
\begin{alignat}{2}
&\!\min_{N, {u}_{0:N-1}}   &\qquad & 
   \sum\limits_{k=0}^{N-1} \lVert \bm{u}_{k} \rVert^{2} + \rho_{t} (\tau_{k} + \zeta_{k+1}T_{r}) + \rho_{c} J_{c,k}
\label{eq:minimal_effort_mpci_search}\\
&\text{subject to} &      &\bm{s}_{d}(\tilde{t})={A}_{df}(\tilde{t})\bm{s}_{d, k} + {B}_{df}(\tilde{t})\bm{u}_{k} \subset \mathcal{S}^{free},  
\notag\\
&                  &      &{\forall} \tilde{t} \in [0, \tau_{k}]
\label{eq:mpci_search_constraint1}\\
&                  &      &\zeta_{k} \in \{0, 1\}, {\forall} k \in \{0, 1, \cdots N-1\}
\label{eq:mpci_search_constraint2}\\
&                  &      &\zeta_{k+1} = 0, if\ \bm{s}_{d}(\tau_{k} + \delta t) \in \mathcal{S}^{free}, \delta t \rightarrow 0
\label{eq:mpci_search_constraint3}\\
&                  &      &\zeta_{k+1} = 1, if\ \bm{s}_{d}(\tau_{k} + \delta t) \in \mathcal{S}^{obs}, \delta t \rightarrow 0
\label{eq:mpci_search_constraint4}\\
&                  &      &\bm{s}_{d,k+1} = \bm{s}_{d}(\tau_{k}), {\forall} \zeta_{k+1} = 0
\label{eq:mpci_search_constraint5}\\
&                  &      &\bm{s}_{d,k+1} = F_{DRR}(\bm{s}_{d}(\tau_{k})), {\forall} \zeta_{k+1} = 1
\label{eq:mpci_search_constraint6}\\
&                  &      &\bm{s}_{d,0} = \bm{s}_{d0}, \bm{s}_{d,N} \in \mathcal{S}^{goal}, \zeta_{0} = 0, \zeta_{N} = 0
\label{eq:mpci_search_constraint7}\\
&                  &      &\bm{u}_{k} \in \mathcal{U}_{M}
\label{eq:mpci_search_constraint8}\\
&                  &      &~^{c}{v}_{k, x}^{-} \in \mathcal{V}^{c}, if\ \zeta_{k} = 1\enspace.
\label{eq:mpci_search_constraint9}
\end{alignat}
\end{subequations}

The optimal cost of~\eqref{mpci serach based planning} is an upper bound to the optimal cost of~\eqref{motion planning inclusive} because~\eqref{mpci serach based planning} is a constrained version of~\eqref{motion planning inclusive}. 
The whole trajectory consists of a set of continuous and collision free primitives of $\tau_{k}$ duration and initial state $\bm{s}_{d, k}$. If the end state of the primitive $\bm{s}_{d}(\tau_{k})$ is state which collides with an obstacle, we modify it based on~\eqref{eq:mpci_search_constraint6}. We modify the final state based on DRR controller model. We make the modified final state as initial state of next primitive. If the end state of the primitive $\bm{s}_{d}(\tau_{k})$ is collision free, we keep the final state similar the collision-avoidance planner making the final state as initial state of the next primitive as \eqref{eq:mpci_search_constraint6}.
Reformulating into a discrete control problem enables the use of several motion planning methods that rely on search-based~\cite{zhou2019robust} or sampling-based~\cite{zha2021exploiting} techniques. We choose to adopt an A* technique similar to~\cite{liu2017search} and concentrate on the creation of effective, guaranteed collision checking and post-collision behavior categorizing methods, as well as an accurate and consistent heuristic since the former can ensure limited time (sub-)optimality. 
\footnote{~We note here that in principle both a search-based (as herein) and a sampling-based global planner is possible. In Section~\ref{subsec:global planner} we demonstrate the differences of the two within collision-inclusive motion planning.}
\vspace{-6pt}

\subsection{Collision Checking and Post-collision Behaviors} \label{subsec:collision check}
For a computed edge $\bm{e}(t) = [\bm{p}(t)^{\top}\ \bm{v}(t)^{\top}\ \bm{a}(t)^{\top}\cdots]^{\top}$, in Alg.~\ref{motion primitive}, we need to check if $\bm{e}(t) \in \mathcal{S}^{free}$ for all $t \in [0, \tau_{f}]$.  
For $\bm{e}(t) \in \mathcal{S}^{free} \wedge \bm{e}(t+\delta t) \in \mathcal{S}^{obs}$ with $\delta t \rightarrow 0$ for all $t \in [0, \tau_{f}]$, we need to modify the edge $\bm{e}(t)$ as in lines 16--21 in Alg.~\ref{motion primitive}. We check collisions in the geometric space $\mathcal{P}^{free} \subset \mathbb{R}^{n}$ separately from enforcing dynamic constraints $\mathcal{D}^{free} \subset \mathbb{R}^{n \times (q-1)}$. An edge $\bm{e}(t)$ is collision-free only if its geometric shape $\bm{p}_{e}(t) \in \mathcal{P}^{free}$ for all $t \in [0, \tau_{f}]$. 

In general, determining collision points for each motion primitive can be very challenging. Herein we model $\mathcal{P}$ as an occupancy grid map, $\mathcal{M}_{o}$. Other representations such as polyhedral maps~\cite{mote2020collision, liu2017planning, deits2015efficient} are also possible but often hard to obtain from a robot's FoV sensor data (e.g., from LiDAR) and hence not pursued herein. Let $\bm{P}_{e} = \{\bm{p}_{e}(t_{i}) \mid t_{i} \in [0, \tau_{f}], i=1, \cdots I \}$ be a set of positions that the system traverses along the trajectory. For collision-free primitives we need $\bm{p}_{e}(t_{i}) \in \mathcal{P}^{free}$ for all $i \in \{0, \cdots I\}$. The duration of the collision-free trajectory is $\tau = \tau_{f}$. For the given polynomial $\bm{p}_{e}(t)$, $t \in [0, \tau_{f}]$, the positions $\bm{p}_{e}(t_{i})$ are sampled by defining 
\begin{equation}
t_{i} = \frac{i}{I}\tau_{f} \qquad \textup{such that} \qquad \frac{\tau}{I}v_{max}\geq \epsilon_{map}\enspace,
\end{equation}
where $\epsilon_{map}$ is the occupancy grid resolution, and 
$v_{max} = \max\{\lvert v_{min} \rvert, \lvert v_{max} \rvert \}$. This condition ensures that the maximum distance between two consecutive samples will not exceed the map resolution.  Since it is an approximation, some cells traversed by $\bm{p}_{e}(t)$ with a portion of the curve within the cell shorter than $\epsilon_{map}$ may be missed, but it guarantees the collision-free trajectory does not hit any obstacles.

In not collision-free $\bm{e}(t)$, the estimated collision time instant $t_{i}$ is when $\bm{p}_{e}(t_{i}) \in \mathcal{P}^{free} \wedge \bm{p}_{e}(t_{i}+\delta t) \in \mathcal{P}^{obs}$ with $\delta t \approx t_{i+1} - t_{i}$ for all $i \in \{0, \cdots I - 1 \}$. Then, we set the duration $\tau$ of the collision-inclusive motion primitives in Alg.~\ref{motion primitive} to $t_{i}$, and modify the end state $\bm{s}_{d,e}$ of $\bm{e}(t)$ as $\bm{s}_{d,e} = \{ \bm{s}_{d,e}^{-}, \bm{s}_{d,e}^{+} \}$ with $\bm{s}_{d,e}^{-} = \bm{e}(t_{i}) = [\bm{p}_{e}(t_{i})^{\top}\ \bm{v}_{e}(t_{i})^{\top}\ \bm{a}_{e}(t_{i})^{\top}\cdots]^{\top}$. We set the duration of this edge $\tau = t_{i}$ and set $\zeta(\tau) = 1$. $\bm{s}_{d,e}^{+} = F_{DRR}(\bm{s}_{d,e}^{-})$ is the post-impact state recovered using the DRR strategy. We discuss how to set $\bm{s}_{d,e}^{+}$ shortly. 

Since $\bm{v}, \bm{a}$ and other higher-order derivatives are polynomial functions, we can compute their extrema within the time period $[0, \tau]$ to check if the respective maximum bounds are violated. 
The polynomials' order is less than $5$ for $n \leq 3$, hence the extrema can be computed quickly in closed form. We eliminate the primitives that cannot be dynamically implemented (i.e. any bounds are exceeded).
For the collision-inclusive primitives, we need to check the $x$ component of the velocity ${v}_{e}^{-}$ in $\mathcal{F}_{c}$ corresponding $~^{c}{v}_{e, x}^{-} \in \mathcal{V}^{c}$. We prune those with $~^{c}{v}_{e, x}^{-} \not \in \mathcal{V}^{c}$ to prevent the robot from flipping over after colliding. 

To generate the frame $\mathcal{F}_{c}$ required for evaluating the collision-inclusive primitives, we need to get the geometric information of each obstacle that the robot collides on. Given a current scan from the mapping sensor (e.g., a LiDAR) we identify all possible collision surfaces and use regression to fit curve equations to the possible collision surfaces. The value of doing so is that it enables a rapid calculation of the tangent and normal unit vectors at selected possible collision points on those collision surfaces. 
Basis vectors of $\mathcal{F}_{c}$ are generated as discussed in Sec.~\ref{sec:deformation problem} whereas the origin of $\mathcal{F}_{c}$ is set to be the estimated position of collision $\bm{p}_{e}^{-}$ in $\bm{s}_{d,e}^{-}$. 

\begin{figure}[!t]
\vspace{6pt}
\centering
\includegraphics[trim={0.3cm 0.3cm 0.3cm 0.3cm}, clip, width=0.3\textwidth]{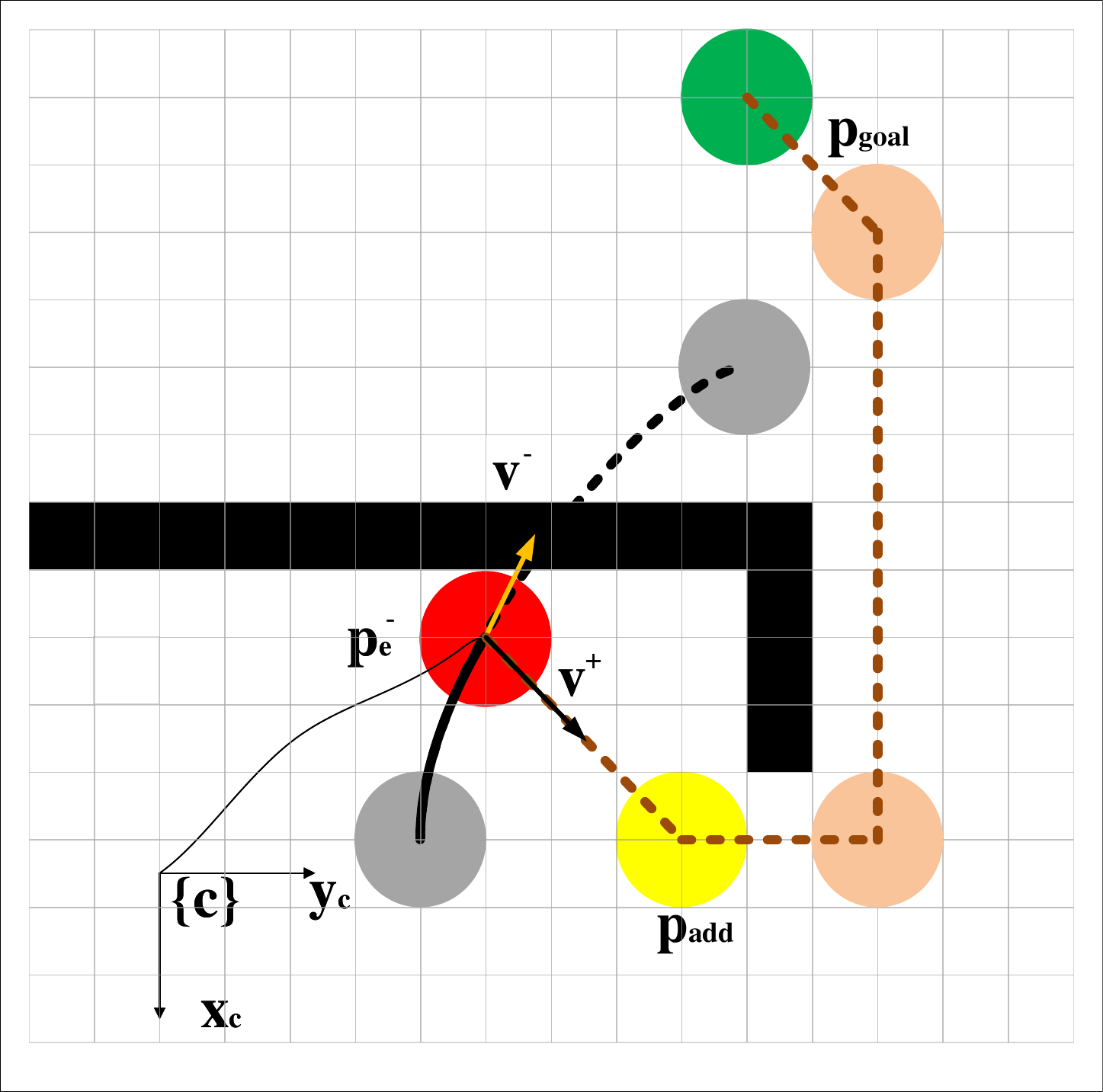}
\vspace{-1pt}
\caption{An example of performing collision and detouring away from a non-convex obstacle by generating a new waypoint between the point of collision and the goal.}
      \label{fig:new waypoint generation}
	\vspace{-3pt}
\end{figure}

After generating $\mathcal{F}_{c}$, we are able to generate $\bm{s}_{d,e}^{+}$ based on the map $\mathcal{M}_{o}$ which we predict the robot will collide on when arriving at $\bm{s}_{d,e}^{-}$ with the given motion primitive. Given the goal position $\bm{p}_{goal}$, we are able to set $\bm{s}_{d, e}^{+}$ according to Alg.~\ref{post collision state}. This way, we can ensure the trajectory generated by the search-based algorithm respect constraint~\eqref{eq:mpci_search_constraint6}. In Fig.~\ref{fig:new waypoint generation}, we show how to generate the intermediate waypoint $\bm{p}_{add}$ based on the jump point search algorithm. If there is no feasible path to the goal, we prune this collision-inclusive motion primitive. Given $^{c}\bm{v}_{e}^{-}$ and $^{c}\bm{v}_{e}^{+}$, we can generate $J_{c}$ of this collision-inclusive motion primitive according to \eqref{collsion cost}.\footnote{~We consider that most of the collision energy can be recovered by the robot via its compliant arms. In practice, precise computation of the dissipated energy is a challenge; however, the DRR strategy accommodates for collision energy losses without any explicit energy dissipation models.}
We set a lower bound to $J_{c}$, 
$\underline{J}_{c}$, to induce a cost if the robot tries to use collisions alone to steer. Tuning $\rho_{c}$ help regulate collision-avoiding and collision-inclusive trajectories. 

We also create an infeasible, $\mathcal{P}^{inf}$, area to link pruned collision-inclusive $\bm{p}_{e}^{-} \in \mathcal{P}^{inf}$. We apply $\mathcal{P}^{inf}$ to prevent the robot from getting into areas where the collisions are difficult to detect using this arm design (i.e. when collision surfaces reduce to almost a point, such as obstacle corners). 


\newlength{\textfloatsepsave} \setlength{\textfloatsepsave}{\textfloatsep} \setlength{\textfloatsep}{3pt}
\begin{algorithm}[h!]
	\caption{Post-collision state generation}
	\label{post collision state}
	\LinesNumbered
	\SetKwInOut{Input}{input}
	\SetKwInOut{Output}{output}
	\SetKwInOut{Parameter}{parameter}
	\SetKwFunction{FMain}{\textsc{GetPostCollisionState}}
	\Input{Pre-collision state in world frame $\bm{s}_{d, e}^{-}$; collision frame $\mathcal{F}_{c}$;  
	goal position in world frame $\bm{p}_{goal}$; $\mathcal{M}_{o}$.}
	\Output{Post-collision state in world frame $\bm{s}_{d,e}^{+}$; behavior type after collision $\xi$.}
	\Parameter{Lower bound and upper bound of the velocity $v_{min}, v_{max}$; upper bound of duration of each primitives $\tau_{f}$, $\delta t$. }
	\SetKwProg{Fn}{Function}{:}{\KwRet $\bm{s}_{d,e}^{+},\ \xi$}
    \Fn{\FMain{$\bm{s}_{d, e}^{-}$, $\bm{p}_{goal}$, $\mathcal{M}_{o}$}}{
	    $\bm{v}^{+} \leftarrow \frac{\bm{p}_{goal} - \bm{p}_{e}^{-}} {\tau_{f}}$\\
	    Generate $~^{c}\bm{v}^{+}$ based on $\mathcal{F}_{c}$ as what is shown in Fig.~\ref{fig:new waypoint generation}\\
	    \eIf{$~^{c}{v}_{x}^{+} < 0$}
	    {$~^{c}{v}_{x}^{+} \leftarrow 0$\\
	    $\xi \leftarrow 2$\\
	    We generate a intermediate waypoint $\bm{p}_{add}$ as what is shown in Fig.~\ref{fig:new waypoint generation} given $\mathcal{M}_{o}$ based on jump point search algorithm. The waypoint should be the last visible waypoint along the path. 
	    $\bm{v}^{+} \leftarrow \frac{\bm{p}_{add} - \bm{p}_{e}^{-}}{\tau_{f}}$
	    }
	    {
	    $\xi \leftarrow 1$\\
	    $\bm{p}_{add} \leftarrow \emptyset$ \\
	    }
	    We adjust each components of $\bm{v}^{+}$ with a saturation function restricting upper bound and lower bound as $v_{min}$ and $v_{max}$. \\
	    We set all derivatives of $\bm{s}_{d,e}^{+}$ as $\bm{p}_{e}^{(q)\ +} = 0$, for all $q \geq 2$, $\bm{p}_{e}^{(1)\ +} = \bm{v}^{+}$, $\bm{p}_{e}^{+} = \bm{p}_{e}^{-}$. \\
	 }
\end{algorithm}

\subsection{Heuristic Function Design}\label{subsec:heuristic}
A heuristic function that is admissible, informative (i.e. provides a tight approximation of the optimal cost), and consistent (i.e. it can be inflated to obtain solutions with bounded sub-optimality efficiently) is required for efficient graph search to solve \eqref{mpci serach based planning}. Similar to~\cite{liu2017search}, we solve a relaxed form of~\eqref{motion planning inclusive} and arrive at a reasonable heuristic function. 
The basic concept is to replace the difficult-to-satisfy $\bm{s}_{d}(t) \in \mathcal{S}^{free}$ and $\bm{u}(t) \in \mathcal{U}$ requirements in~\eqref{motion planning inclusive} with a constraint on time $T$. Next, we demonstrate that a relaxation of ~\eqref{motion planning inclusive} that includes motion planning may be solved optimally and effectively. We add a constraint to ensure that the robot will travel through the recently added waypoint $\bm{p}_{add}$, avoiding the obstacle it collided with and preventing repeated collisions with it if $\bm{p}_{add} \neq \emptyset$.

\textit{1) Lower Bound of Time: } Limits on maximum speed, acceleration, jerk, etc. imposed by $\mathcal{S}^{obs}$ and $\mathcal{U}$ can help create a lower in \eqref{motion planning inclusive} of $T$. If $\bm{p}_{add} = \emptyset$, the minimum time to reach the nearest state $\bm{s}_{d, goal}$ in the goal region $\mathcal{S}^{goal}$ is constrained by $\underline{T}_{v} = \frac{\lVert \bm{p}_{goal} - \bm{p}_{0} \rVert_{\infty}}{v_{max}}$. This is because the system’s maximum velocity is bounded by $v_{max}$ along each axis.
The system’s maximum acceleration is bounded by $a_{max}$, hence the state $\bm{s}_{d,goal} = [\bm{p}_{goal}^{\top} \ \bm{v}_{goal}^{\top}]$ cannot be reached faster than
\vspace{-3pt}
\begin{subequations} \label{minimal time acceleration}
\begin{alignat}{2}
&\!\min_{\bm{a},\ \underline{T}_{a}}   &\qquad& 
   \underline{T}_{a}  
\label{eq:minimal_effort_mta}\\
&\text{subject to} &      &\dot{\bm{s}}_{d}(t)={A}_{f}\bm{s}_{d}(t) + {B}_{f}\bm{u}(t), \bm{u}(t) = \bm{a}(t),
\notag\\
&                   &      & {\forall} t \in [0, \underline{T}_{a}] 
\label{eq:mta_constraint1}\\
&                  &      &  \lVert \bm{a}(t) \rVert \leq a_{max}
\label{eq:mta_constraint2}\\
&                  &      &  \bm{s}_{d}(0) = [\bm{p}_{0}^{\top} \ \bm{v}_{0}^{\top}]^{\top}, \bm{s}(\underline{T}_{a}) = [\bm{p}_{goal}^{\top} \ \bm{v}_{goal}^{\top}]^{\top}
\label{eq:mta_constraint3}
\end{alignat}
\end{subequations}

The above is a minimum-time optimal control problem with input constraints, which can be solved in closed form along each individual axis to obtain the lower bound $\underline{T}_{a} = \min \{ \underline{T}_{a,x}, \underline{T}_{a,y}, \underline{T}_{a,z} \}$ \cite[ch.~5]{lewis2012optimal}. This procedure applies for constraints in higher-order derivatives, but in practice the computed times are less likely to provide better bounds while requiring higher computational effort. Hence, even though we can define a lower bound on the minimum achievable time via $\underline{T} = \min\{ \underline{T}_{v}, \underline{T}_{a}, \cdots \}$, for computational expediency we use the efficiently-computed (but less tight) bound $\underline{T} = \underline{T}_{v}$. For those cases with $\bm{p}_{add} \neq \emptyset$, we generate $\underline{T}_{1}$ and $\underline{T}_{2}$ for path segments $\bm{p}_{0} \rightarrow \bm{p}_{add}$ and $\bm{p}_{add} \rightarrow \bm{p}_{goal}$ as $\underline{T}_{1} = \underline{T}_{v, 1} = \frac{\lVert \bm{p}_{add} - \bm{p}_{0} \rVert_{\infty}}{v_{max}}$ and $\underline{T}_{2} = \underline{T}_{v, 2} \frac{\lVert \bm{p}_{goal} - \bm{p}_{add} \rVert_{\infty}}{v_{max}}$.

\textit{2) Velocity Control Linear Quadratic Minimum Time Heuristic: } 
The lower bound $\underline{T}$ can help relax~\eqref{motion planning inclusive} by replacing the state and input constraints. 
If ${p}_{add} = \emptyset$, then
\vspace{-3pt}
\begin{subequations} \label{motion planning LQMT}
\begin{alignat}{2}
&\!\min_{\bm{D}, T_{g}}   &\qquad& 
   J_{s}(\bm{D}) + \rho_{t} T_{g}  
\label{eq:minimal_effort_mpLQMT}\\
&\text{subject to} &      &\dot{\bm{s}}_{d}(t)={A}_{f}\bm{s}_{d}(t) + {B}_{f}\bm{u}(t), {\forall} t \in [0, T_{g}] 
\label{eq:mpLQMT_constraint1}\\
&                  &      &  \bm{s}_{d}(0) = \bm{s}_{d0}, \bm{s}_{d}(T_{g}) \in \mathcal{S}^{goal} 
\label{eq:mpLQMT_constraint2}\\
&                  &      &  T_{g} \geq \underline{T}\enspace.
\label{eq:mpLQMT_constraint3}
\end{alignat}
\end{subequations}
The relaxed problem \eqref{motion planning LQMT} is in fact the classical Linear Quadratic Minimum-Time Problem \cite{verriest1991linear}. The optimal cost generated from \eqref{motion planning LQMT} according to \cite{liu2017planning} is
\begin{equation} \label{heuristic function}
h(\bm{s}_{d,0}) = {\delta}_{T}^{\top} {W}_{T}^{-1} {\delta}_{T} + 
\rho_{t}{T_{g}}\enspace.
\end{equation}
We define ${\delta}_{T} = \bm{s}_{d,goal} - e^{{A}_{f}T_{g}} \bm{s}_{d,0}$ and the controllability Gramian ${W}_{T} = \int_{0}^{T_{g}}e^{{A}_{f}t}{B}_{f}^{\top}e^{{A}_{f}^{\top}t}{B}_{f}dt$.

Let us consider velocity control as an illustrative example of~\eqref{heuristic function}. Given $T_{g}$, $\bm{s}_{d,0} = \bm{p}_{0}$, $\bm{s}_{d,goal} = \bm{p}_{goal}$, we can rewrite the optimal cost of \eqref{motion planning LQMT} shown in \eqref{heuristic function} as
\begin{equation} \label{heuristic function v control 1}
h_{v}(\bm{s}_{d,0}) = C^{*}(T_{g}) = \frac{\lVert \bm{p}_{goal} - \bm{p}_{0} \rVert^{2}}{T_{g}} + \rho_{t} T_{g}\enspace.
\end{equation}
By minimizing $C^{*}$ in~\eqref{heuristic function v control 1} with the constraint $T^{*}_{g} \geq \underline{T}$, we are able to obtain the ideal $T^{*}_{g}$. If the positive real root $root^{+} \geq \underline{T}$, then the solution is the positive real root of $\frac{dC^{*}}{dT_{g}} = 0$.  Otherwise, $T^{*}_{g} = \underline{T}$. Furthermore, the optimal cost is $C^{*}(T^{*}_{g})$.
For the case where $\bm{p}_{add} \neq \emptyset$, we modify \eqref{heuristic function v control 1} to
\begin{equation} \label{heuristic function v control 2}
\begin{aligned}
& h_{v}(\bm{s}_{d,0}) = C^{*}(T_{1}, T_{2}) = \frac{\lVert \bm{p}_{add} - \bm{p}_{0} \rVert^{2}}{T_{1}} + \rho_{t} T_{1} + \\
& \frac{\lVert \bm{p}_{goal} - \bm{p}_{add} \rVert^{2}}{T_{2}} + \rho_{t} T_{2}
\end{aligned}
\end{equation}
Similarly, we are able to derive the optimal $T^{*}_{1}$ and $T^{*}_{2}$ by minimizing $C^{*}$ in \eqref{heuristic function v control 2} with constraints $T^{*}_{1} \geq \underline{T}_{1}$ and $T^{*}_{2} \geq \underline{T}_{2}$. We can get the solution of this optimization problem by solving the positive real root of $\frac{\partial C^{*}}{\partial T_{1}} = 0$ and $\frac{\partial C^{*}}{\partial T_{2}} = 0$. The optimal cost then is $C^{*}(T^{*}_{1}, T^{*}_{2})$. 

\subsection{Jump Point-based Computation to Improve Efficiency} \label{subsec:jump point}
Previous analyses (Sec.~\ref{subsec:problem formulation} to~\ref{subsec:heuristic}), yield the overall structure of our proposed collision-inclusive search-based motion planning algorithm, based on A* graph search. From \eqref{mpci serach based planning}, we notice that we extend the feasible set of the optimization problem compared to the collision avoidance planning problem~\cite{liu2017search}. Extending the feasible set forces our method to traverse more nodes on the graph compared to the collision avoidance method. Even though our method can generate a less conservative result with less cost compared to the collision avoidance method, the computational time of our can be larger compared to collision avoidance. 

To improve computational efficiency and reduce the computational time of our method, we can replace the post-impact motion primitive generation technique introduced in Sec.~\ref{subsec:collision check} in A* graph search with a more efficient variant that is inspired by jump point search~\cite{harabor2011online}. 
Specifically, we notice that when the robot needs to add a new waypoint between the collision point $\bm{p}_{e}^{-}$ of the motion primitive and the goal $\bm{p}_{goal}$ ($\xi = 2$), we can modify $\bm{p}_{e}^{+} = \bm{p}_{add}$. Performing this modification will help us eliminate traversing multiple nodes with the same $p_{add}$. This way, the number of nodes we are traversing can reduce, thus reducing computational time. Even though applying this technique can be at expense of optimality of the solution, solving the planning problem with less computational time can be more important in practice. 

If colliding with an obstacle (as shown in Fig.~\ref{fig:new waypoint generation}), we modify $\bm{p}_{e}^{+}$ and duration $\tau$ as $\bm{p}_{e}^{+} \leftarrow \bm{p}_{add}$ and $\tau \leftarrow \tau + \tau_{add}$ with $\tau_{add} = \frac{\lVert \bm{p}_{add, y} - \bm{p}_{e}^{-} \rVert}{\lVert {v}_{e}^{+} \rVert}$. The cost will be updated with new $\tau \leftarrow \tau + \tau_{add}$. 
When we go through edges with $\xi = 2$, we split the trajectory of this given edge with two segments, given the start and the end waypoints as $\bm{p}_{0}$ and $\bm{p}_{e}^{-}$ for the first segment, $\bm{p}_{e}^{-}$ and $\bm{p}_{e}^{+}$ for the second segment. The time duration of the first segment is $\tau - \tau_{add}$ and the time duration of the second segment is $\tau_{add}$. We set the $\xi = 0$ and $\zeta = 0$ with respect to the waypoint $\bm{p}_{e}^{+}$. 

\subsection{Trajectory Refinement}\label{subsec:refinement}
Following the aforementioned approach results in a collision-free trajectory including specific times needed to reach each waypoint. This is then fed as a prior to create smooth trajectories in higher dimensions. The refined trajectory $\bm{s}_{d}^{*}(t)$ is derived from solving a gradient-based trajectory generation problem similar to the one in Sec.~\ref{subsec:replanner formulation} with given initial and end states $\bm{s}_{d,0}$ and $\bm{s}_{d,goal}$ and intermediate waypoints $\bm{p}_{k}, k \in \{0, 1, \cdots, N\}$. The time for each trajectory segment $\tau_{k}$ is also given from the prior trajectory. All $\bm{p}_{k}$ are stored in $\mathbf{waypoint\_list}$ (see Alg.~\ref{pseudo replanner}). 

We apply a two-step optimization strategy similar to \cite{gao2017gradient} which can be summarized as follows: 1) First, optimize the collision cost of the path generated from waypoints only. Positions of intermediate waypoints on the initial path are left as free variables, and will be pushed away from the obstacles. 2) Second, revise the time scaling of the trajectory according to current waypoints’ positions. Then optimize the objective with additional smoothness and dynamical penalty terms.

\begin{figure}[!t]
\vspace{6pt}
\centering
\includegraphics[trim=0.5cm 3.5cm 1.0cm 8cm, clip,width=0.4\linewidth]{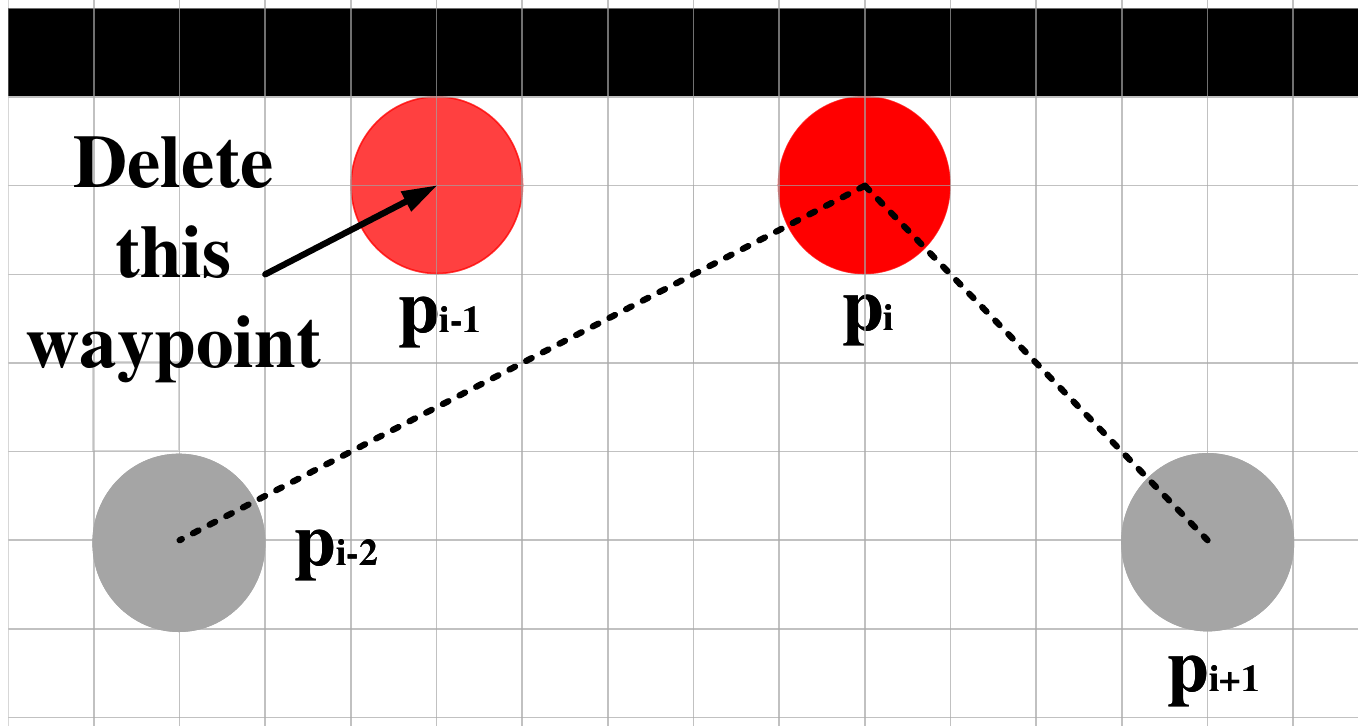}
\vspace{0pt}
\caption{Collision state merging.}%
\label{fig:Collision merging}
\vspace{-18pt}
\end{figure}

The output trajectory comprises $N_{c} + 1$ continuous polynomial trajectory segments. 
The differential variable of the waypoint with $\xi_{k} \geq 1$ is fixed end variable in the collision-inclusive method, which indicates $\bm{s}_{d,ed,i} = \bm{s}_{d,k}^{-}, k \in \{0, 1, \cdots, N\}, i \in \{1, \cdots, N_{c}\},\ \forall \xi_{k} \geq 1$. The differential variable of the next fixed initial state is $\bm{s}_{d,st,i+1} = \bm{s}_{d,k}^{+}, k \in \{0, 1, \cdots, N\}, i \in \{1, \cdots, N_{c}\},\ \forall \xi_{k} \geq 1$. 
The collision state waypoints $\xi_{k} \geq 1$ are generated from a grid map with augmented obstacles. We need to adjust those waypoints before trajectory generation by relocating them so that the distance to the closest block is $d_{c} \leq r_{rob}$; this way we ensure that planned collisions occur. 
The output of the search-based algorithm may have two colliding and reflecting states that are close-by (Fig.~\ref{fig:Collision merging}). In this case, if $\bm{p}_{i}$ is visible to both $\bm{p}_{i-2}$ and $\bm{p}_{i+1}$, we can delete $\bm{p}_{i-1}$ to reduce redundant collisions.
We also disregard obstacles that the robot planned to collide on ($\xi_{k} \geq 1$) when computing the potential field for trajectory generation in the second step for computational expediency. The trajectory after refinement is $n$-th order continuous. 

It is important to note that even though the refinement step produces a smoother trajectory, the refined course might be dynamically infeasible; we need to perform time scaling as in~\cite{liu2017planning} to reduce the maximum dynamics of the refined trajectory. The refined trajectory might collide with the obstacle in the trajectory segment that is checked to be collision free according to Sec.~\ref{subsec:collision check}. In such cases, our DRR strategy ensures robustness and safety. 

\setlength{\textfloatsep}{\textfloatsepsave}

\section{Experimental Results}\label{sec:experimental result}
We validate the effectiveness of our unified framework for collision-inclusive motion planning and control by presenting several benchmark testing results in simulation and via physical experimentation with our robot. 1) First, we test the deformation controller on the robot to evaluate its performance and generate a post-collision model which is required in simulation. 2) Second, we test the local DRR trajectory generation component experimentally with our robot. 3) Then, we test our global planning method in a double corridor environment and compare it with state-of-the-art search-based collision avoidance and sampling-based collision-inclusive methods. 4) After that we test the overall planning strategy in simulation in unknown, partially-observable environments. 5) Lastly, we evaluate the overall method with our impact-resilient robot experimentally. 

\subsection{Experimental Setup and Implementation Details}
Testing the deformation controller (Sec.~\ref{subsect:deformation controller}) and DRR (Sec.~\ref{subsec:drr_testing}) experimentally takes place in a $2.0\times2.5$\;m area with a rectangular pillar serving as a static polygon-shaped obstacle. 
The overall method is tested experimentally (Sec.~\ref{subsec:overall_testing}) in a $2.5\times3.5$\;m area with a long rectangular pillar in the middle to form a U-shaped corridor environment. 

We use the two active omni-directional impact-resilient wheeled robots we built in-house (Fig.~\ref{fig:robots}). The main chassis is connected to a deflection `ring' via $4$ or $8$ arms that feature a visco-elastic prismatic joint each. Each arm has embedded Hall effect sensors to measure the length of the arm and detect collisions along each of their direction when the deformation exceeds a certain threshold. 
In physical experiments, odometry feedback is provided by a 12-camera VICON motion capture system. An onboard Intel NUC mini PC ($2.3$\;GHz i7 CPU; $16$\;GB RAM) processes odometry data and sends control commands to the robot at $10$\;Hz. The robot is equipped with a single-beam LiDAR (RPLidar A2) with $8$\;m range to detect the obstacles in the environment. 

The robot may flip when colliding with velocities over a bound. To identify a theoretical collision velocity bound to avoid flipping, we use an energy conservation argument. Assume the kinetic energy before collision transfers into elastic potential energy of the arm, and the gravitational potential energy of the robot with small flipping angle counters the negative work input from the controller, i.e. 
\vspace{-10pt}

{\small
    {\begin{equation*}
        E_{k, t^{-}}(v_{max}) = E_{ep}(l_{e}) - E_{ep}(l_{s}) + E_{gp}(\sigma_{max}) + ma_{in, max}(l_{s} - l_{e})
    \end{equation*}
    }%
}

Then,

{\small{
\begin{equation*}
\begin{aligned}
    v_{max} = & \{\frac{k[(l_{e} - l_{0})^{2} - (l_{s}-l_{0})^{2}]}{2m} + g(\rho - l_{s} + l_{e})\sin{\sigma_{max}}\\
    + & a_{in, max}(l_{s}-l_{e})\}^{\frac{1}{2}} \enspace. 
\end{aligned} 
\end{equation*}
}%
}

The robot's radius is $0.3$\;m. The difference between the initial and neutral position of each arm is $l_{s} = 30.0$\;mm, the maximum load length is $l_{e} = 15.0$\;mm, and the neutral length is $l_{0} = 41.5$\;mm. The spring coefficient is $k = 2.31$\;N/mm. We set the largest flip angle $\sigma_{max} = \ang{3}$. For the 4-arm robot, the maximum acceleration input is $a_{in, max} = 5.0$\;m/s$^{2}$, and its mass is $6.0$\;kg. Then, we compute an upper theoretical velocity bound of $v_{max} \approx 0.7$\;m/s.\footnote{ The 8-arm robot features motors with higher torque and different gear ratio that increase $a_{in,max}$ and despite the mass increase to $8$\;kg, the same upper theoretical velocity bound remains valid.}

Simulated comparison against other methods (Sec.~\ref{subsec:global planner}) takes place in a double-corridor environment, whereas simulated benchmark testing of our method when noise is added takes place in the same double-corridor environment but with added isolated obstacles added as well (Sec.~\ref{subsec:simulation framework}). We further consider a similar environment that features non-convex obstacles (Sec.~\ref{subsec:simulation framework}).

We use a rigid cylinder body to emulate the robot. A numerical model is generated from the experiments for the deformation recovery controller to determine the output velocity after collision. The output velocity is generated by adding uniform random noise to the reference velocity $~^{c}{v}_{T}$. Then, we use a ray-casting algorithm to emulate the LiDAR (we consider the range of the LiDAR can cover all visible operating space). We implement simulation benchmarks in a python environment. All simulations run on an workstation with Intel Core Xeon-E2146G CPU.

\subsection{Experimental Testing of the Deformation Controller}\label{subsect:deformation controller}
To examine the deformation controller's effect in local trajectory generation, we command the robot to collide with an obstacle and then apply the proposed deformation recovery controller. 
We perform $10$ trials of various input-output velocity combinations~\cite[Table 2]{lu2021deformation}. Collision detection is very accurate; only $9$ out of $249$ were not detected. 

Results suggest that the deformation controller generates a negative velocity to make the robot detach from the obstacle after collision. Actual output velocity $\bar{{v}}_{out}$ is determined by the actual input velocity $\bar{{v}}_{in}$ and the set output value ${v}_{out,set}$ though the latter may not be reached in practice. That is because feedback linearization is not robust to system parameter uncertainties that occur in practice.
We observe that the velocity along the normal to collision direction is closer to the set velocity than the velocity along the tangent direction. This is because most of the uncertainties in system parameters enter as unmodeled friction dynamics along the tangent direction. Further, the sensor is more accurate when the input velocity is along the normal direction; the average value of deformation detected in this case is $29\%$ larger. 

\vspace{0pt}
\subsection{Experimental Testing of the DRR Strategy}\label{subsec:drr_testing}

We test our DRR strategy with a trajectory generated based on using the online safe trajectory generation method in~\cite{richter2016polynomial} with time scaling as in~\cite{liu2017planning} without collision checking. We compare the strategies in two cases: 1) when the previous path does not intersect with the collision surface; and 2) when the previous path intersects with the collision surface.

\begin{figure}[!h]
\vspace{-6pt}
      \centering
      \begin{subfigure}{0.225\textwidth}
        \includegraphics[trim={6cm 0.2cm 6cm 6.25cm}, clip, width=0.925\textwidth]{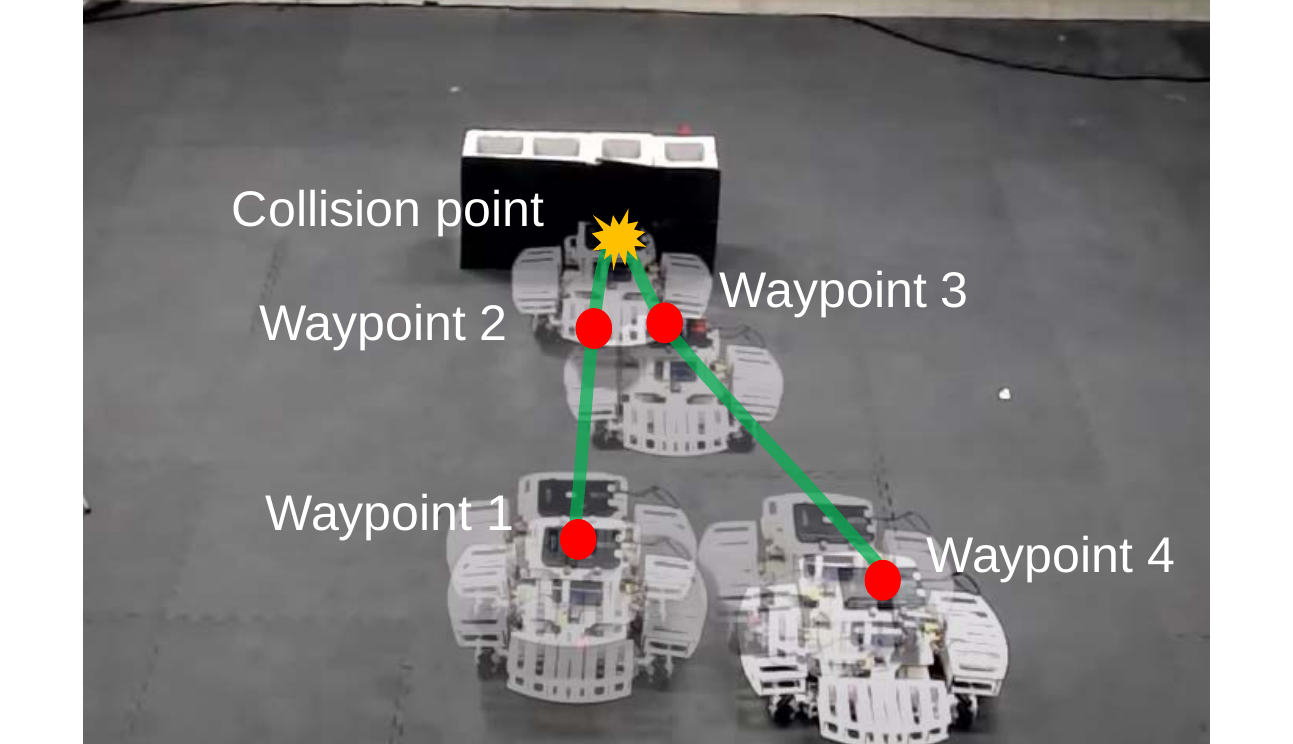}
      \end{subfigure}
      \begin{subfigure}{0.225\textwidth}
        \includegraphics[trim={6cm 0.2cm 6cm 6.25cm}, clip, width=0.925\textwidth]{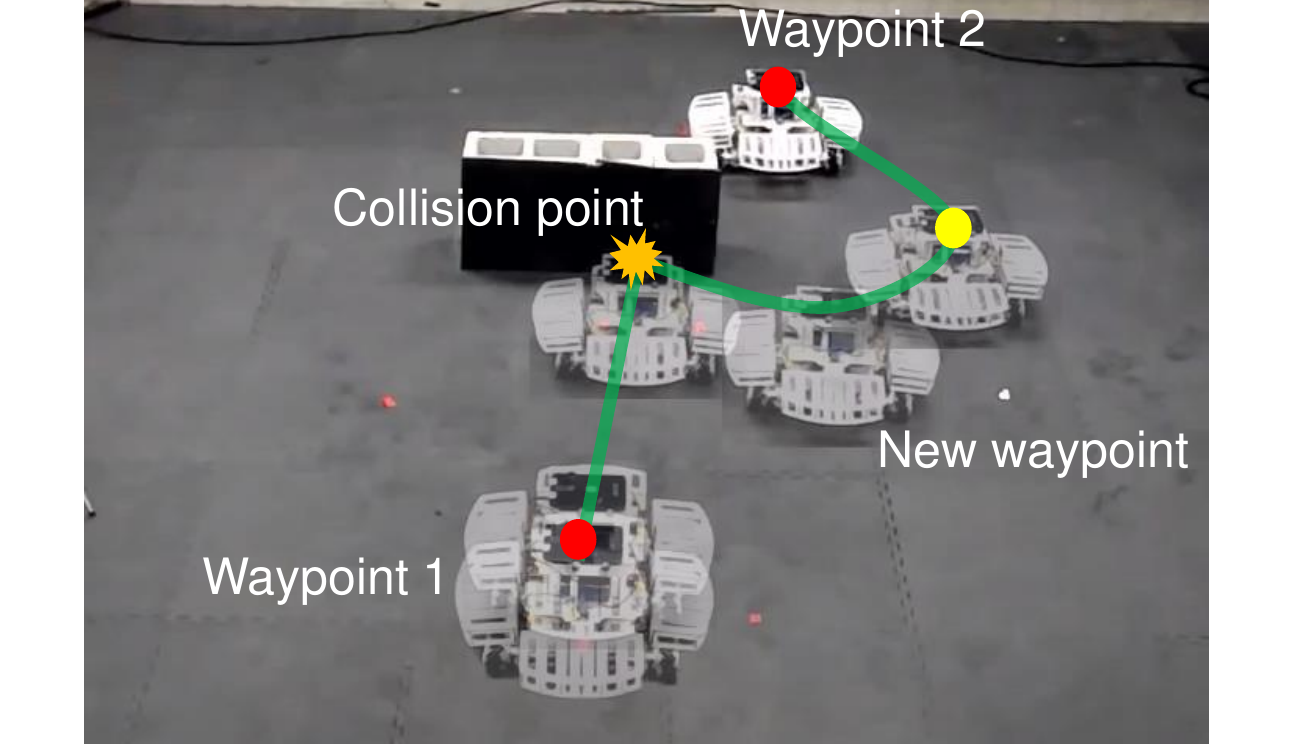}
      \end{subfigure}
      \caption{Composite images (taken every $2$\;s) of a sample experiment with DRR. The robot must go from start to goal passing through all waypoints, including intermediate ones created post-collision. 
      } 
      \label{fig:ExperimentDRR}
	\vspace{-9pt}
\end{figure}

\begin{figure}[!t]
\vspace{6pt}
 \begin{subfigure}{.235\textwidth}
  \centering
  \includegraphics[trim={4cm 0.2cm 4cm 0.2cm}, clip, width=0.75\linewidth]{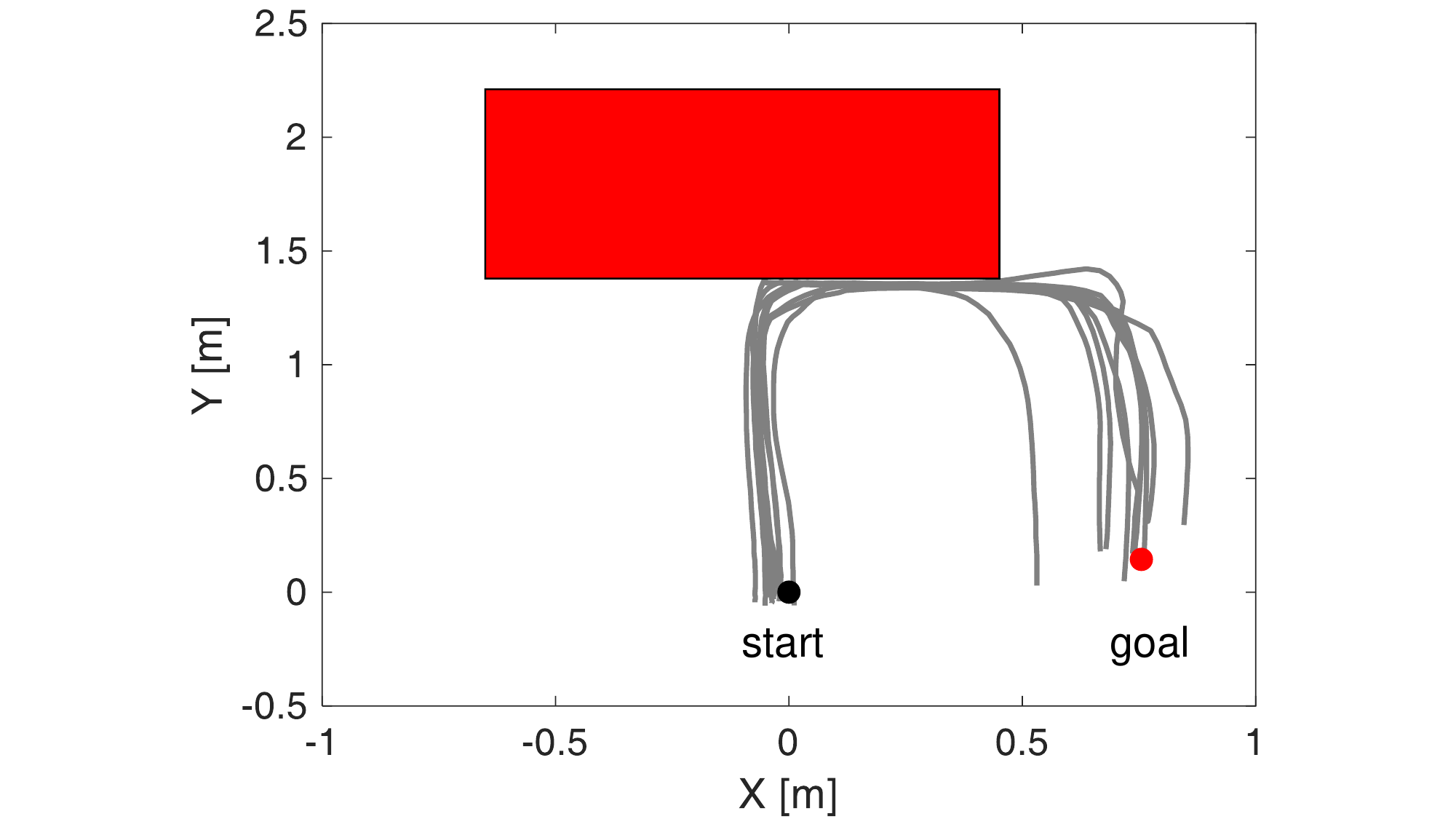}
  \vspace{-3pt}
  \caption{Case 1 collision avoidance.}
 \end{subfigure}%
 \begin{subfigure}{.235\textwidth}
  \centering
  \includegraphics[trim={4cm 0.2cm 4cm 0.2cm}, clip,width=0.75\linewidth]{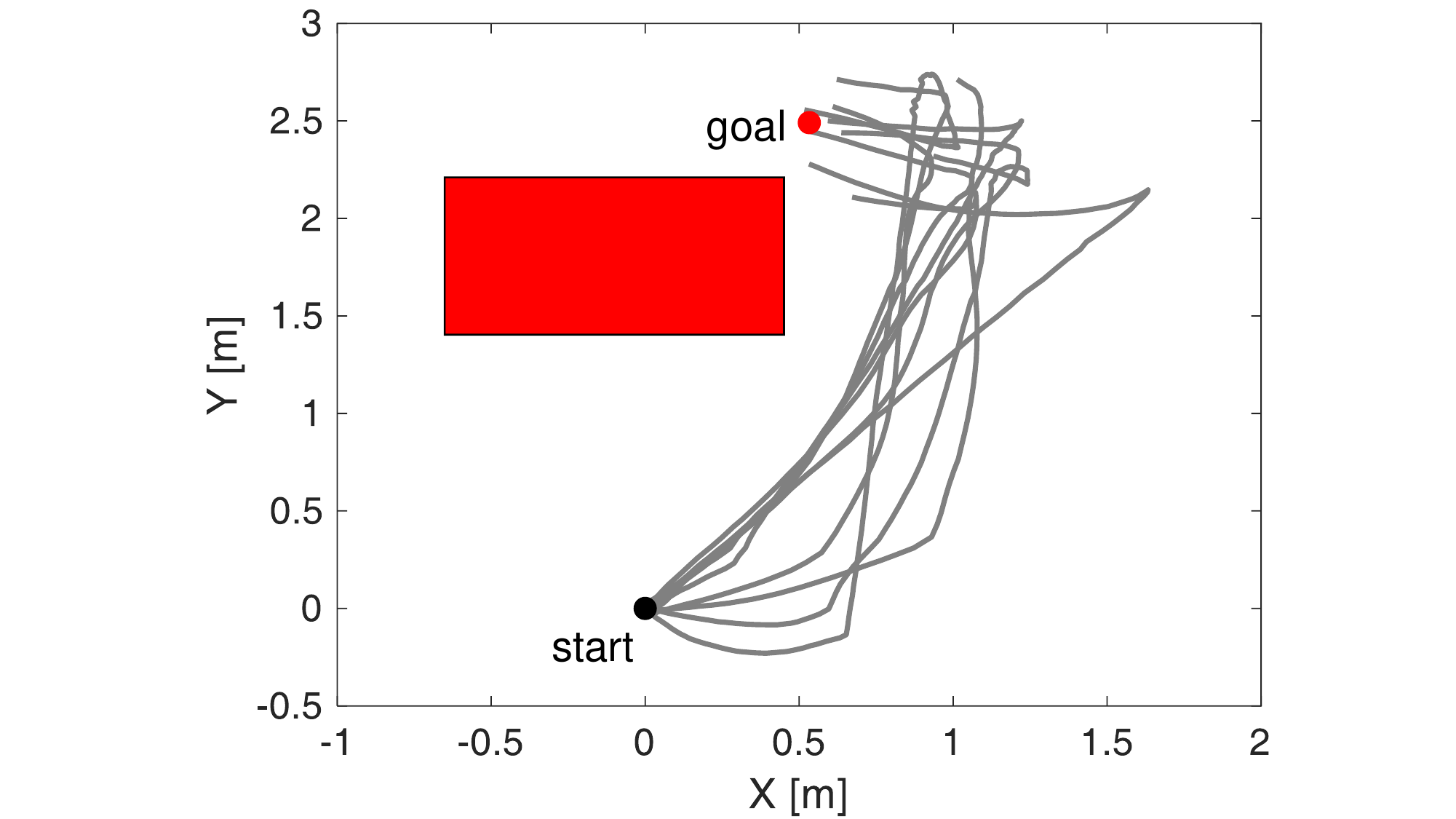}
  \vspace{-3pt}
  \caption{Case 2 collision avoidance.}
 \end{subfigure}
 \vspace{-1pt}
 \medskip
 \begin{subfigure}{.235\textwidth}
  \centering
  \includegraphics[trim={4cm 0.2cm 4cm 0.2cm}, clip,width=0.75\linewidth]{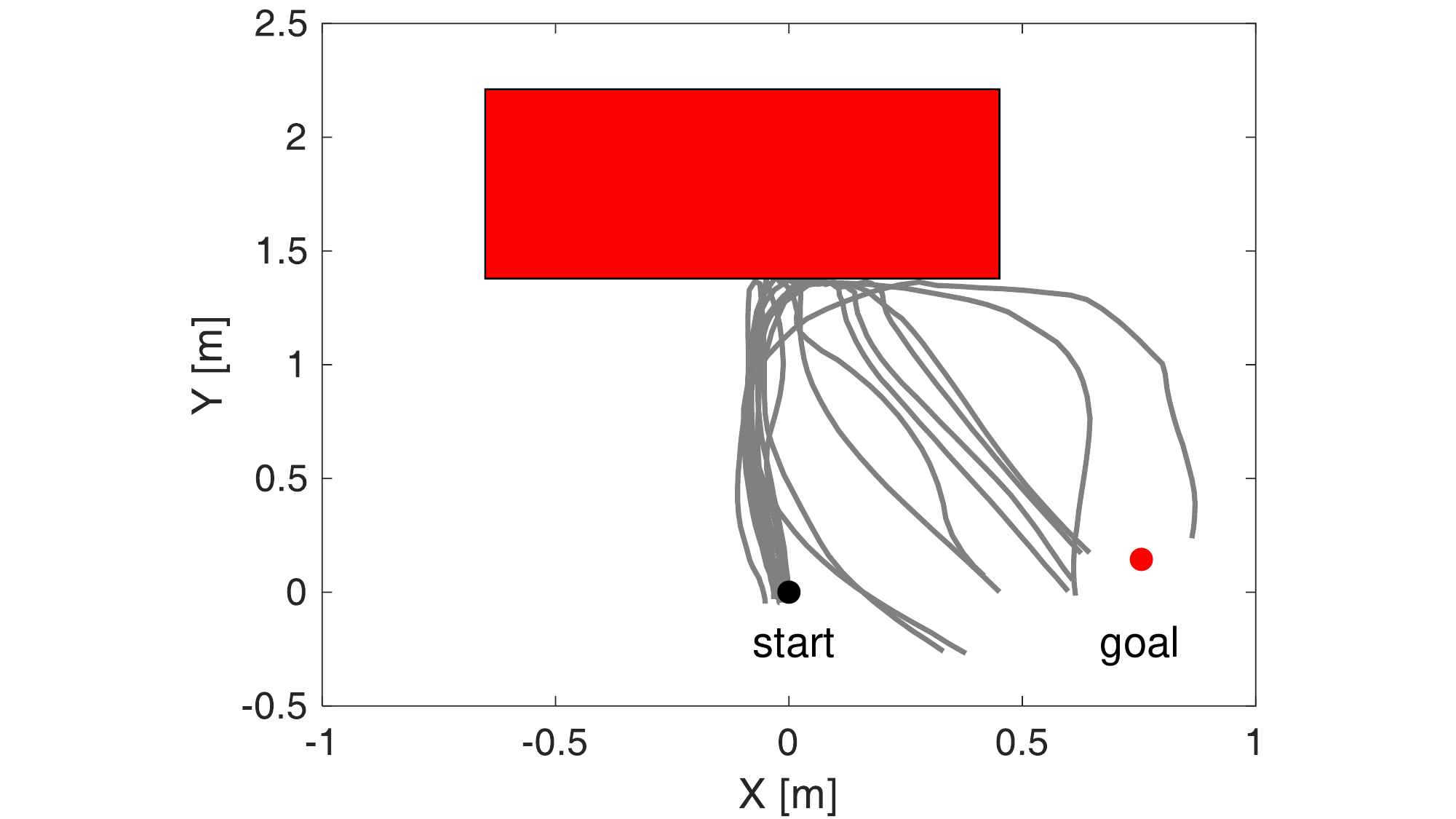}
  \vspace{-3pt}
  \caption{Case 1 DRR.}
 \end{subfigure}%
 \begin{subfigure}{.235\textwidth}
  \centering
  \includegraphics[trim={4cm 0.2cm 4cm 0.2cm}, clip,width=0.75\linewidth]{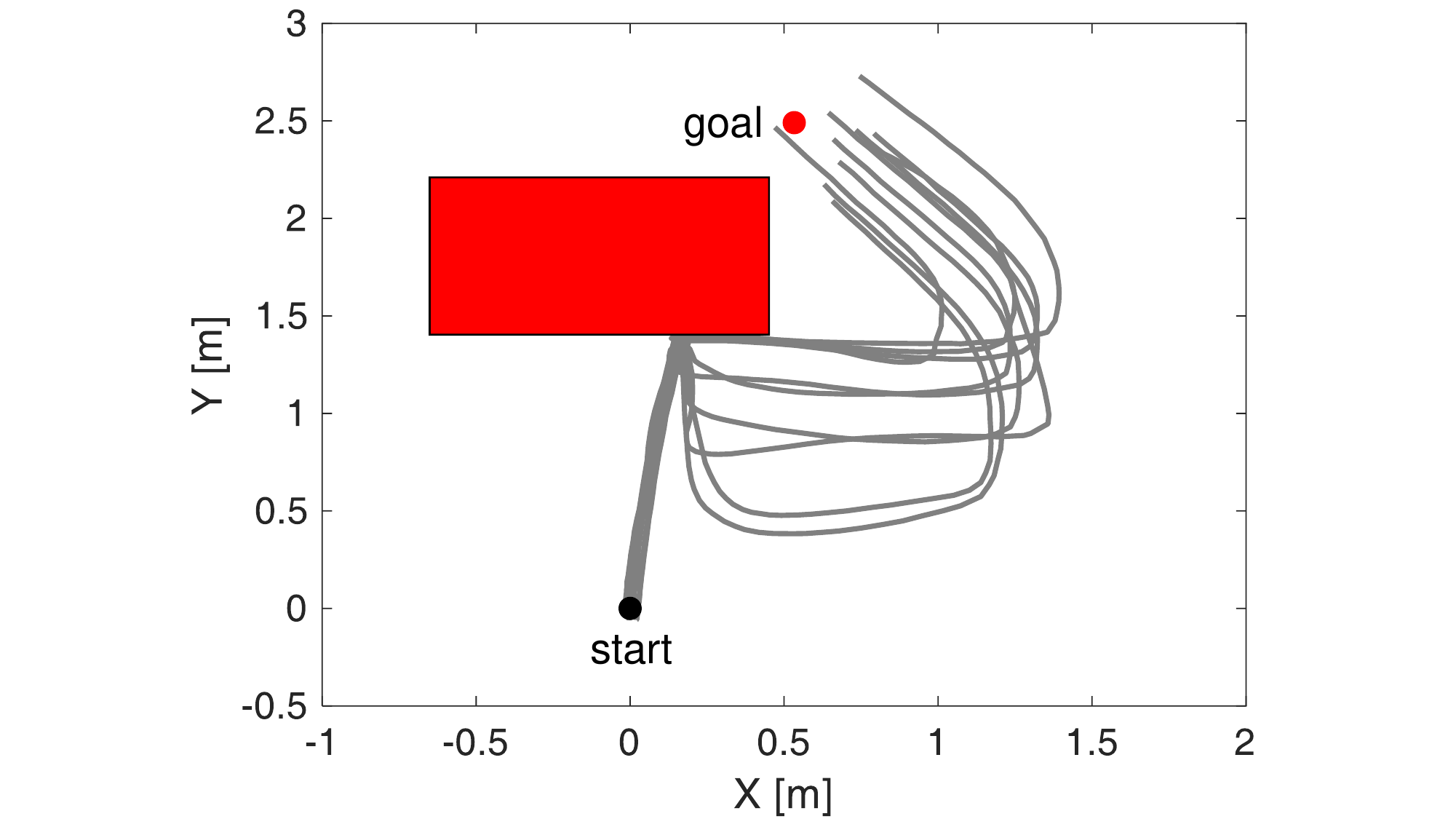}
  \vspace{-3pt}
  \caption{Case 2 DRR.}
 \end{subfigure}%
 \vspace{-6pt}
 \caption{Experimental trajectories generated from DRR and collision avoidance \cite{richter2016polynomial} when the preplanned path intersects or does not intersect with the obstacle. (In all cases we conduct 10 trials).} 
 \label{fig:compare DRR and avoidance}
 \vspace{-9pt}
\end{figure}

Case 1 tests the condition in Fig.~\ref{fig:no adjustment}, i.e. no waypoint is added as per Alg.~\ref{pseudo replanner}. Case 2 tests Fig.~\ref{fig:cases of waypoint adjustment}, i.e. a waypoint is added to the list. In case 2, we run RRT* to generate a collision free path and perform path simplification to remove nodes without affecting the path's collision safety~\cite[Fig.~4]{lu2021deformation}. The path simplification technique removes intermediate waypoints between two waypoints if a line segment between those two does not intersect with the obstacle. Then, we use the trajectory generation strategy in~\cite{richter2016polynomial}. We perform $10$ trials for each case. Instances of DRR and all experimental trajectories are shown in Fig.~\ref{fig:ExperimentDRR} and Fig.~\ref{fig:compare DRR and avoidance}, respectively.

Even though we design a collision-free desired trajectory with the strategy in \cite{richter2016polynomial}, the robot may still collide with the environment given for instance unmodeled dynamics such as drift. In case 1 there are $3$ out of $10$ trials that the robot in fact collides with the obstacle applying trajectory generation~\cite{richter2016polynomial} that aims to avoid collisions.  Table~\ref{table:compare} shows statistics on mean arrival times, path lengths and control energy.

\begin{table}[!h]
\vspace{3pt}
    \caption{Comparison of trajectory generation strategy in \cite{richter2016polynomial} (Collision-avoidance) and DRR (Collision-inclusive) strategies.}
    \vspace{-6pt}
    \label{table:compare}
    \begin{center}
    \begin{tabular}{c c c c c}
    \toprule
    & \multicolumn{2}{c}{Strategy in \cite{richter2016polynomial}} & \multicolumn{2}{c}{DRR (our method)}\\
    \toprule
    & Case 1 & Case 2 & Case 1 & Case 2\\
    \toprule
    $\bar{T}_{end}$ [$s$] 
    & $7.7$
    & $8.71$
    & $6.16$
    & $9.17$\\
    \midrule
    $STD(T_{end})$ 
    & $0$
    & $2.20$
    & $0.22$
    & $0.31$\\
    \midrule
    $\bar{s}$ [$m$]
    & $3.153$
    & $3.448$
    & $2.977$
    & $4.38$\\
    \midrule
    $STD(s)$ 
    & $0.117$
    & $0.495$
    & $0.150$
    & $0.451$\\
    \midrule
    $\bar{E_{c}}$ [$m^{2}/s^{3}$]
    & $56.83$
    & $80.62$
    & $58.42$
    & $255.96$\\
    \midrule
    $STD(E_{c})$ 
    & $29.34$
    & $54.38$
    & $32.92$
    & $133.54$\\
    \bottomrule
    \end{tabular}
    \end{center}
    \vspace{-18pt}
\end{table}

In case $1$ for DRR, mean arrival times $\bar{T}_{end}$ and path lengths $\bar{s}$ decrease by $25\%$ and $6\%$, while the control energy increases by $2.8\%$ on average. However, the error in the end point increases by $25\%$. In case $2$, mean arrival times and path lengths increase by $5.2\%$ and $27\%$, and control energy increases by $258\%$. 
This is because the output velocity of DRR is not flat since the robot decelerates and then accelerates during boundary following. The path generated by the boundary following is not the shortest. %
However, since the path between the collision point and the new inserted waypoint is close to the obstacle surface, the existence of the obstacle decreases the control error in free space. The error in the end point decreases by $12\%$. These results show the tradeoff between online reactive execution (whereby collision checking is skipped) and collision avoidance.

\subsection{Simulated Tests of the Collision-inclusive Global Planner}\label{subsec:global planner}
\begin{figure}[!t]
\vspace{4pt}
\centering
     \begin{subfigure}{.235\textwidth}
      \centering
      \includegraphics[trim={0.5cm 0.2cm 1.0cm 1.0cm}, clip, width=0.9\linewidth]{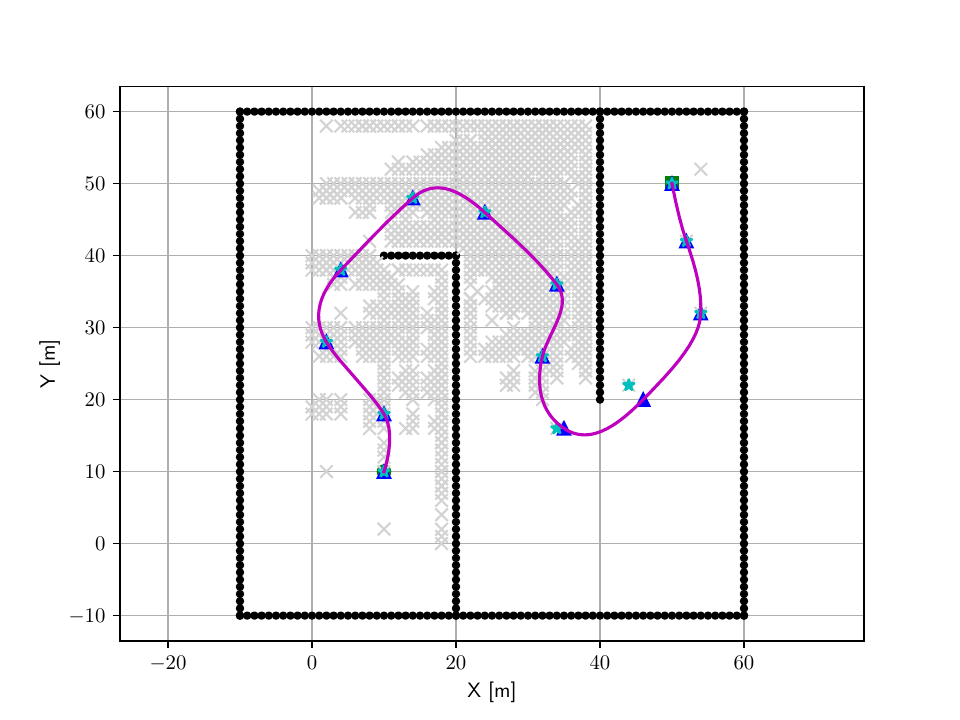}
      \vspace{-3pt}
      \caption{Collision-avoidance without\\ node pruning.}
      \label{fig:avoidance without}
     \end{subfigure}%
     \begin{subfigure}{.235\textwidth}
     \centering
      \includegraphics[trim={0.5cm 0.2cm 1.0cm 1.0cm}, clip,width=0.9\linewidth]{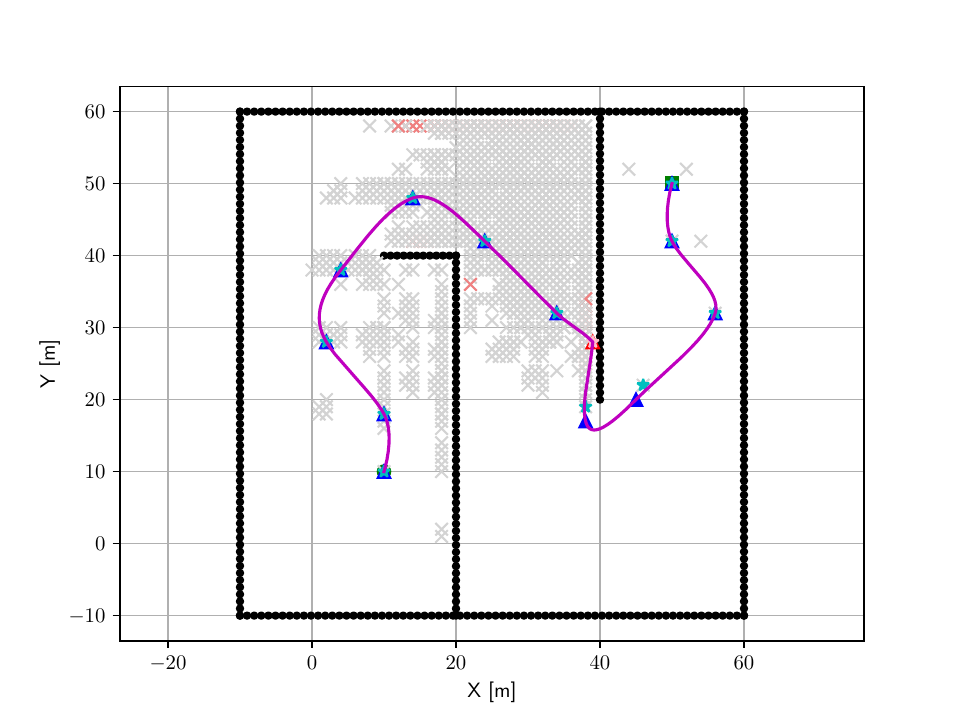}
      \vspace{-3pt}
      \caption{Collision-inclusive without\\ jump points ($\rho_{c} = 1.0$)}
      \label{fig:avoidance with}
     \end{subfigure}
     \vspace{0pt}
     
     \begin{subfigure}{.235\textwidth}
      \centering
      \includegraphics[trim={0.5cm 0.2cm 1.0cm 1.0cm}, clip,width=0.9\linewidth]{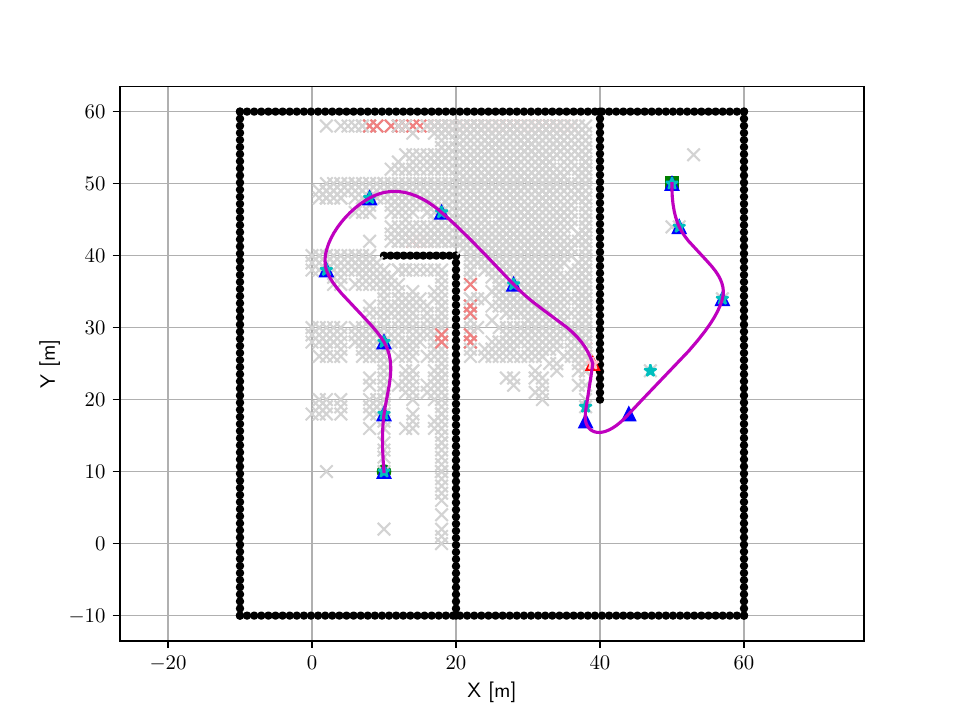}
      \vspace{-3pt}
      \caption{Collision-inclusive without\\ jump points ($\rho_{c} = 10.0$).}
      \label{fig:inclusive without}
     \end{subfigure}%
     \begin{subfigure}{.235\textwidth}
      \centering
      \includegraphics[trim={0.5cm 0.2cm 1.0cm 1.0cm}, clip,width=0.9\linewidth]{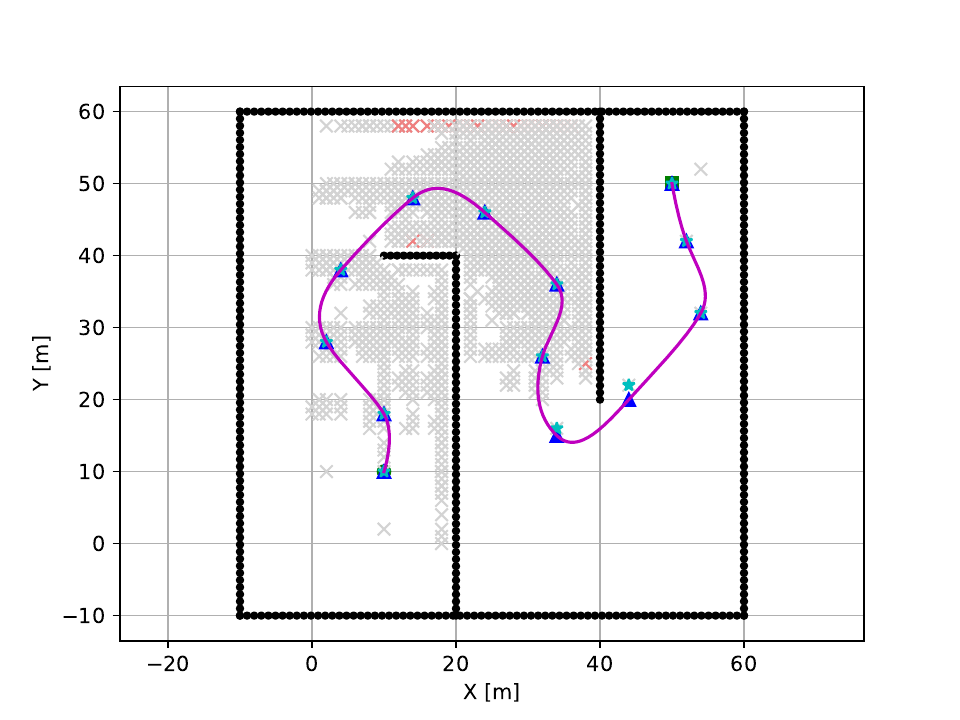}
      \vspace{-3pt}
      \caption{Collision-inclusive without\\ jump points ($\rho_{c} = 100.0$).}
      \label{fig:inclusive with}
     \end{subfigure}%
     \vspace{0pt}
     
     \begin{subfigure}{.235\textwidth}
      \centering
      \includegraphics[trim={0.5cm 0.2cm 1.0cm 1.0cm}, clip,width=0.9\linewidth]{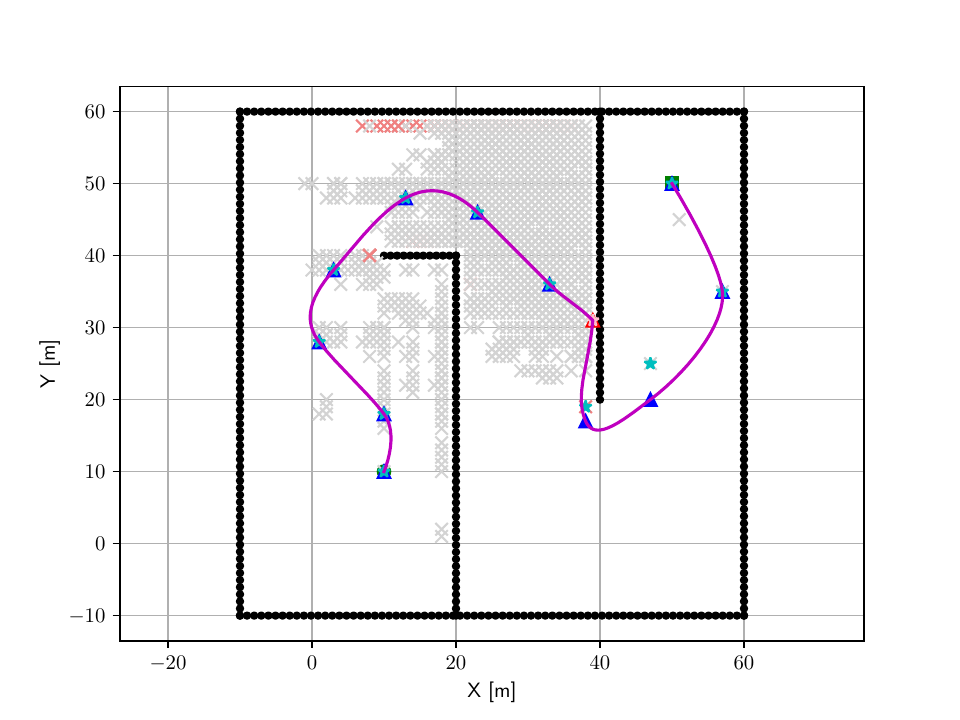}
      \vspace{-3pt}
      \caption{Collision-inclusive with\\ jump points ($\rho_{c} = 1.0$).}
      \label{fig:inclusive without 100}
     \end{subfigure}%
     \begin{subfigure}{.235\textwidth}
      \centering
      \includegraphics[trim={0.5cm 0.2cm 1.0cm 1.0cm}, clip,width=0.9\linewidth]{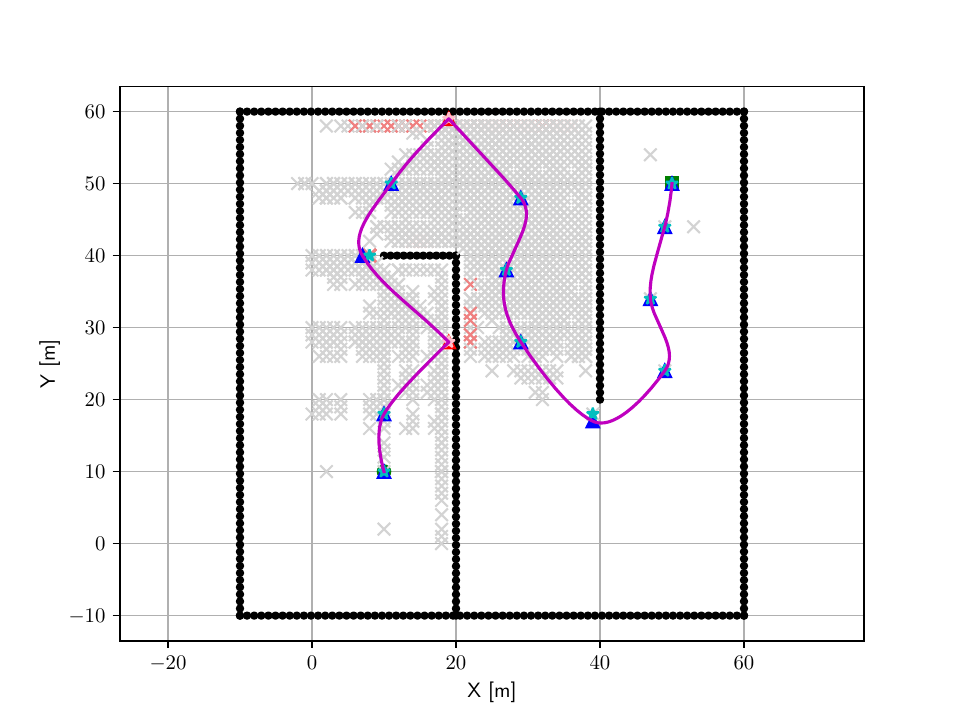}
      \vspace{-3pt}
      \caption{Collision-inclusive with\\ jump points ($\rho_{c} = 10.0$).}
      \label{fig:inclusive with 1000}
     \end{subfigure}%
 \vspace{0pt}
 \caption{Search-based collision avoidance global planner~\cite{liu2017search} (top panels) and our proposed search-based collision-inclusive global planner for three different $\rho_c$ values. Higher $\rho_c$ values (bottom panels) penalize collisions more, thus recovering behaviors that resemble collision avoidance. The collision-inclusive with jump points case when $\rho_c=100.0$ is very similar to the case without jump points in panel (d) in the sense of number of closed nodes $N_p$ (c.f. Table~\ref{table:globalplanner}, and hence not shown here for brevity.}
 \label{fig:globalplannertest}
 \vspace{-21pt}
\end{figure}

To test our proposed framework in simulation, we first benchmark it in a double corridor environment to test our search-based collision-inclusive global planner. We compare our method for global planning against two methods: 1) a search-based collision-avoidance motion planning algorithm~\cite{liu2017search}, and 2) an RRT*-based (sampling-based) collision-inclusive planning algorithm adapted from~\cite{zha2021exploiting} to ensure fair comparisons.\footnote{~No open-source python code is available for either~\cite{liu2017search} and~\cite{zha2021exploiting}, so we implemented both ourselves to the best possible extent.} 
In all tests, the dynamic limits are set as $a_{max} = 5.0\ {\rm m/s^{2}}$. The holonomic robot only translates but does not rotate during the process.\footnote{~Constant orientation is maintained via a separate stabilizing controller.} We set the upper bound of the robot velocity $v_{max} = 2.0$\;m/s. The cost function in all methods considers $\rho_{t}=1.0$. The overall size of the map is $70\times70$\;m. The position resolution of the grid map in the benchmark is $1.0$\;m, and the position resolution of the velocity map in the benchmark is $0.1$\;m/s. The time interval for each motion primitive is set as $\tau = 5.0$\;s (selected via an ablation study the results of which are shown in Table~\ref{table:globalplannerprune}) and the resolution of acceleration $r = 1.0\ {\rm m/s^{2}}$ also selected via an ablation study the results of which are shown in Table~\ref{table:globalplannerres}). We set $\lambda_{s} = 0.5$, $\lambda_{o} = 1.0$ and $\lambda_{v} = 10.0$.\footnote{~The values were selected empirically to improve trajectory refinement.} Parameter $\rho_c$ in the cost function is one of the most critical ones as it determines how much collisions are being penalized. We select $\rho_c=1.0$, with the ablation studies to determine this value being demonstrated qualitatively in Fig.~\ref{fig:globalplannertest} and expanded in more depth in Table~\ref{table:globalplanner}.

Results from testing the global planner are shown in Fig.~\ref{fig:globalplannertest}; both collision avoidance as per~\cite{liu2017search} (panel a) and collision-inclusive (our method, panel b-f) results are highlighted. We demonstrate our method's results with and without implementing jump points. We also consider three cases for varying values of parameter $\rho_c$ which affects how much collisions are being penalized in the cost function: 1) $\rho_c=1.0$ (panels b and e) which corresponds to minimal penalty; 2) $\rho_c=10.0$ (panels c and f) which corresponds to a medium penalty; and 3) $\rho_c=100.0$ (panels d) which corresponds to a severe penalty. It can be readily verified that both cases of $\rho_c=\{1.0,10.0\}$ can lead to paths that contain collisions, although in some cases (especially when jump points are considered) a higher $\rho_c$ value of $10.0$ may make the output trajectory unnecessarily complex and suboptimal (panel f). As such, if collisions are to be considered, setting $\rho_c=1.0$ should be preferred. At very high $\rho_c$ values (of $100.0$), we observe that our method can recover collision avoidance behaviors (c.f. panels a and d). This highlights our global planner's ability to switch between collision-inclusive and collision avoidance planning on-demand by only updating the value of a single parameter.

\begin{table}[!h]
\vspace{-6pt}
    \caption{Comparison of global planners' performance between our method and that to prune primitives.}
    \vspace{-6pt}
    \label{table:globalplannerprune}
    \begin{center}
    \resizebox{0.45\textwidth}{!}{
    \begin{tabular}{c c c c c c c c}
    \toprule
    \multicolumn{2}{c}{$r = 1.0$} & {$\tau$} & Comp. & \multirow{2}*{\shortstack{$N_{p}$}} & Traj. & Ctrl.\\
    \multicolumn{2}{c}{[$\rm m/s^{2}$]} & [s] & Time[s] &  & Time[s] & Cost[${\rm m^{2}/s^{3}}$]]\\
    \toprule
    \multirow{2}*{\shortstack{method \cite{liu2017search} \\ without pruning}} &  & \multirow{2}*{$5.0$} & \multirow{2}*{$31.86$} & \multirow{2}*{$1427$} & \multirow{2}*{$71.5$} & \multirow{2}*{$12.01$}\\
    \\
    \midrule
    \multirow{3}*{\shortstack{Our method \\ without pruning\\ $\rho_{c} = 1.0$}} &  & \multirow{3}*{$5.0$} & \multirow{3}*{$34.60$} & \multirow{3}*{$1340$} & \multirow{3}*{${69.1}$} & \multirow{3}*{$13.25$} \\
    \\
    \\
    \midrule
    \multirow{2}*{\shortstack{method \cite{liu2017search} \\ with pruning}} &  & \multirow{2}*{$0.5$} & \multirow{2}*{$18629.12$} & \multirow{2}*{$102836$} & \multirow{2}*{$91.1$} & \multirow{2}*{$201.98$}\\
    \\
    \midrule
    \multirow{3}*{\shortstack{Our method \\ with pruning\\ $\rho_{c} = 1.0$}} &  & \multirow{3}*{$0.5$} & \multirow{3}*{$15168.97$} & \multirow{3}*{$89450$} & \multirow{3}*{${90.9}$} & \multirow{3}*{$200.72$} \\
    \\
    \\
    \bottomrule
    \end{tabular}
    }
    \end{center}
    \vspace{-15pt}
\end{table}

We also demonstrate the utility of formulating the motion primitives as discussed in Sec.~\ref{subsec:motion primitives} as compared to directly pruning dynamically infeasible primitives. Our proposed method can feature primitives of longer duration $\tau$, which in fact helps increase the efficiency of exploring the space. Table~\ref{table:globalplannerprune} provides comparative numbers for both collision avoidance~\cite{liu2017search} and our collision-inclusive method. Results verify that our selected primitives generation method can explore the space with significantly less computational time when compared to the approach of pruning infeasible states. 

Table~\ref{table:globalplannerres} contains the results from the ablation study on the resolution parameter $r$. We found that in the environment with simple obstacles (as in Fig.~\ref{fig:strategy simulation m1 5} shown later), a high resolution of $r = 0.5$ leads to computational times for both collision-avoidance and collision-inclusive methods that are much higher since the graph is denser. The computational time of the collision-inclusive method is higher than the collision-avoidance method since we modify those primitives that intersect with the obstacles instead of pruning them altogether. With lower resolution $r = 1.0$ or $r = 2.0$, the computational time of both collision-avoidance and collision-inclusive methods rapidly decreases. When $r = 1.0$ collision-avoidance and our collision-inclusive method have comparable computational performance. However, as we further increase the resolution ($r = 2.0$), the computational time of our collision-inclusive method gets much lower than the collision-avoidance method, which appears to be affected less by this change. These observations suggest that with lower resolution, collision-inclusive primitives can explore the space with higher efficiency. However, the trajectory time and control cost are higher than applying higher resolution. Similar observations can be made when testing in a more complicated environment (Fig.~\ref{fig:strategy simulation m2 5}). Therefore, taken computational time, trajectory time and control cost into consideration, we select $r = 1.0$ for which both collision-avoidance and collision-inclusive method have better results.

\begin{table}[!h]
\vspace{6pt}
    \caption{Comparison of global planners' performance for different values of parameter $r$}.
    \vspace{-6pt}
    \label{table:globalplannerres}
    \begin{center}
    \resizebox{0.45\textwidth}{!}{
    \begin{tabular}{c c c c c c c}
    \toprule
    non-convex env. & $r$ & Comp. & \multirow{2}*{\shortstack{$N_{p}$}} & Traj. & Ctrl.\\
    Fig.~\ref{fig:strategy simulation m1 5}& [${\rm m/s^{2}}$] & Time[s] &  & Time[s] & Cost[${\rm m^{2}/s^{3}}$]\\
    \toprule
    \multirow{2}*{\shortstack{method \cite{liu2017search}}} &  \multirow{2}*{$0.5$} & \multirow{2}*{$480.19$} & \multirow{2}*{$6752$} & \multirow{2}*{$63.1$} & \multirow{2}*{$11.86$}\\
    \\
    \midrule
    \multirow{3}*{\shortstack{Our method \\ without jump point\\ $\rho_{c} = 1.0$}} & \multirow{3}*{$0.5$} & \multirow{3}*{$873.78$} & \multirow{3}*{$9311$} & \multirow{3}*{${66.3}$} & \multirow{3}*{$14.09$} \\
    \\
    \\
    \midrule
    \multirow{2}*{\shortstack{method \cite{liu2017search}}} &  \multirow{2}*{$1.0$} & \multirow{2}*{$31.86$} & \multirow{2}*{$1427$} & \multirow{2}*{$71.5$} & \multirow{2}*{$12.01$}\\
    \\
    \midrule
    \multirow{3}*{\shortstack{Our method \\ without jump point\\ $\rho_{c} = 1.0$}} & \multirow{3}*{$1.0$} & \multirow{3}*{$34.60$} & \multirow{3}*{$1340$} & \multirow{3}*{${69.1}$} & \multirow{3}*{$13.25$} \\
    \\
    \\
    \midrule
    \multirow{2}*{\shortstack{method \cite{liu2017search}}} &  \multirow{2}*{$2.0$} & \multirow{2}*{$34.68$} & \multirow{2}*{$1657$} & \multirow{2}*{$74.6$} & \multirow{2}*{$17.48$}\\
    \\
    \midrule
    \multirow{3}*{\shortstack{Our method \\ without jump point\\ $\rho_{c} = 1.0$}} & \multirow{3}*{$2.0$} & \multirow{3}*{$9.70$} & \multirow{3}*{$557$} & \multirow{3}*{${74.7}$} & \multirow{3}*{$12.57$} \\
    \\
    \\
    \bottomrule
    \toprule
    non-convex env. &  &  &  & \\
    Fig.~\ref{fig:strategy simulation m2 5} &  &  &  & \\
    \toprule
    \multirow{2}*{\shortstack{method \cite{liu2017search}}} &  \multirow{2}*{$0.5$} & \multirow{2}*{$564.95$} & \multirow{2}*{$8385$} & \multirow{2}*{$74.9$} & \multirow{2}*{$9.58$}\\
    \\
    \midrule
    \multirow{3}*{\shortstack{Our method \\ without jump point\\ $\rho_{c} = 1.0$}} & \multirow{3}*{$0.5$} & \multirow{3}*{$1995.16$} & \multirow{3}*{$15908$} & \multirow{3}*{${81.3}$} & \multirow{3}*{$11.94$} \\
    \\
    \\
    \midrule
    \multirow{2}*{\shortstack{method \cite{liu2017search}}} &  \multirow{2}*{$1.0$} & \multirow{2}*{$55.11$} & \multirow{2}*{$2388$} & \multirow{2}*{$79.9$} & \multirow{2}*{$12.43$}\\
    \\
    \midrule
    \multirow{3}*{\shortstack{Our method \\ without jump point\\ $\rho_{c} = 1.0$}} & \multirow{3}*{$1.0$} & \multirow{3}*{$52.69$} & \multirow{3}*{$1799$} & \multirow{3}*{${76.6}$} & \multirow{3}*{$20.15$} \\
    \\
    \\
    \midrule
    \multirow{2}*{\shortstack{method \cite{liu2017search}}} &  \multirow{2}*{$2.0$} & \multirow{2}*{$48.33$} & \multirow{2}*{$2151$} & \multirow{2}*{$89.0$} & \multirow{2}*{$20.84$}\\
    \\
    \midrule
    \multirow{3}*{\shortstack{Our method \\ without jump point\\ $\rho_{c} = 1.0$}} & \multirow{3}*{$2.0$} & \multirow{3}*{$24.39$} & \multirow{3}*{$1726$} & \multirow{3}*{${79.6}$} & \multirow{3}*{$12.33$} \\
    \\
    \\
    \bottomrule
    \end{tabular}
    }
    \end{center}
     \vspace{-21pt}
\end{table}

\begin{table}[!h]
\vspace{6pt}
    \caption{Comparison of global planners' performance for different values of parameter $\rho_c$.}
    \label{table:globalplanner}
    \vspace{-6pt}
    \begin{center}
    \resizebox{0.45\textwidth}{!}{
    \begin{tabular}{c c c c c c c c}
    \toprule
    \multicolumn{2}{c}{${v}_{0} = [0\ 0]^{\top}$}& Comp. & \multirow{2}*{\shortstack{$N_{p}$}} & Traj. & Ctrl. & Succ.\\
    \multicolumn{2}{c}{[m/s]} & Time[s] &  & Time[s] & Cost[${\rm m^{2}/s^{3}}$] & Rate[$\%$]\\
    \toprule
    \multirow{2}*{\shortstack{search based \\ method \cite{liu2017search}}} &  & \multirow{2}*{$29.02$} & \multirow{2}*{$1427$} & \multirow{2}*{$71.5$} & \multirow{2}*{$12.01$} & \multirow{2}*{$100.0$}\\
    \\
    \midrule
    \multirow{3}*{\shortstack{Our method \\ no jump point\\ $\rho_{c} = 1.0$}} &  & \multirow{3}*{$34.60$} & \multirow{3}*{$1340$} & \multirow{3}*{$69.1$} & \multirow{3}*{${13.25}$} & \multirow{3}*{$100.0$}\\
    \\
    \\
    \midrule
    \multirow{3}*{\shortstack{Our method \\ with jump point\\ $\rho_{c} = 1.0$}} &  & \multirow{3}*{$30.75$} & \multirow{3}*{$1388$} & \multirow{3}*{${73.3}$} & \multirow{3}*{$10.101$} &  \multirow{3}*{$100.0$}\\
    \\
    \\
    \midrule
    \multirow{3}*{\shortstack{Our method \\ no jump point\\ $\rho_{c} = 10.0$}} &  & \multirow{3}*{$48.28$} & \multirow{3}*{$1701$} & \multirow{3}*{$74.1$} & \multirow{3}*{$10.74$} &  \multirow{3}*{$100.0$}\\
    \\
    \\
    \midrule
    \multirow{3}*{\shortstack{Our method \\ with jump point\\ $\rho_{c} = 10.0$}} &  & \multirow{3}*{$44.32$} & \multirow{3}*{$1747$} & \multirow{3}*{$76.7$} & \multirow{3}*{$12.60$} &  \multirow{3}*{$100.0$}\\
    \\
    \\
    \midrule
    \multirow{3}*{\shortstack{Our method \\ no jump point\\ $\rho_{c} = 100.0$}} &  & \multirow{3}*{$44.93$} & \multirow{3}*{$1588$} & \multirow{3}*{$71.5$} & \multirow{3}*{$12.01$} &  \multirow{3}*{$100.0$}\\
    \\
    \\
    \midrule
    \multirow{3}*{\shortstack{Our method \\ with jump point\\ $\rho_{c} = 100.0$}} &  & \multirow{3}*{$41.74$} & \multirow{3}*{$1602$} & \multirow{3}*{$71.5$} & \multirow{3}*{$12.01$} &  \multirow{3}*{$100.0$}\\
    \\
    \\
    \midrule
    \multirow{4}*{\shortstack{sampling based \\ method \cite{zha2021exploiting}}} & mean & $124.94$ & $211$ & $101.0$ & $9.86$ &  \multirow{4}*{$70.0$}\\
    & std & $55.63$ & $38.2$ & $9.86$ & $2.91$ &  \\
    & min & $93.72$ & $183$ & $84.2$ & $5.77$ &  \\
    & max & $259.40$ & $298$ & $114.8$ & $15.27$ &  \\
    \bottomrule
    \toprule
    \multicolumn{2}{c}{${v}_{0} = [2\ 2]^{\top}$} &  &  &  & \\
    \toprule
    \multirow{2}*{\shortstack{search based \\ method \cite{liu2017search}}} &  & \multirow{2}*{$22.41$} & \multirow{2}*{$1113$} & \multirow{2}*{$71.7$} & \multirow{2}*{$14.02$} & \multirow{2}*{$100.0$}\\
    \\
    \midrule
    \multirow{3}*{\shortstack{Our method \\ no jump point\\ $\rho_{c} = 1.0$}} &  & \multirow{3}*{$24.45$} & \multirow{3}*{$1012$} & \multirow{3}*{${81.0}$} & \multirow{3}*{$15.03$} &  \multirow{3}*{$100.0$}\\
    \\
    \\
    \midrule
    \multirow{3}*{\shortstack{Our method \\ with jump point\\ $\rho_{c} = 1.0$}} &  & \multirow{3}*{$23.78$} & \multirow{3}*{$1077$} & \multirow{3}*{${78.9}$} & \multirow{3}*{$8.20$} &  \multirow{3}*{$100.0$}\\
    \\
    \\
    \midrule
    \multirow{3}*{\shortstack{Our method \\ no jump point\\ $\rho_{c} = 10.0$}} &  & \multirow{3}*{$37.30$} & \multirow{3}*{$1388$} & \multirow{3}*{$74.9$} & \multirow{3}*{$13.28$} &  \multirow{3}*{$100.0$}\\
    \\
    \\
    \midrule
    \multirow{3}*{\shortstack{Our method \\ with jump point\\ $\rho_{c} = 10.0$}} &  & \multirow{3}*{$36.67$} & \multirow{3}*{$1417$} & \multirow{3}*{$74.9$} & \multirow{3}*{$13.28$} &  \multirow{3}*{$100.0$}\\
    \\
    \\
    \midrule
    \multirow{3}*{\shortstack{Our method \\ no jump point\\ $\rho_{c} = 100.0$}} &  & \multirow{3}*{$30.97$} & \multirow{3}*{$1145$} & \multirow{3}*{$71.7$} & \multirow{3}*{$14.02$} &  \multirow{3}*{$100.0$}\\
    \\
    \\
    \midrule
    \multirow{3}*{\shortstack{Our method \\ with jump point\\ $\rho_{c} = 100.0$}} &  & \multirow{3}*{$30.38$} & \multirow{3}*{$1145$} & \multirow{3}*{$71.7$} & \multirow{3}*{$14.02$} &  \multirow{3}*{$100.0$}\\
    \\
    \\
    \midrule
    \multirow{4}*{\shortstack{sampling based \\ method \cite{zha2021exploiting}}} & mean & $88.50$ & $169$ & $117.0$ & $14.18$ &  \multirow{4}*{$70.0$}\\
    & std & $43.88$ & $55.7$ & $10.58$ & $5.14$ &  \\
    & min & $17.09$ & $63$ & $103.6$ & $8.28$ &  \\
    & max & $142.83$ & $235$ & $134.9$ & $21.83$ &  \\
    \bottomrule
    \end{tabular}
    }
    \end{center}
    \vspace{-21pt}
\end{table}

\vspace{-1pt}
Further, we conduct a more extensive analysis to evaluate the effect of different values of parameter $\rho_c$ in more detail. Table~\ref{table:globalplanner} contains more detailed results and also presents comparisons against the sampling-based (RRT*) method in~\cite{zha2021exploiting}, which was adapted herein to feature a trapezoidal velocity pattern to connect any two nodes in the tree to better match our search-based global planning method and enable fair comparisons. Due to the non-deterministic nature of this method, we perform $10$ trials and report statistics. In all other cases (that are deterministic), we perform a single trial. 

Both our collision-inclusive method and collision avoidance in~\cite{liu2017search} can generate kinodynamically-feasible trajectories. When the initial velocity is $\bm{v}_{0} = [0\ 0]^{\top}$\;m/s and $\rho_{c} = 1.0$, our method without jump points tends to generate a path with the shortest duration compared to both the collision-inclusive planner with jump points and the collision-avoidance planner. However, the control cost for doing so is slightly higher. The computational time of the collision-inclusive planner with jump points is the second-lowest among all the methods. Comparing these results with those obtained by the RRT* method in~\cite{zha2021exploiting}, we find the RRT*-based approach is time consuming since node rewiring requires significant computational time (about 88\% of total time). Also, results are not deterministic compared to the search-based method. Thus, we deduce that the search-based collision-inclusive method with jump points can be the global planner in our unified collision-inclusive motion planning and control framework with parameters selected in this section.

\subsection{Simulated Tests of our Unified Collision-inclusive Method}\label{subsec:simulation framework}
We first test our unified collision-inclusive motion planning and control strategy in a double corridor environment with online sensing, and compare its performance against that of a collision avoidance framework similar to~\cite{tordesillas2021faster}. 
In the collision avoidance framework, the global planner is the search-based method in~\cite{liu2017search}; we also make the optimistic assumption of treating the unknown space as free. The local trajectory generation method is based on gradients~\cite{gao2017gradient} and time duration adjustment~\cite{liu2017planning}. We design a backup safety trajectory to ensure the robot will stop at the frontier. 
Then, we test both methods in a double corridor environment populated with circular isolated obstacles of increasing density. 
In all cases, each method is run for $10$ times with the same initial configuration and parameter settings. 

\begin{figure}[!t]
\vspace{6pt}
 \begin{subfigure}{.25\textwidth}
  \centering
  \includegraphics[trim={4cm 0.2cm 4cm 0.2cm}, clip, width=0.925\linewidth]{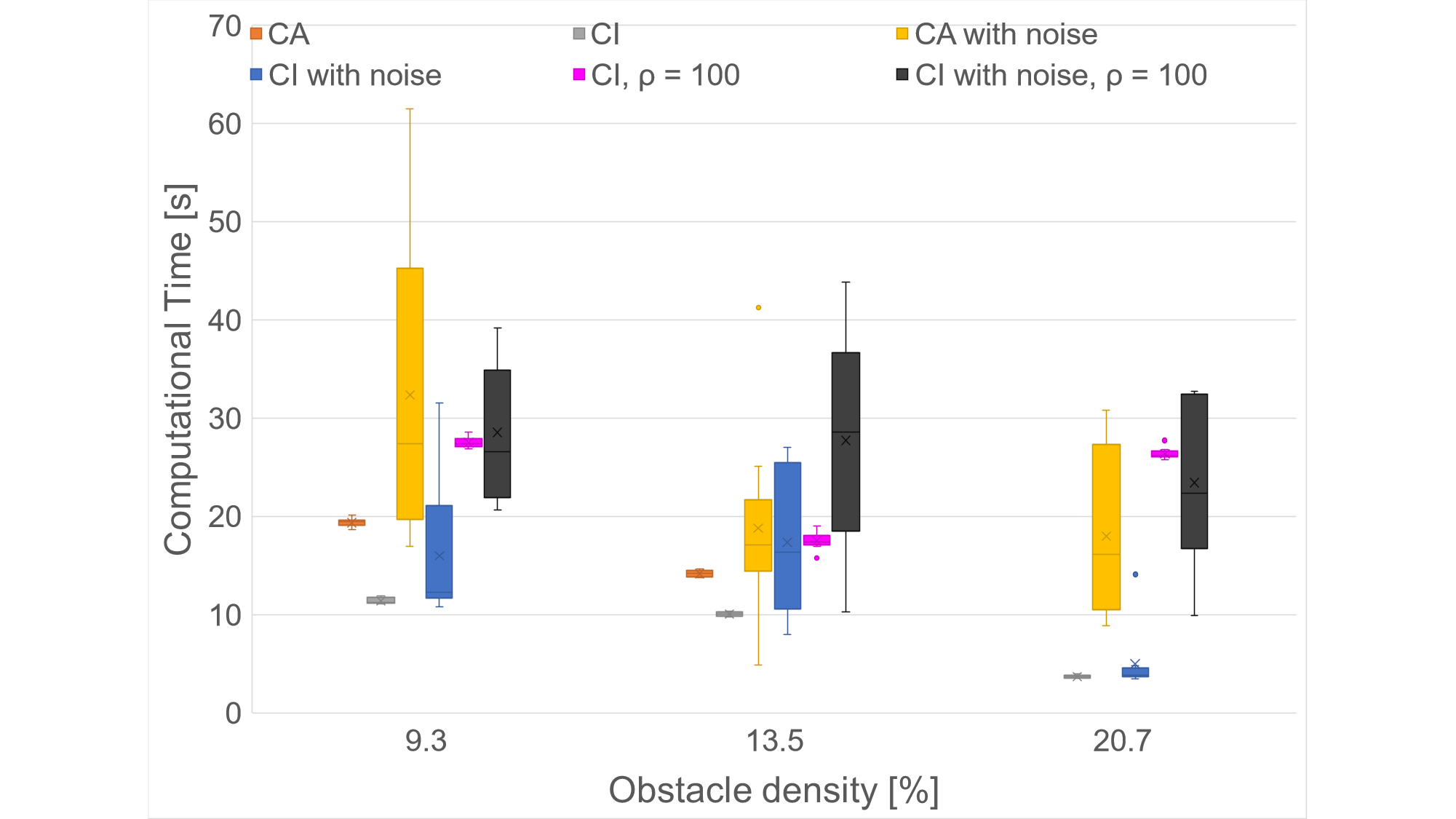}
  \vspace{-3pt}
  \caption{$T_{rep} = 5.0$\;s.}
 \end{subfigure}%
 \begin{subfigure}{.25\textwidth}
  \centering
  \includegraphics[trim={4cm 0.2cm 4cm 0.2cm}, clip,width=0.925\linewidth]{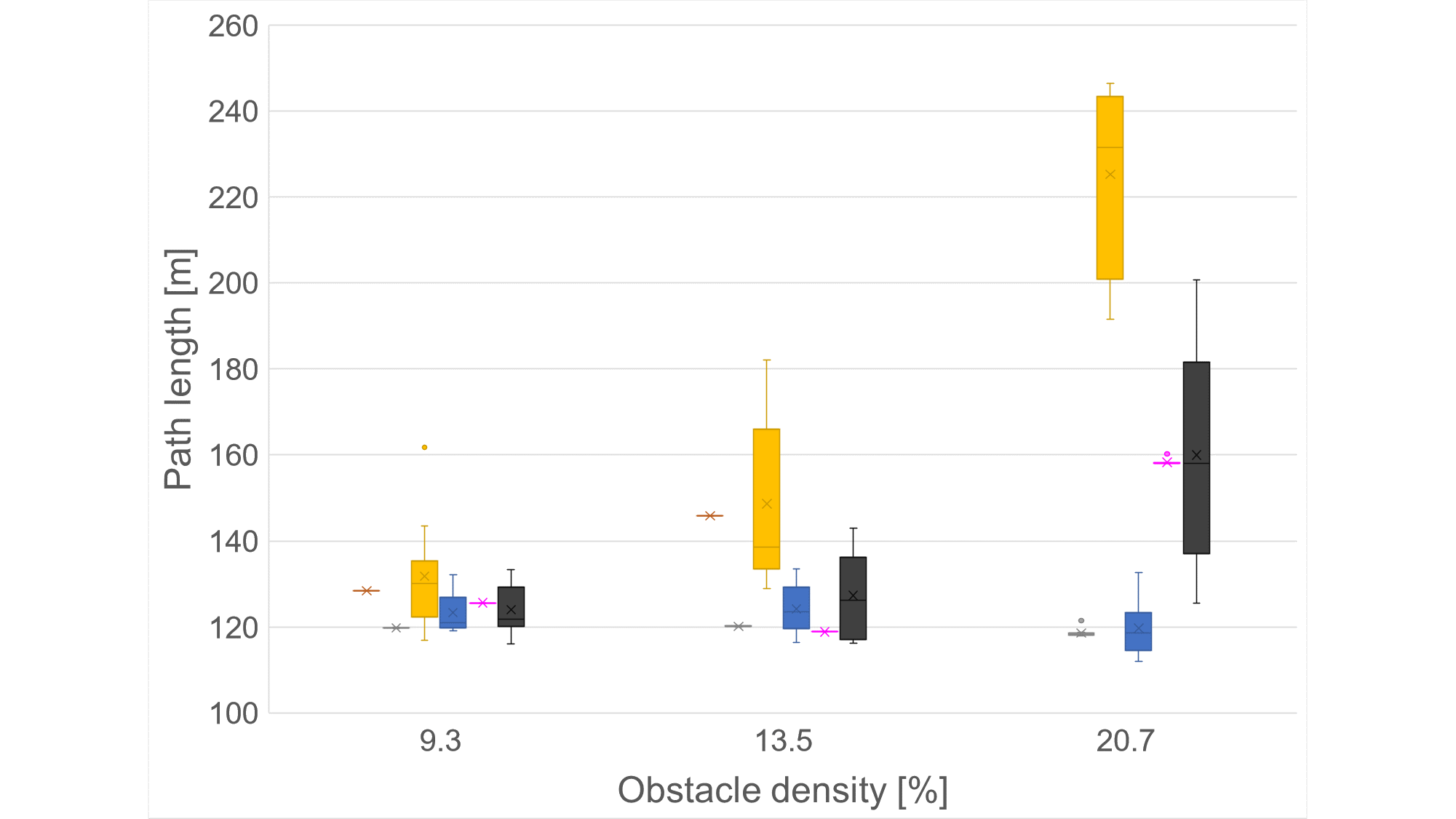}
  \vspace{-3pt}
  \caption{$T_{rep} = 5.0$\;s.}
 \end{subfigure}
 \vspace{0pt}
 \medskip
 \begin{subfigure}{.25\textwidth}
  \centering
  \includegraphics[trim={4cm 0.2cm 4cm 0.2cm}, clip,width=0.925\linewidth]{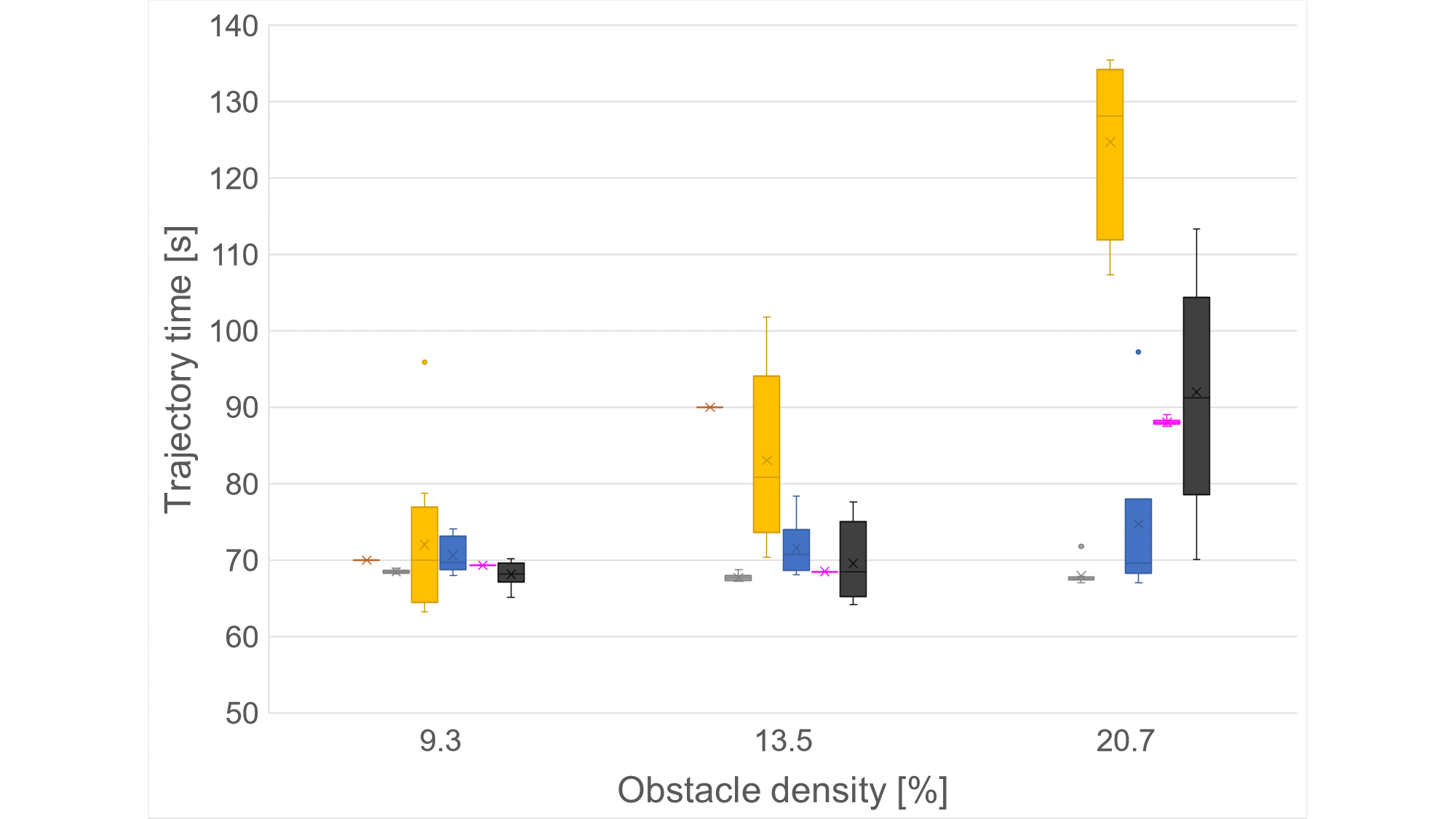}
  \vspace{-3pt}
  \caption{$T_{rep} = 5.0$\;s.}
 \end{subfigure}%
 \begin{subfigure}{.25\textwidth}
  \centering
  \includegraphics[trim={4cm 0.2cm 4cm 0.2cm}, clip,width=0.925\linewidth]{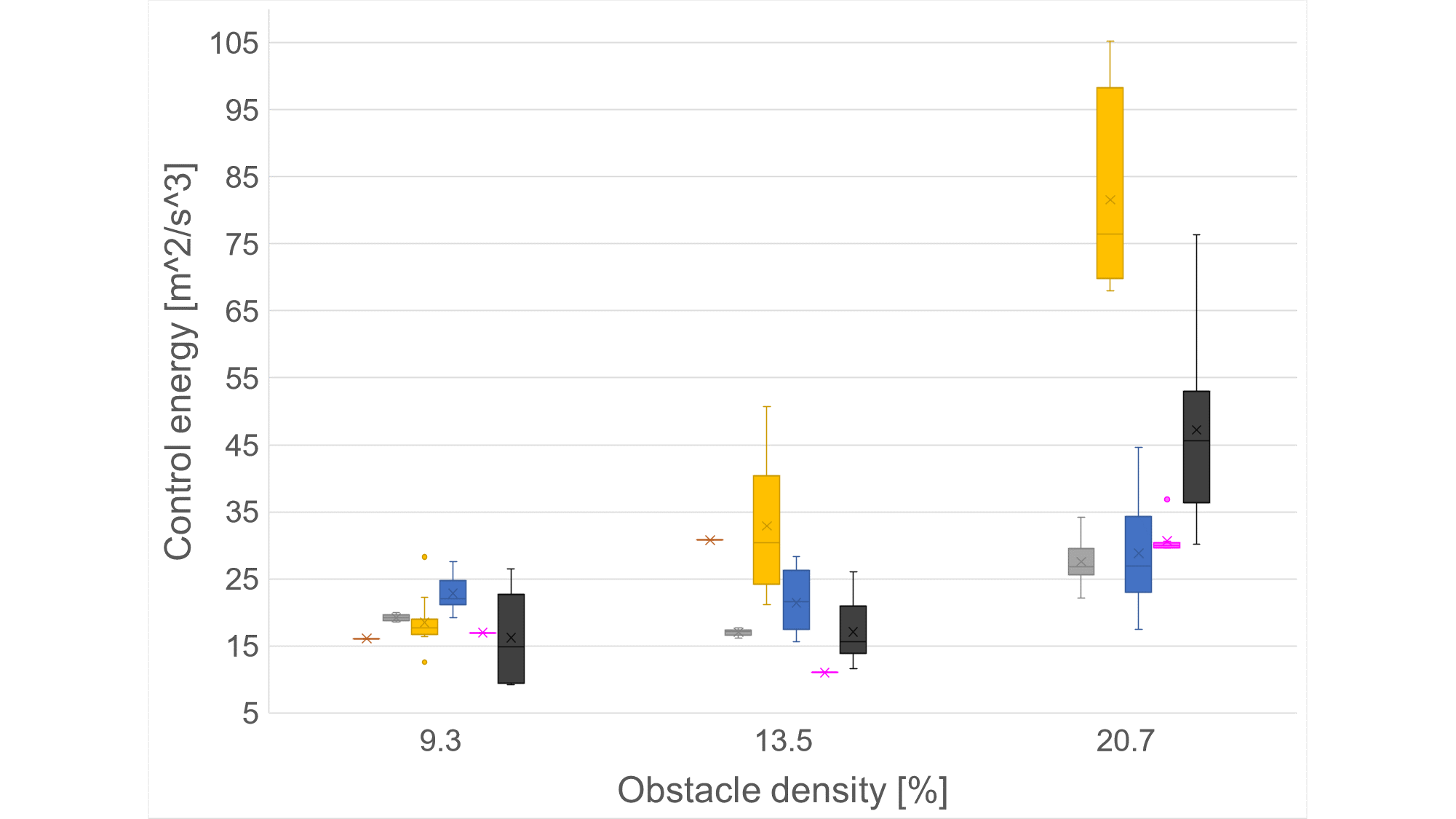}
  \vspace{-3pt}
  \caption{$T_{rep} = 5.0$\;s.}
 \end{subfigure}%

 \begin{subfigure}{.25\textwidth}
  \centering
  \includegraphics[trim={4cm 0.2cm 4cm 0.2cm}, clip, width=0.925\linewidth]{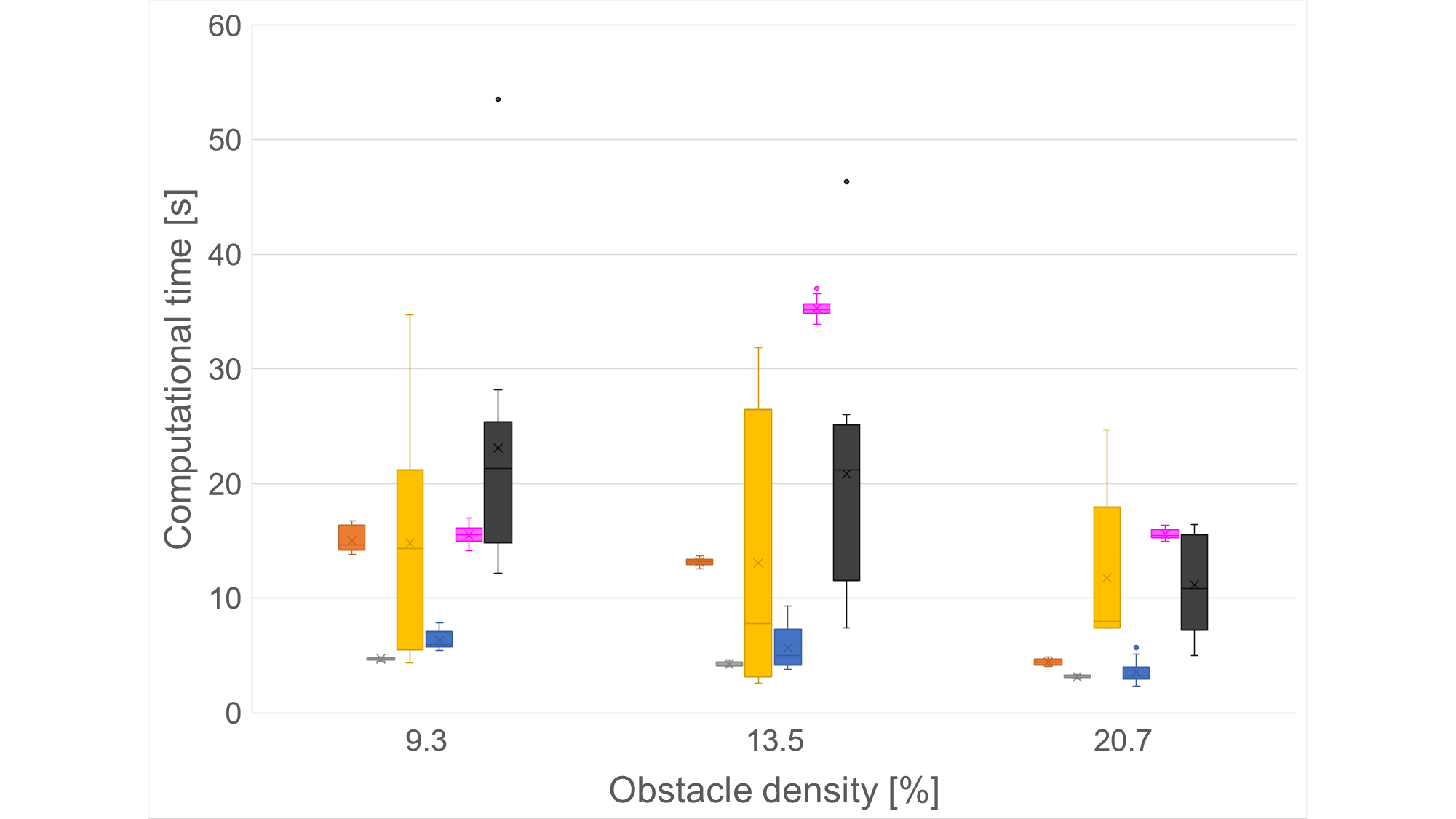}
  \vspace{-3pt}
  \caption{$T_{rep} = 10.0$\;s.}
 \end{subfigure}%
 \begin{subfigure}{.25\textwidth}
  \centering
  \includegraphics[trim={4cm 0.2cm 4cm 0.2cm}, clip,width=0.925\linewidth]{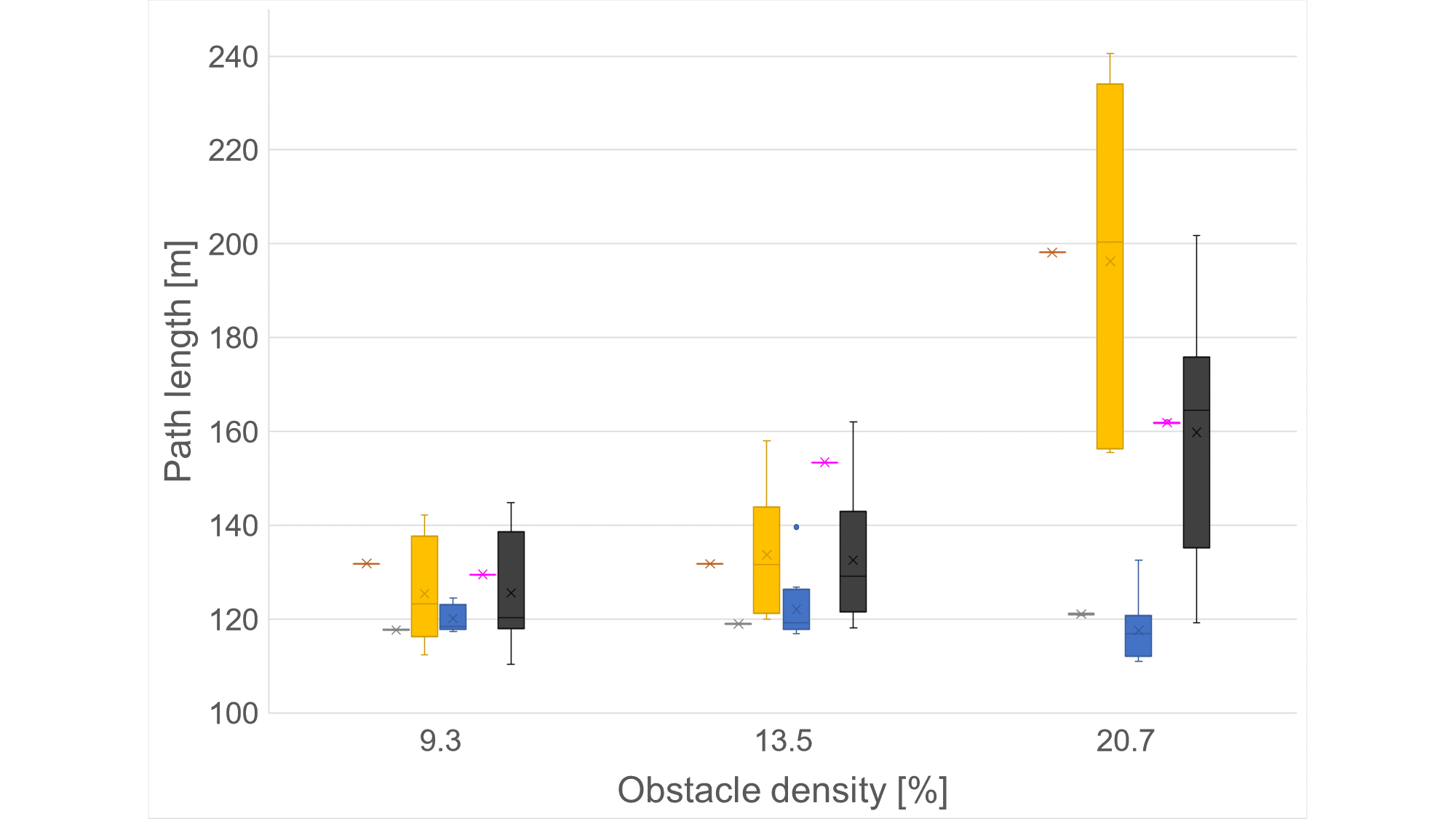}
  \vspace{-3pt}
  \caption{$T_{rep} = 10.0$\;s.}
 \end{subfigure}
 \vspace{0pt}
 \medskip
 \begin{subfigure}{.25\textwidth}
  \centering
  \includegraphics[trim={4cm 0.2cm 4cm 0.2cm}, clip,width=0.925\linewidth]{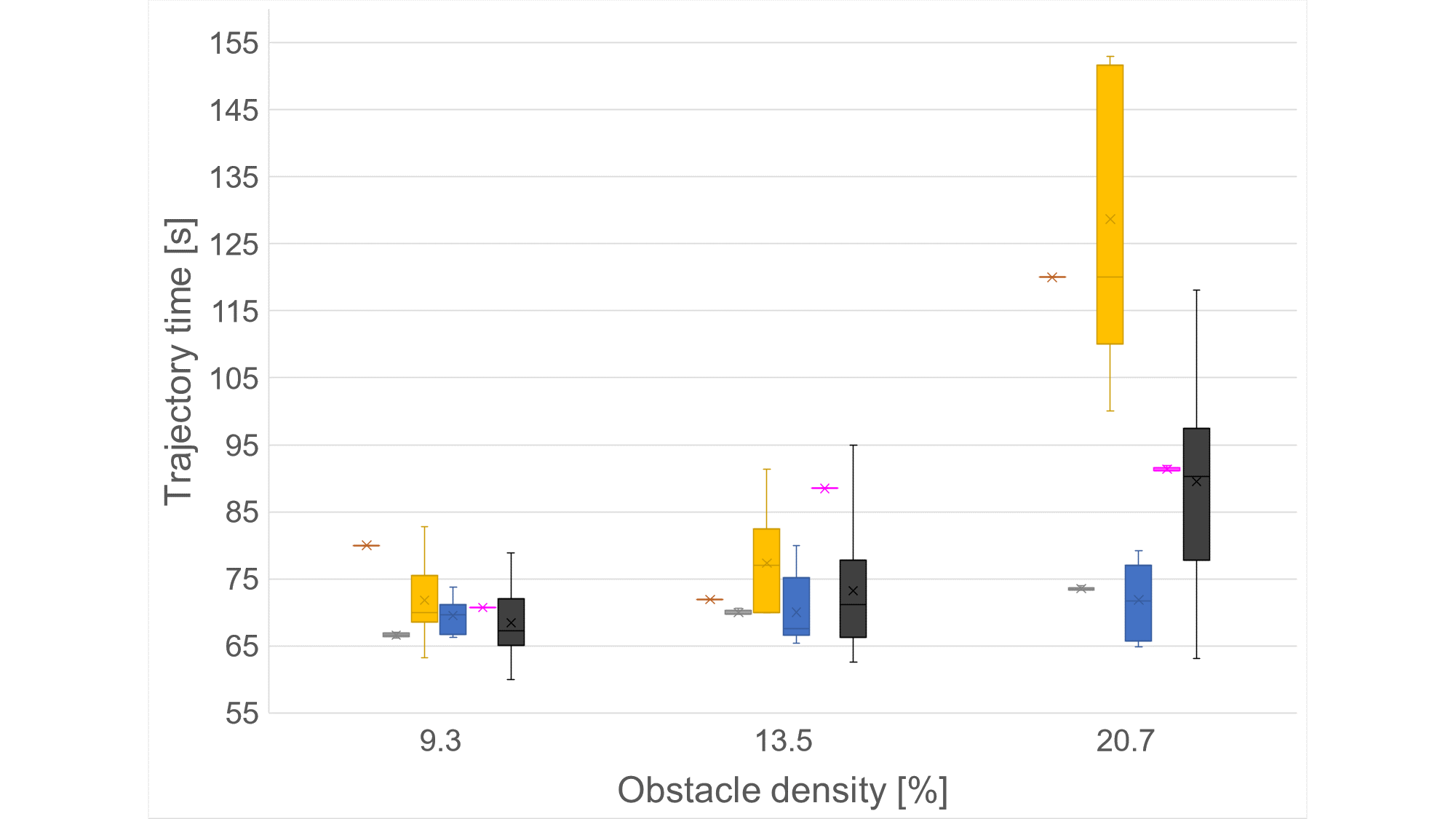}
  \vspace{-3pt}
  \caption{$T_{rep} = 10.0$\;s.}
 \end{subfigure}%
 \begin{subfigure}{.25\textwidth}
  \centering
  \includegraphics[trim={4cm 0.2cm 4cm 0.2cm}, clip,width=0.925\linewidth]{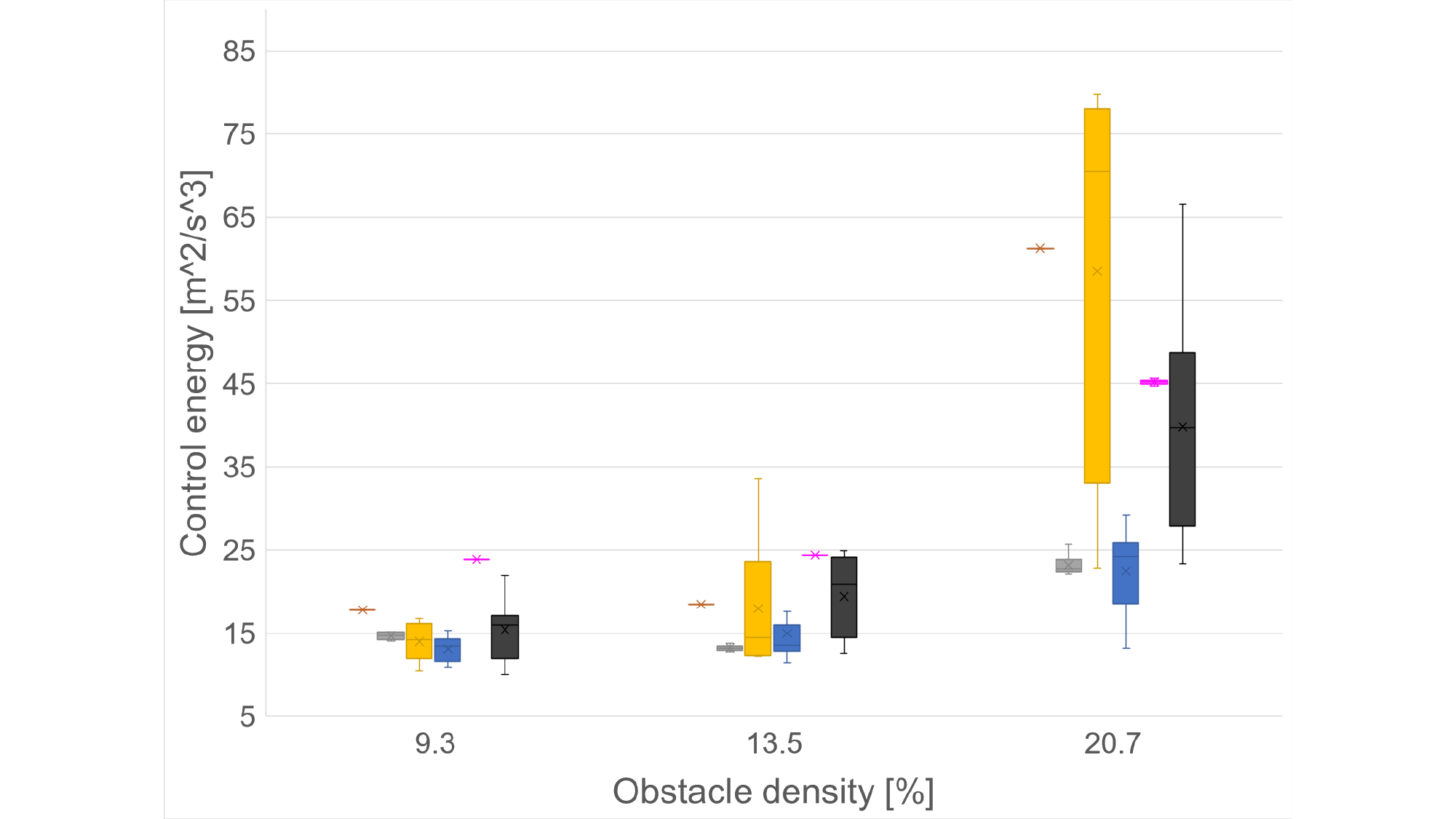}
  \vspace{-3pt}
  \caption{$T_{rep} = 10.0$\;s.}
 \end{subfigure}%
 \vspace{-6pt}
 \caption{Computational time, path length, trajectory time and control energy for collision-avoidance (CA) without (in red) and with (in yellow) noise, collision-inclusive (CI) without (in grey) and with (in blue) noise with $\rho_{c} = 1.0$ and collision-inclusive (CI) without (in magenta) and with (in black) noise with $\rho_{c} = 100.0$ methods, for two cases of replanning time $T_{rep} = 5.0$\;s and $T_{rep} = 10.0$\;s. }
 \label{fig:static replan}
 \vspace{0pt}
\end{figure}

\begin{figure}[!h]
\vspace{-9pt}
 \begin{subfigure}{.22\textwidth}
  \centering
  \includegraphics[trim={6cm 0.2cm 6cm 0.2cm}, clip, height=0.9\textwidth]{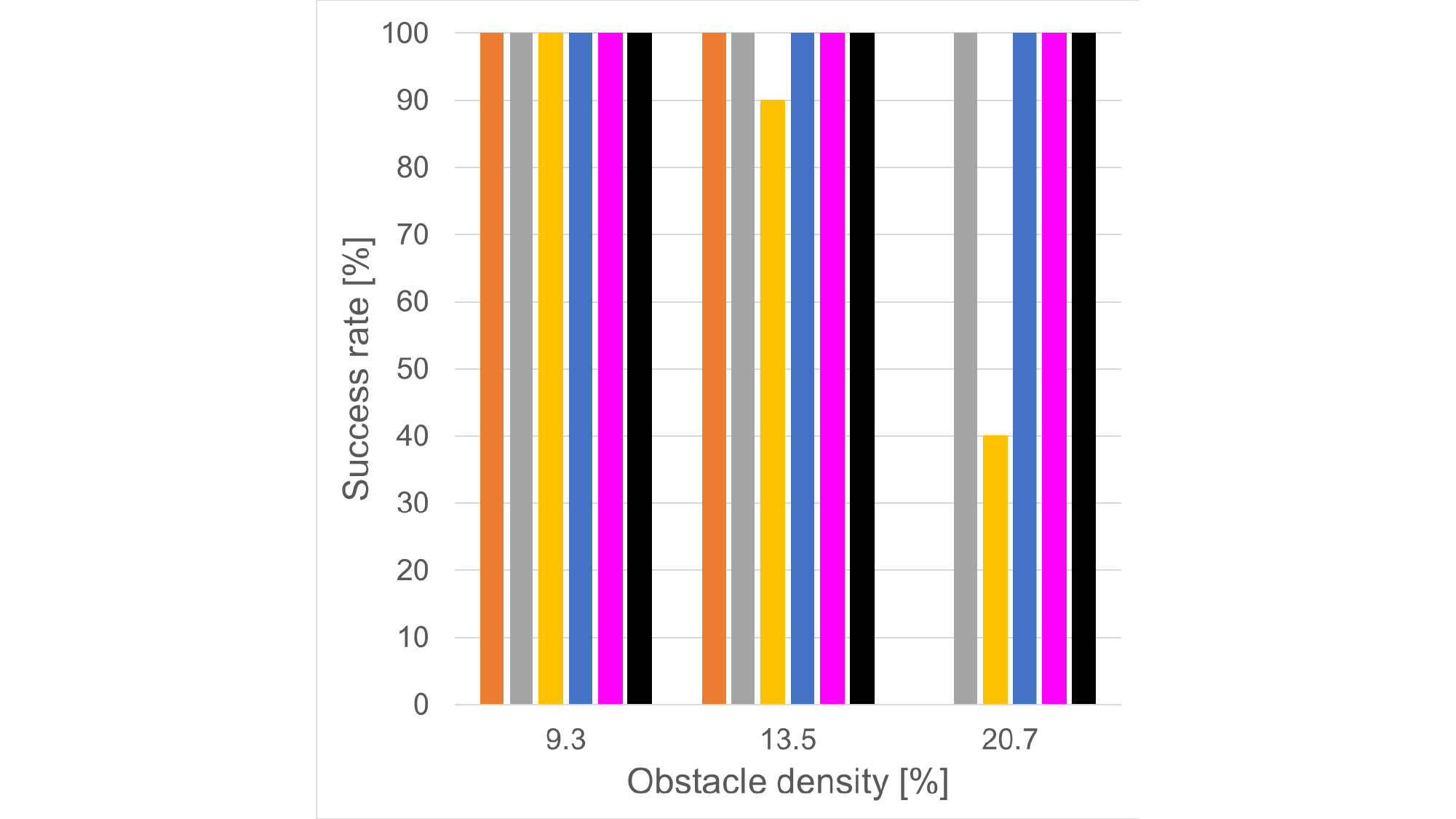}
  \vspace{-15pt}
  \caption{$T_{rep} = 5.0$\;s.}
 \end{subfigure}%
 \hspace{-6pt}
 \begin{subfigure}{.235\textwidth}
  \centering
  \includegraphics[trim={4cm 0.2cm 4.5cm 0.2cm}, clip, height=0.85\textwidth]{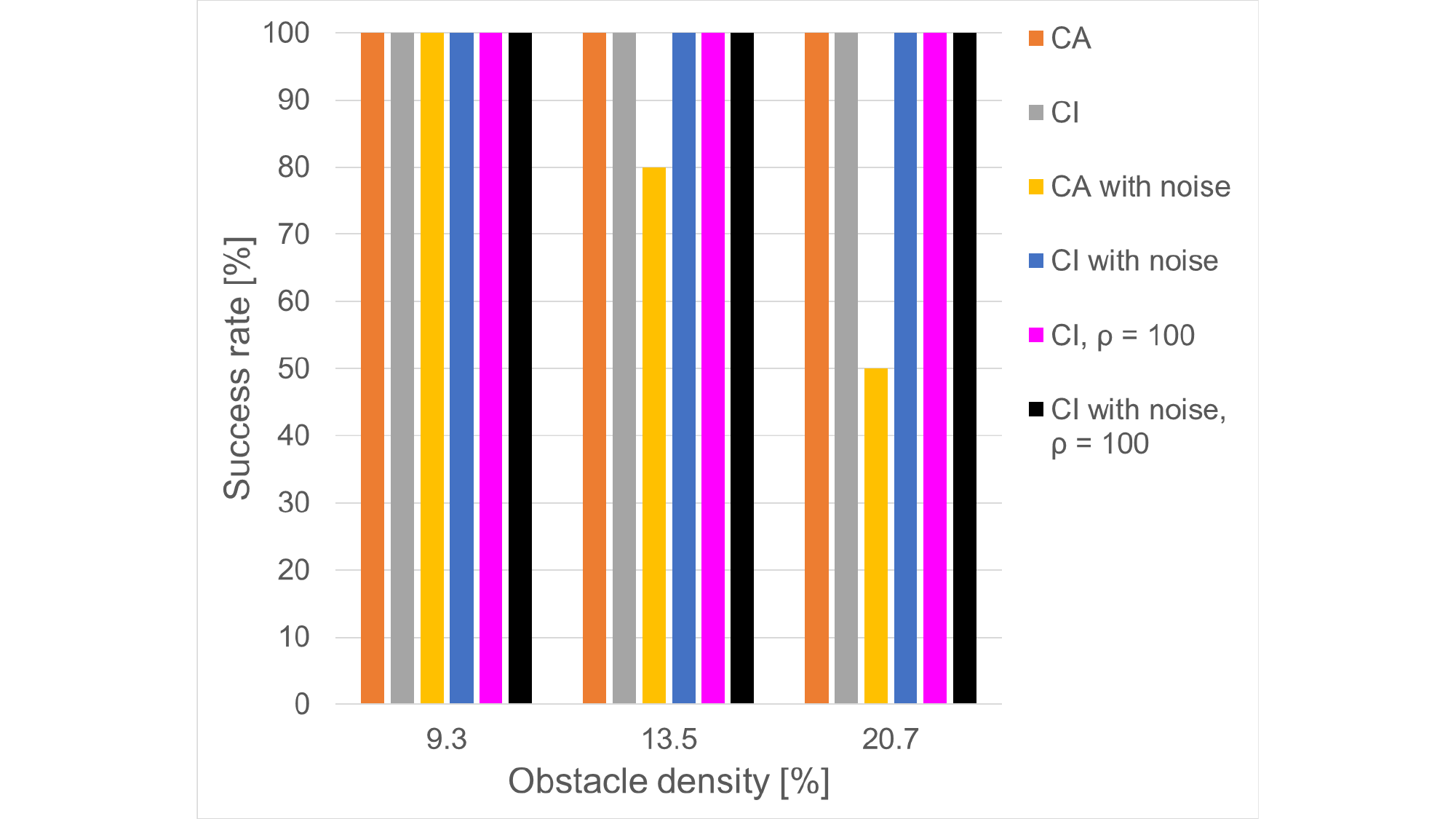}
  \vspace{-15pt}
  \caption{$T_{rep} = 10.0$\;s.}
 \end{subfigure}
 \vspace{-3pt}
 \caption{Success rates of both collision-avoidance and collision-inclusive frameworks (same notation as in Fig.~\ref{fig:static replan}). } 
 \label{fig:success rate}
 \vspace{-21pt}
\end{figure}

\begin{figure*}[!h]
	\vspace{6pt}
	\centering
	\begin{subfigure}[b]{0.325\textwidth}
		\includegraphics[width=0.85\linewidth]{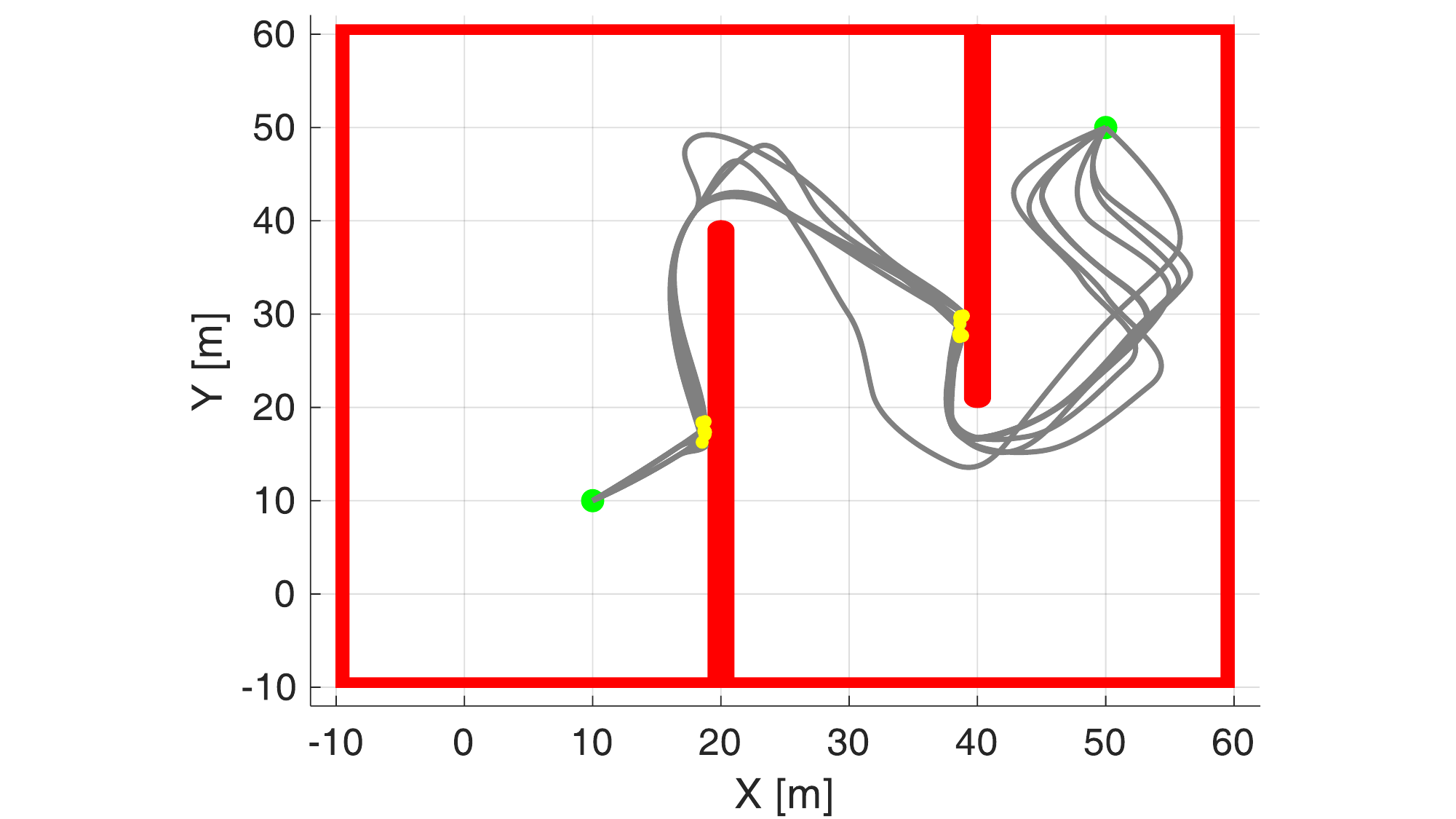}
		\vspace{-3pt}
		\caption{$T_{rep}=5.0$, obstacle density=$9.3 \%$.}
		\label{fig:5.0, 0.02}
	\end{subfigure}
	\begin{subfigure}[b]{0.325\textwidth}
		\includegraphics[width=0.85\linewidth]{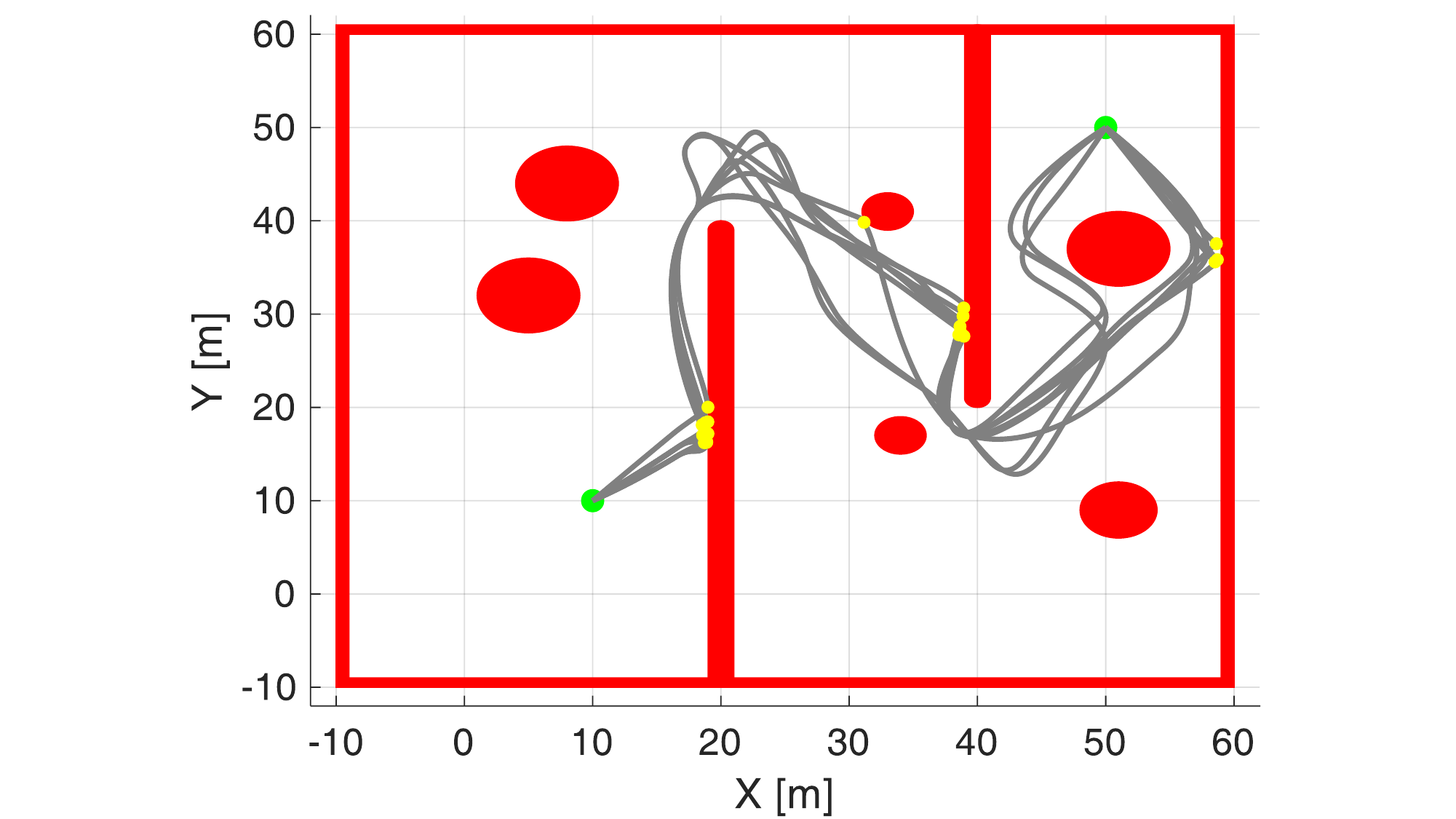}
		\vspace{-3pt}
		\caption{$T_{rep}=5.0$, obstacle density=$13.5 \%$.}
		\label{fig:5.0, 0.04}
	\end{subfigure}
	\begin{subfigure}[b]{0.325\textwidth}
		\includegraphics[width=0.85\linewidth]{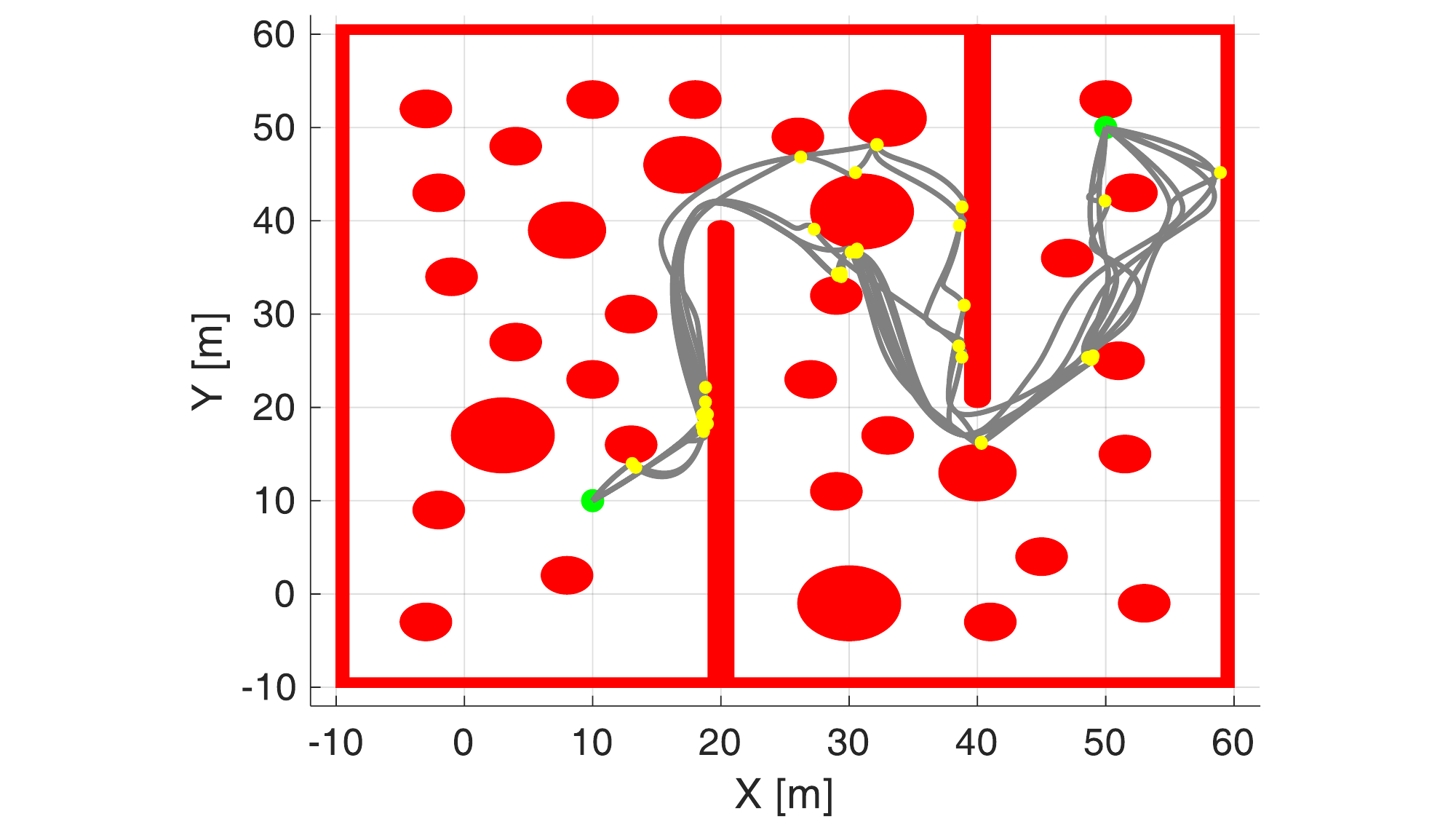}
		\vspace{-3pt}
		\caption{$T_{rep}=5.0$, obstacle density=$20.7 \%$.}
		\label{fig:5.0, 0.06}
	\end{subfigure}
	\vspace{-1pt}
	
	\vspace{3pt}
	\centering
	\begin{subfigure}[b]{0.325\textwidth}
		\includegraphics[width=0.85\linewidth]{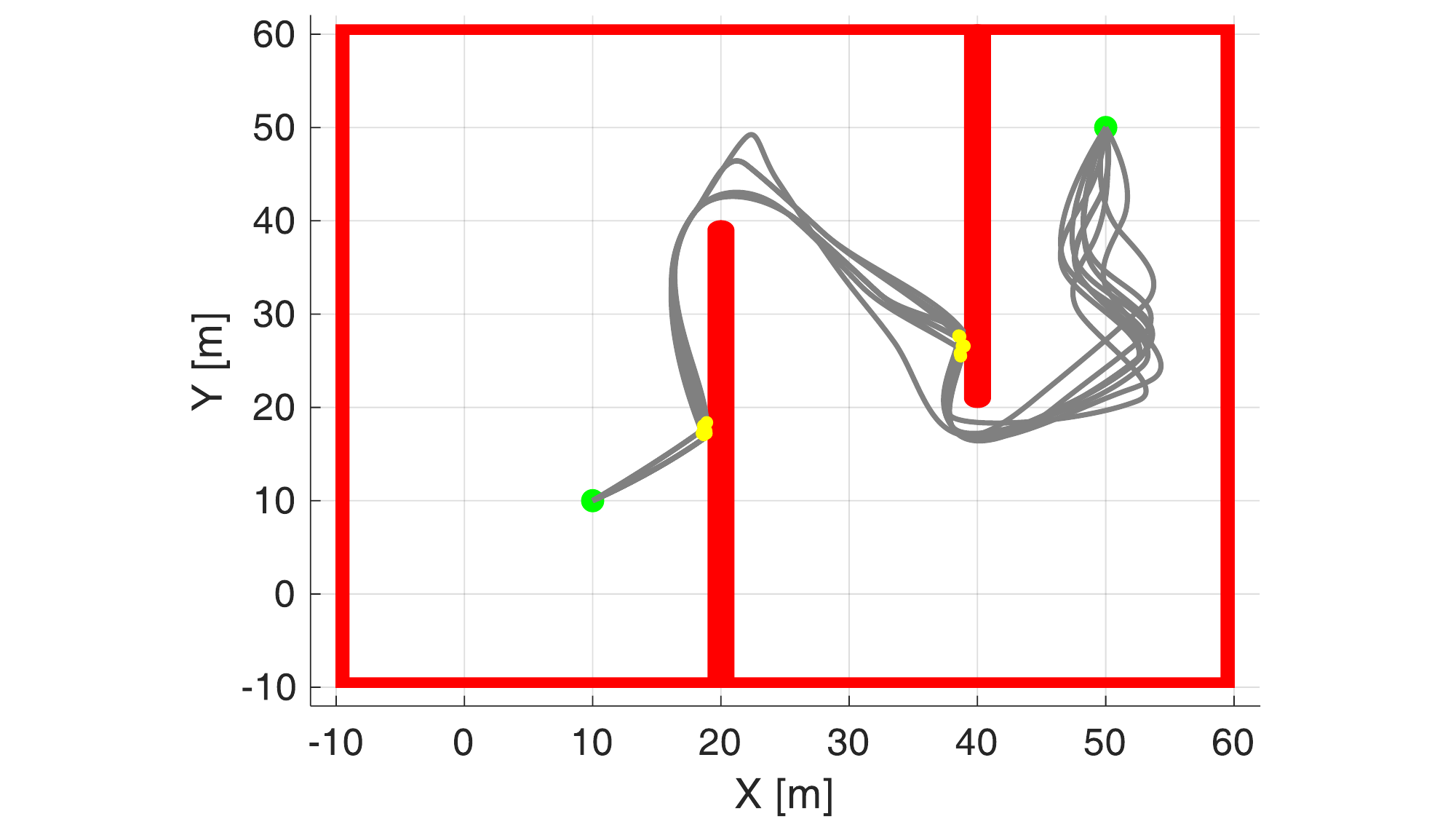}
		\vspace{-3pt}
		\caption{$T_{rep}=10.0$, obstacle density=$9.3 \%$.}
		\label{fig:10.0, 0.02}
	\end{subfigure}
	\begin{subfigure}[b]{0.325\textwidth}
		\includegraphics[width=0.85\linewidth]{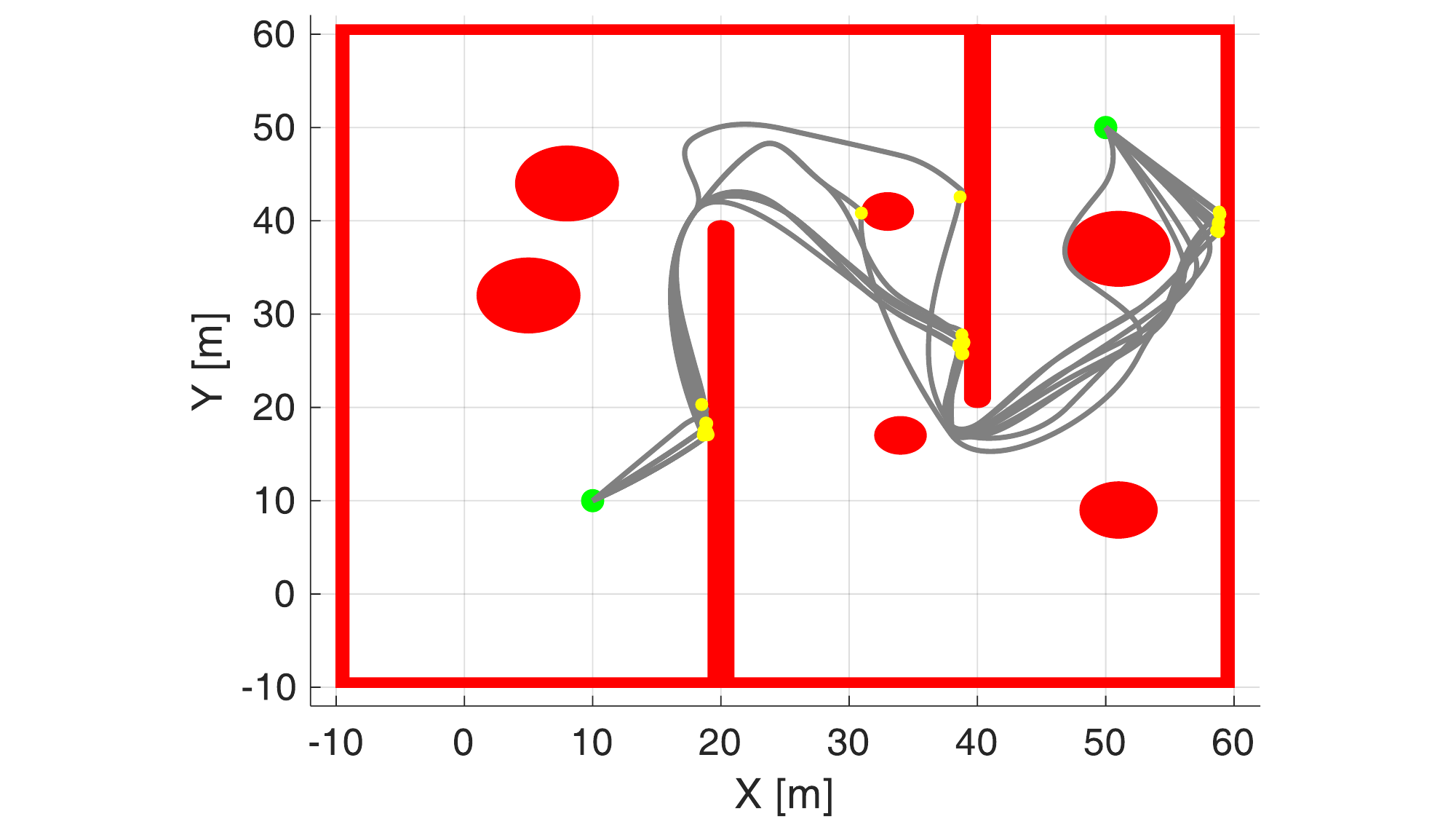}
		\vspace{-3pt}
		\caption{$T_{rep}=10.0$, obstacle density=$13.5 \%$.}
		\label{fig:10.0, 0.04}
	\end{subfigure}
	\begin{subfigure}[b]{0.325\textwidth}
		\includegraphics[width=0.85\linewidth]{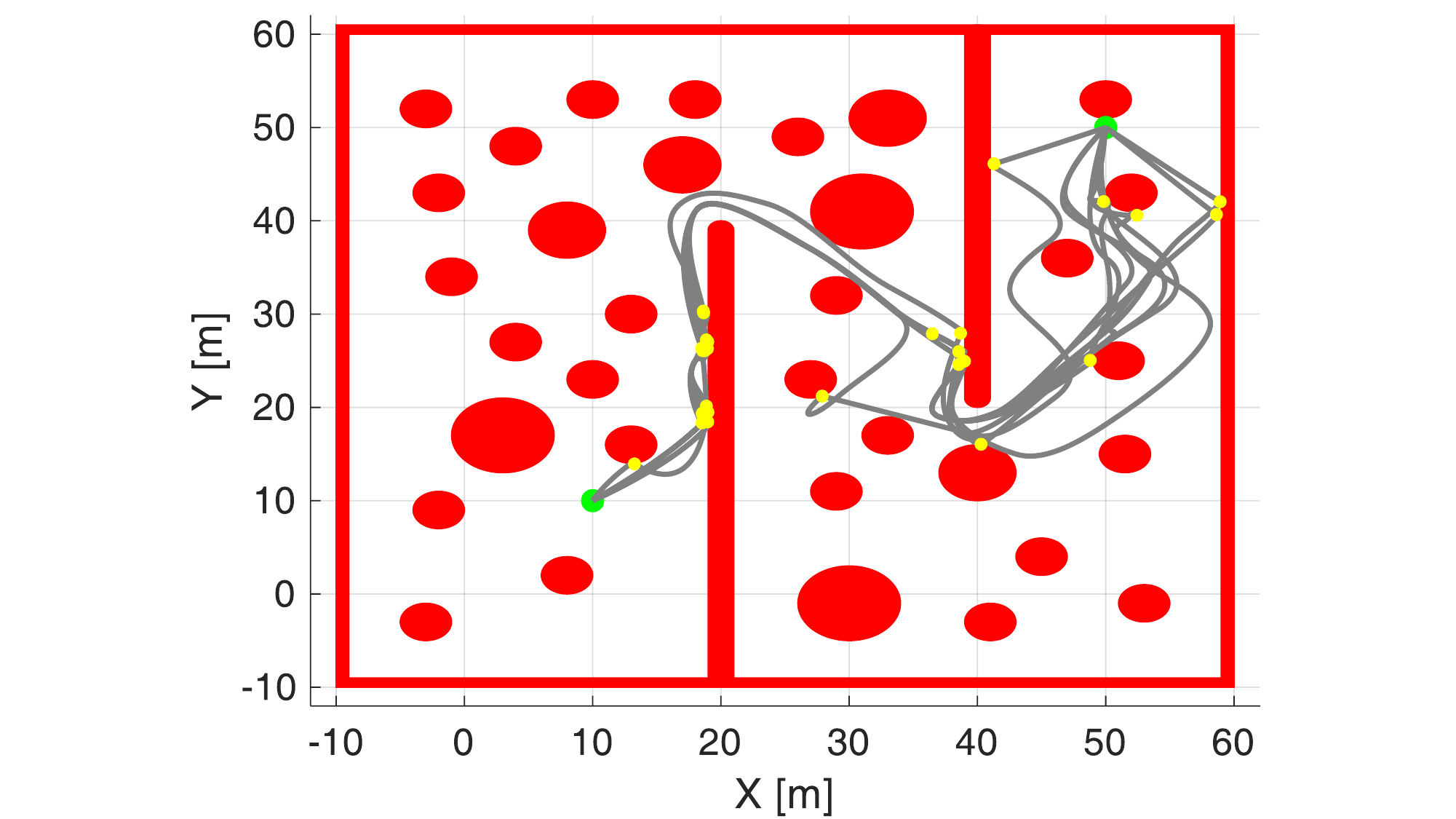}
		\vspace{-3pt}
		\caption{$T_{rep}=10.0$, obstacle density=$20.7 \%$.}
		\label{fig:10.0, 0.06}
	\end{subfigure}
	\vspace{-1pt}
	\caption{Simulated trajectories of the unified collision-inclusive motion planning and control framework with online sensing and noise added to the input of the global planner in the double corridor for environment with increasing obstacle density.}
	\vspace{-12pt}
	\label{fig:simulation examples}
\end{figure*}

We test with and without additive estimation noise in the global planner. Position estimation noise is zero-mean truncated Gaussian with variance of $0.3$ and bounds of $\pm0.9$. 
Velocity estimation noise is zero-mean truncated Gaussian with variance of $0.1$ and bounds of $\pm0.3$. 
Comparative results are presented in Figs.~\ref{fig:static replan} and~\ref{fig:success rate}. Output trajectories of our method (with added noise) are shown in Fig.~\ref{fig:simulation examples}.

With reference to Fig.~\ref{fig:static replan}, when replanning every $5$ sec, our method generates shorter paths with a lower trajectory time on average. When the obstacle density is low ($9.3 \%$), our method generates trajectories with higher control energy; however, when the obstacle density increases ($13.5 \% \& 20.7 \%$) our method requires lower control energy since the robot can utilize the obstacles to change its heading.\footnote{~In collision avoidance, and with $T_{rep} = 5.0$\;s, the robot can get trapped oscillating in a area to avoid collisions and cannot reach the goal; however, adding some random behavior may help the robot break the tie.}
When replanning every $10$ sec, our method consistently generates paths with lower length on average. Our method also has lower trajectory times and control energy. 

With $\rho_{c} = 100$, $T_{rep} = 5.0$\;s, our algorithm causes oscillations around the obstacle by avoiding collisions which increase path length and trajectory time. However, its ability to use collisions makes its output trajectory better than the collision-avoidance strategy in terms of control energy, trajectory and path length. 
With $T_{rep} = 10.0$\;s the path length, trajectory time and control energy of the collision-inclusive trajectory is higher since there is no safety maneuver making the robot stop before the frontier; hence it will have to turn sharply and possibly oscillate when replanning. The computational time of the collision-inclusive planning is higher since it visits more nodes in the graph.

Further, success rates of the proposed collision-inclusive method and collision avoidance are shown in Fig.~\ref{fig:success rate}. Our method has higher success rates as it addresses over-conservativeness in collision avoidance to ensure safety.

\begin{figure}[!h]
\vspace{6pt}
 \begin{subfigure}{.235\textwidth}
  \centering
  \includegraphics[trim={4cm 0.2cm 4cm 0.2cm}, clip, width=0.75\linewidth]{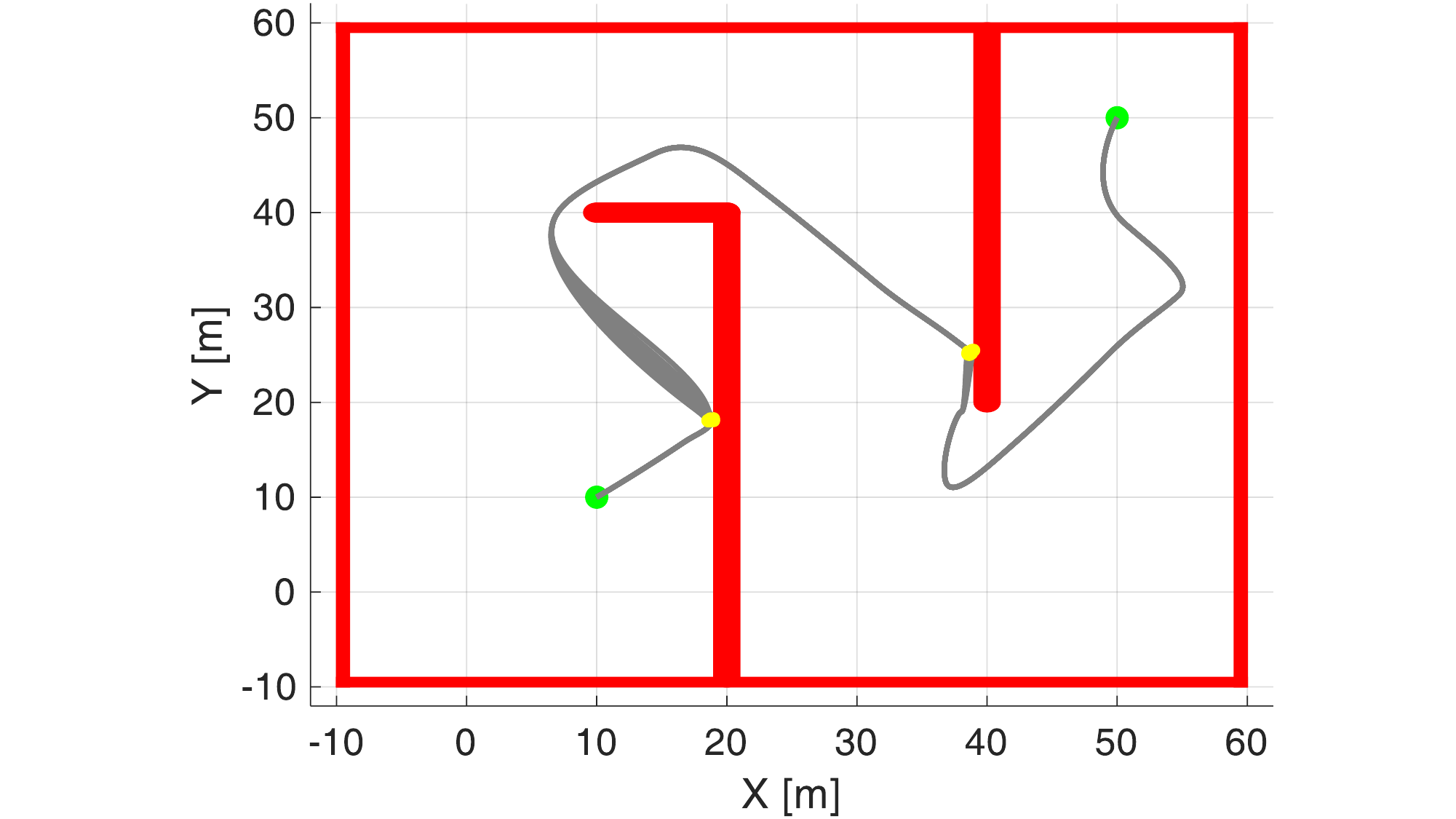}
  \vspace{-3pt}
  \caption{$T_{rep} = 5.0$\;s.}
  \label{fig:strategy simulation m1 5}
 \end{subfigure}%
 \begin{subfigure}{.235\textwidth}
  \centering
  \includegraphics[trim={4cm 0.2cm 4cm 0.2cm}, clip,width=0.75\linewidth]{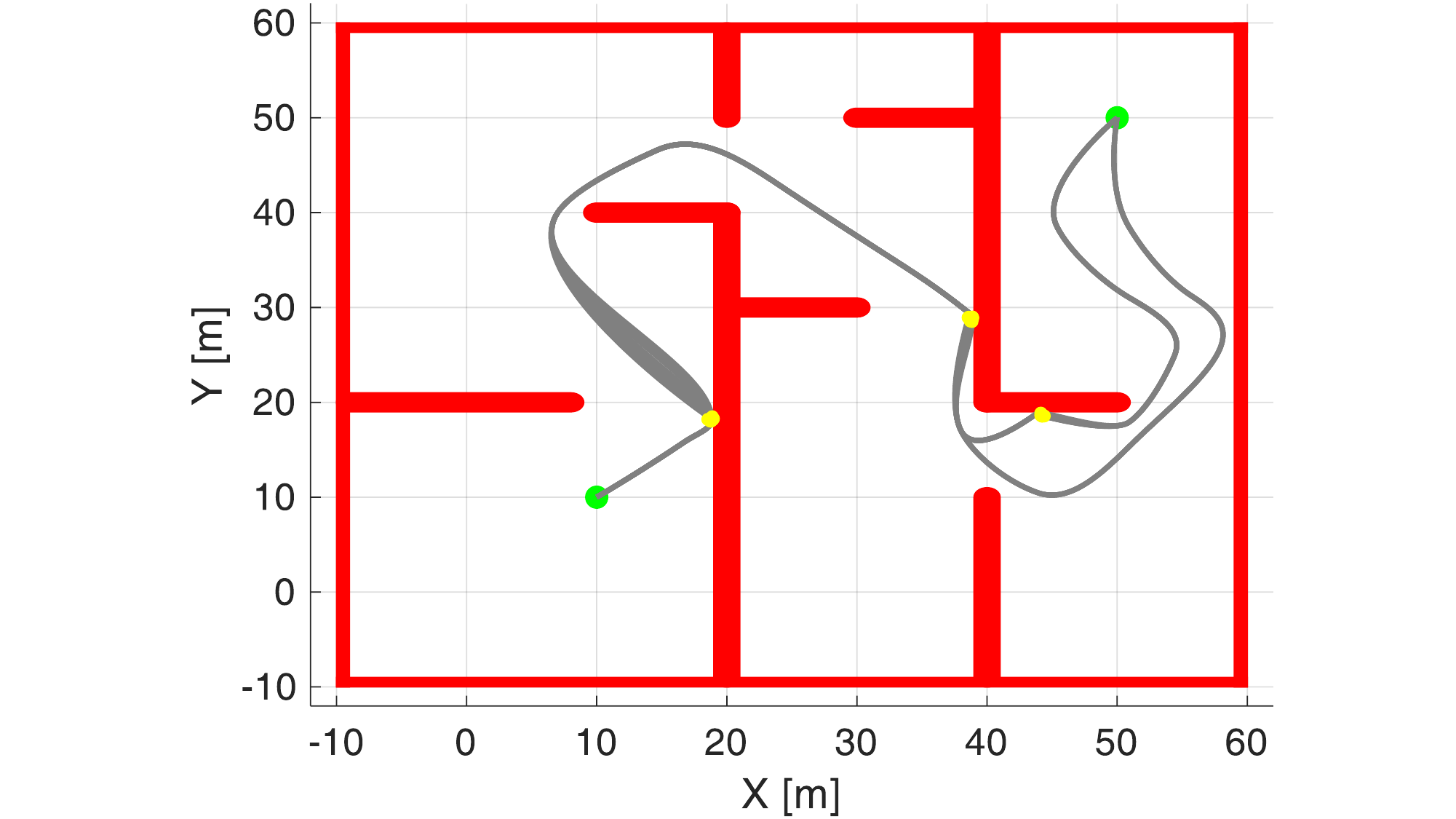}
  \vspace{-3pt}
  \caption{$T_{rep} = 5.0$\;s.}
  \label{fig:strategy simulation m2 5}
 \end{subfigure}
 \vspace{-1pt}
 \medskip
 \begin{subfigure}{.235\textwidth}
  \centering
  \includegraphics[trim={4cm 0.2cm 4cm 0.2cm}, clip,width=0.75\linewidth]{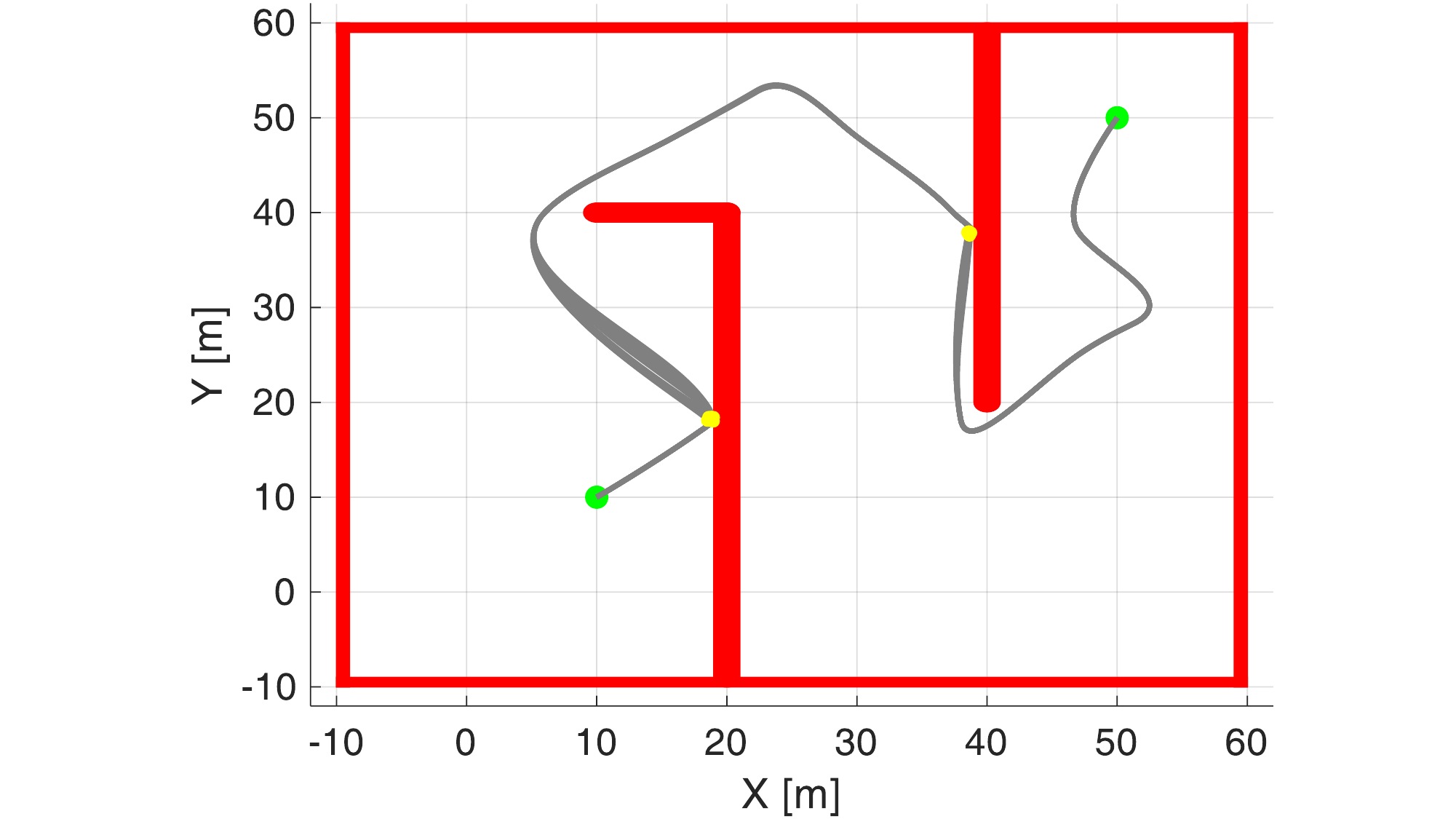}
  \vspace{-3pt}
  \caption{$T_{rep} = 10.0$\;s.}
  \label{fig:strategy simulation m1 10}
 \end{subfigure}%
 \begin{subfigure}{.235\textwidth}
  \centering
  \includegraphics[trim={4cm 0.2cm 4cm 0.2cm}, clip,width=0.75\linewidth]{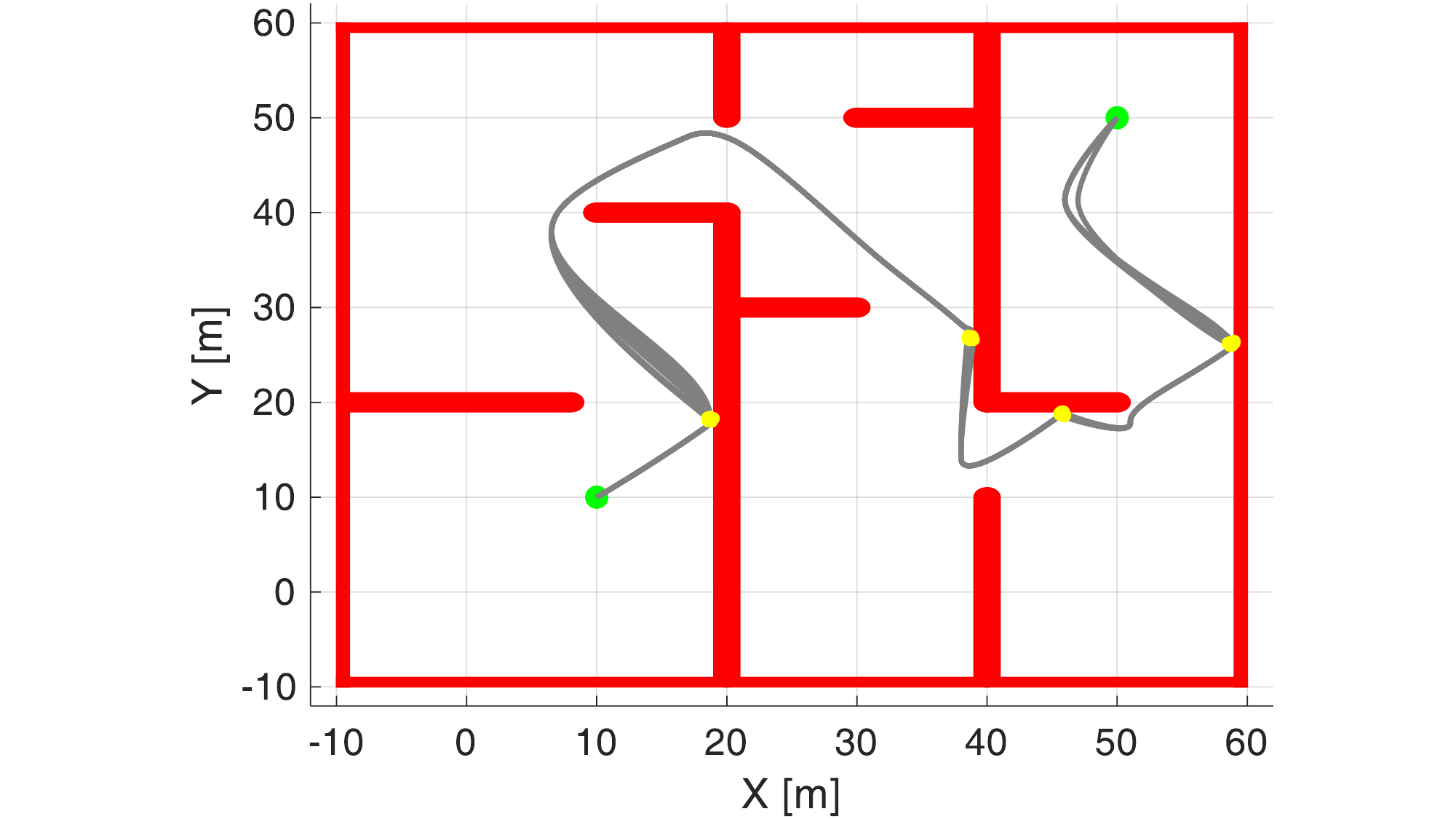}
  \vspace{-3pt}
  \caption{$T_{rep} = 10.0$\;s.}
  \label{fig:strategy simulation m2 10}
 \end{subfigure}%
 \vspace{-6pt}
 \caption{Simulated trajectories generated from our method in environments with isolated non-convex obstacles.}
 \label{fig:strategy simulation non-convex}
 \vspace{-21pt}
\end{figure}

We then test our proposed unified collision-inclusive motion planning and control strategy in environments with non-convex obstacles (Table~\ref{table:strategy simulation non-convex} and Fig.~\ref{fig:strategy simulation non-convex}).
With reference to the environment shown in Fig.~\ref{fig:strategy simulation m1 5}, and using a replanning time interval $T_{rep} = 5.0$\;s, our strategy can reach the goal with lower computational time, trajectory time and path length but higher control energy since the robot trajectories oscillate at the corner of the obstacle after collision-recovery and detouring. Similar patterns are observed for $T_{rep} = 10.0$\;s. 
In the case shown in Fig.~\ref{fig:strategy simulation m2 5}, and with $T_{rep} = 5.0$\;s, our strategy is better than collision avoidance in terms of computational time, trajectory time, path length and control energy. When $T_{rep} = 10.0$\;s, the control energy of our method increases because of oscillations around corners.

\begin{table}[!h]
\vspace{-3pt}
    \caption{Results by testing with non-convex obstacles.}
    \vspace{-6pt}
    \label{table:strategy simulation non-convex}
    \begin{center}
    \resizebox{0.45\textwidth}{!}{
    \begin{tabular}{c c c c c c c c}
    \toprule
    non-convex env. & & Comp. & Traj. & Ctrl. & Path\\
    Fig~\ref{fig:strategy simulation m1 5}
    & & Time[s] & Time[s] & Cost[${\rm m^{2}/s^{3}}$] & Len.[m]\\
    \toprule
    \multirow{2}*{\shortstack{Our method \\ $T_{rep} = 5.0$}} & mean & $35.90$ & $81.34$ & $34.24$ & $144.29$ \\
     & std & $0.953$ & $0.398$ & $0.434$ & $0.231$ \\
    \midrule
    \multirow{2}*{\shortstack{Collision-avoidance \\ $T_{rep} = 5.0$}} & & \multirow{2}*{$52.55$} & \multirow{2}*{$85.0$} & \multirow{2}*{$25.814$} & \multirow{2}*{$160.67$} \\
    \\
    \midrule
    \multirow{2}*{\shortstack{Our method \\ $T_{rep} = 10.0$}} & mean & $16.71$ & $80.43$ & $21.15$ & $146.51$ \\
     & std & $0.243$ & $0.857$ & $0.312$ & $0.121$ \\
    \midrule
    \multirow{2}*{\shortstack{Collision-avoidance \\ $T_{rep} =10.0$}} & & \multirow{2}*{$18.62$} & \multirow{2}*{$90.0$} & \multirow{2}*{$15.11$} & \multirow{2}*{$167.25$} \\
    \\
    \bottomrule
   \toprule
    non-convex env. & & & & & \\
    Fig~\ref{fig:strategy simulation m2 5}
    & & & & & \\
    \toprule
    \multirow{2}*{\shortstack{Our method \\ $T_{rep} = 5.0$}} & mean & $9.09$ & $80.48$ & $19.72$ & $146.02$ \\
     & std & $0.210$ & $0.623$ & $2.205$ & $2.543$ \\
    \midrule
    \multirow{2}*{\shortstack{Collision-avoidance \\ $T_{rep} = 5.0$}} & & \multirow{2}*{$32.62$} & \multirow{2}*{$93.4$} & \multirow{2}*{$27.38$} & \multirow{2}*{$173.60$} \\
    \\
    \midrule
    \multirow{2}*{\shortstack{Our method \\ $T_{rep} = 10.0$}} & mean & $6.98$ & $85.33$ & $32.54$ & $153.17$ \\
     & std & $0.199$ & $0.857$ & $0.556$ & $0.395$ \\
    \midrule
    \multirow{2}*{\shortstack{Collision-avoidance \\ $T_{rep} =10.0$}} & & \multirow{2}*{$14.45$} & \multirow{2}*{$94.4$} & \multirow{2}*{$19.50$} & \multirow{2}*{$166.94$} \\
    \\
    \bottomrule
    \end{tabular}
    }
    \end{center}
    \vspace{-15pt}
\end{table}

\subsection{Experimental Validation of our Framework}\label{subsec:overall_testing}
\begin{figure}[!t]
\vspace{-4pt}
      \centering
      \begin{subfigure}{0.235\textwidth}
        \includegraphics[trim={5cm 3.0cm 5cm 3.0cm}, clip, width=0.925\textwidth]{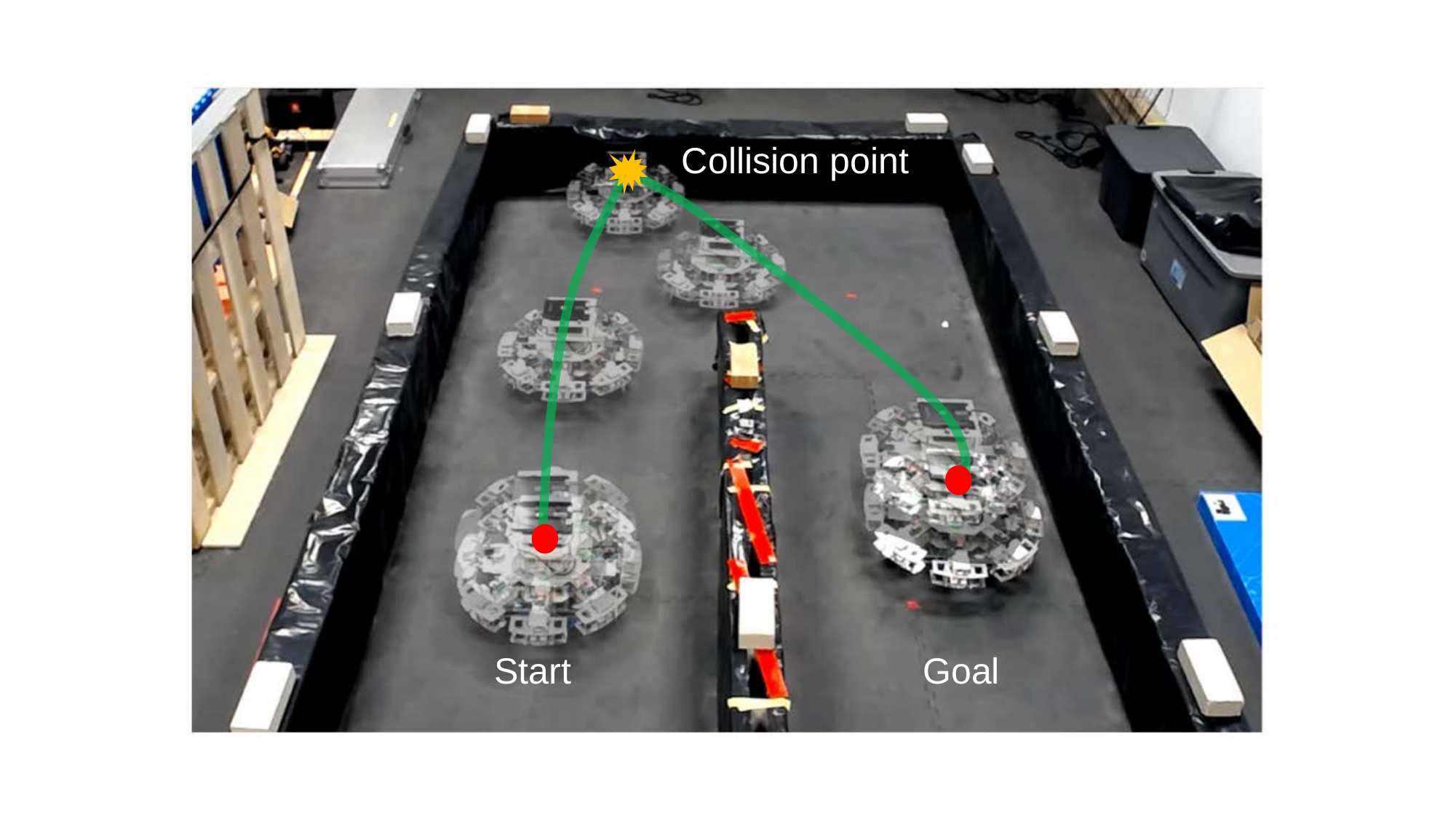}
      \end{subfigure}
      \begin{subfigure}{0.235\textwidth}
       \centering
        \includegraphics[trim={5cm 3.0cm 5cm 3.0cm}, clip, width=0.925\textwidth]{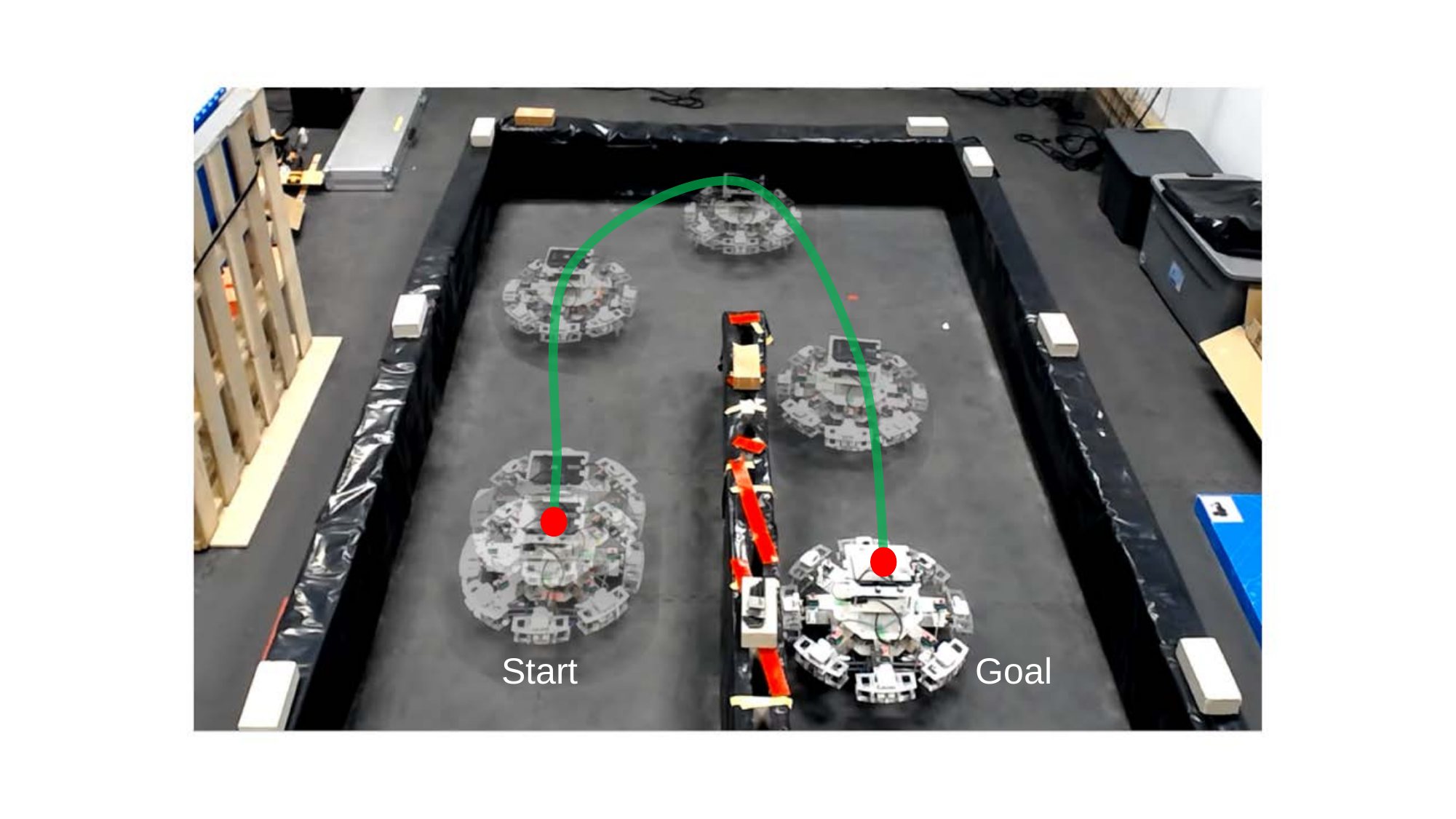}
      \end{subfigure}
      \caption{Composite images of a sample experiment with our unified collision-inclusive motion planning and control framework (left) and of a collision avoidance sample experiment (right). Snapshots shown every $2.5$\;s. (See supplementary video file for more details.)} 
      \label{fig:ExperimentPlanner}
	\vspace{-9pt}
\end{figure}

\begin{figure}[h!]
\vspace{-2pt}
  \begin{subfigure}{.235\textwidth}
  \centering
  \includegraphics[trim={4cm 0.2cm 4cm 0.2cm}, clip,width=0.9\linewidth]{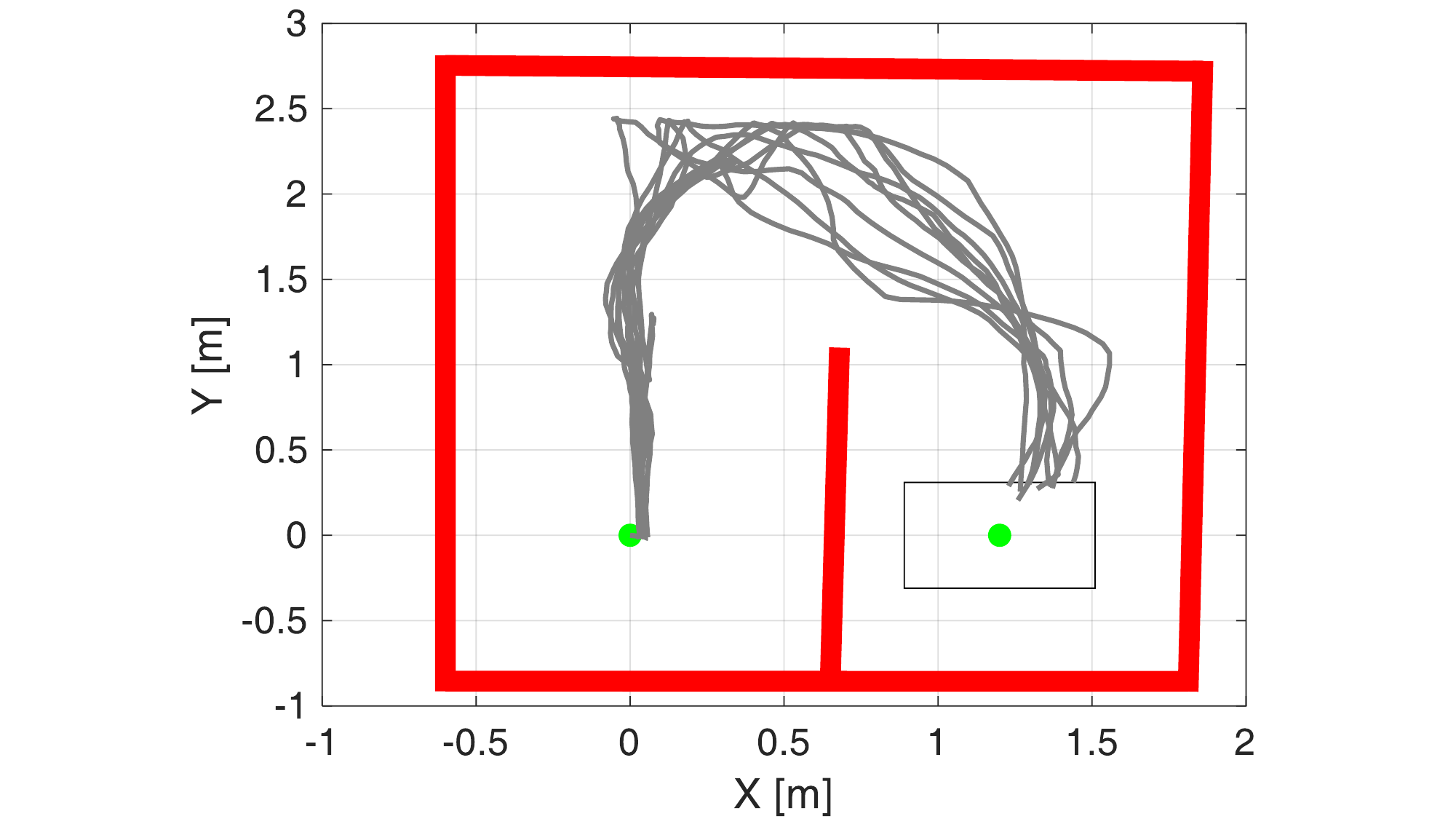}
  \vspace{-3pt}
 \end{subfigure}%
  \begin{subfigure}{.235\textwidth}
  \centering
  \includegraphics[trim={4cm 0.2cm 4cm 0.2cm}, clip, width=0.9\linewidth]{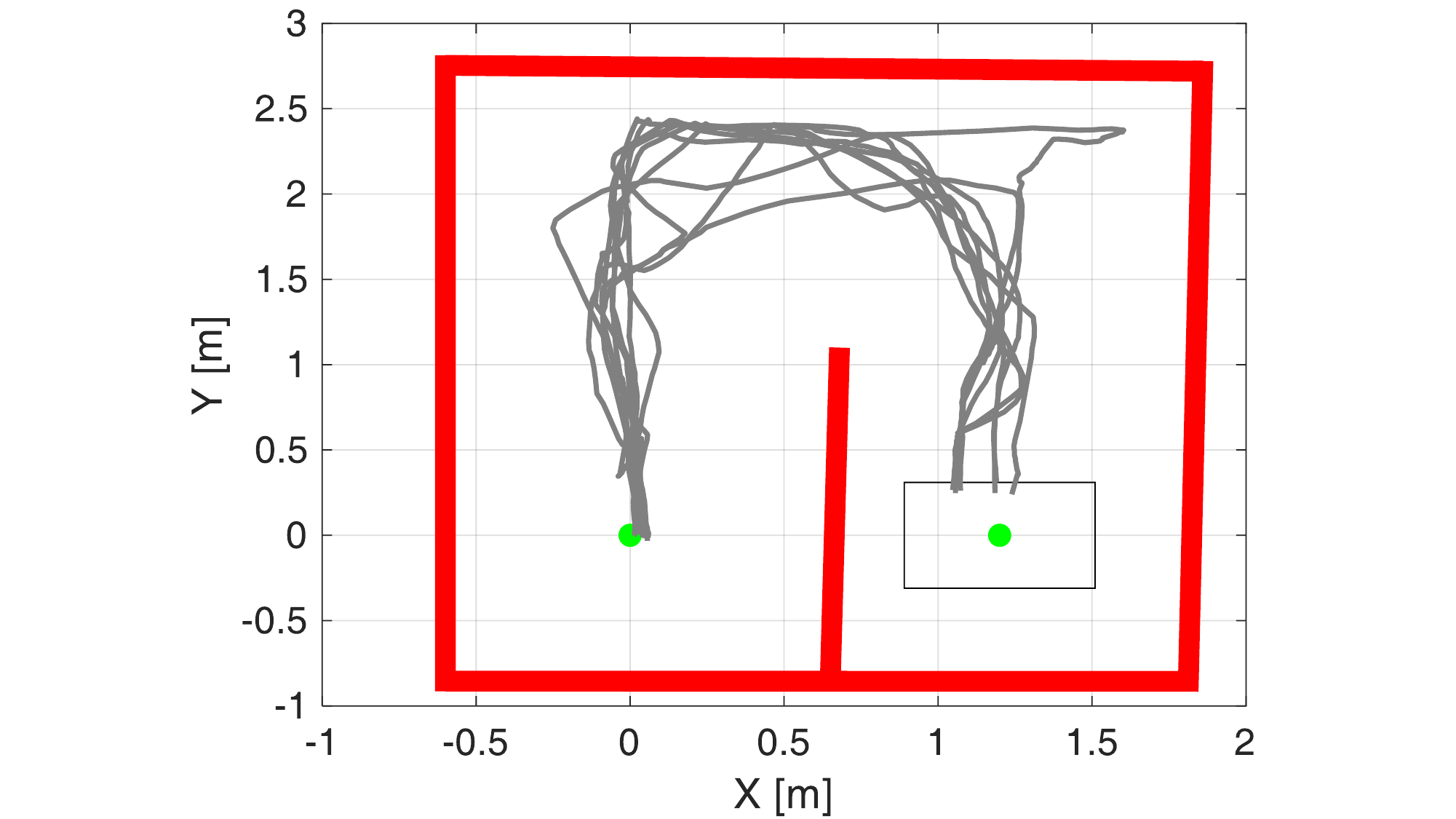}
  \vspace{-3pt}
 \end{subfigure}%
 \vspace{-1pt}
 \caption{Experimental trajectories generated from our proposed collision-inclusive method (left) the collision avoidance method in Sec.~\ref{subsec:simulation framework} (right). (In all cases we conduct 10 trials). }
 \label{fig:trajectory hardware}
 \vspace{-9pt}
\end{figure}

Finally, we validate our proposed framework experimentally, and also test is against the collision avoidance strategy in Sec.~\ref{subsec:simulation framework}, in a single corridor environment similar to~\cite{mote2020collision} (Fig.~\ref{fig:ExperimentPlanner}). Each method is repeated for $10$ times using the same parameter settings. Output trajectories are depicted in Fig.~\ref{fig:trajectory hardware}, while detailed numerical results are given in Table~\ref{table:overall strategy}. 
By implementing our proposed collision-inclusive planning method, the robot can reach the goal area with higher success rates since unmodeled dynamics in physical testing make the robot collide with the obstacle even if the reference trajectory generated from collision avoidance is designed to be collision-free. Further, by utilizing collisions, the robot can reach the goal faster while requiring less control energy by trading off the average path length. 

\begin{table}[!h]
\vspace{0pt}
    \caption{Comparison of collision-inclusive and collision-avoidance frameworks in the physical robot.}
    \vspace{-6pt}
    \label{table:overall strategy}
    \begin{center}
    \resizebox{0.45\textwidth}{!}{
    \begin{tabular}{c c c c c c c}
    \toprule
     & & Path & Traj. & Ctrl. & Succ.\\
     & & Len.[m] & Time[s] & Cost[${\rm m^{2}/s^{3}}$] & Rate[$\%$]\\
    \toprule
    \multirow{2}*{\shortstack{Collision-avoidance}} & mean & $4.99$ & $12.74$ & $38.32$ &  \multirow{2}*{$90.0$}\\
    & std & $1.80$ & $2.61$ & $16.75$ &  \\
    \midrule
    \multirow{2}*{\shortstack{Collision-inclusive}} & mean & $5.36$ & $11.98$ & $32.45$ &  \multirow{2}*{$100.0$}\\
    & std & $0.33$ & $1.19$ & $10.05$ &  \\
    \bottomrule
    \end{tabular}
    }
    \end{center}
    \vspace{-21pt}
\end{table}

\section{Discussion and Conclusions}\label{sec:conclusion}

\subsection{Summary of Contributions and Main Findings} In this article, we proposed a unified collision-inclusive motion planning and control framework applied for navigation in unknown environment. A global search-based method is devised to generate a path which contains explicit information about collisions. The effect of the collisions is explored in the global planner. The local planner is enhanced by a lower-level deformation recovery control and trajectory replanning strategy, which enables the robot to detect and recover from collisions and move toward the goal. The deformation controller is designed based on robot dynamics, which herein is a holonomic omni-directional wheeled robot.  

The planning system was evaluated extensively through several benchmark comparisons in simulation as well as via physical experimental testing. The conducted ablation study demonstrated the utility of certain key design choices made in this work (e.g., not pruning primitives altogether), and evaluated the effect of key parameters (e.g., how much collisions are to be penalized via parameter $\rho_c$). The proposed collision-inclusive planning method is implemented in simulation first and then integrated with state estimation, mapping and control into our custom-made robot platform to check the feasibility of the method in physical world experiments. Results show that the proposed method is robust and can generate fast and safe trajectories compared to collision-avoidance methods. Overall, this work pushes forward the state-of-the-art in collision-inclusive motion planning and control, and provides a competitive alternative to traditional collision avoidance methods for a class of impact-resilient mobile robots operating in partially-observable environments populated with isolated (non-)convex obstacles.

\subsection{Discussion of Key Selections in our Framework}
\paragraph*{\textbf{Application to Other Robots in 2D and 3D}} We considered the family of omni-directional wheeled robots (Fig.~\ref{fig:robots}). Yet, we anticipate that our proposed framework can apply to other impact-resilient robots in 2D (e.g., wheeled~\cite{StagerICRA19} or aerial~\cite{briod2013contact,zha2020collision,mulgaonkar2020tiercel, de2020collision,liu2021toward} robots) provided that they can adjust their position and redirect post impact by using the collision to save energy. The omni-directional wheeled robot employed here is one example along those types of robots. The higher-level part of the framework can readily apply in 3D for such systems; same holds for the overall methodology as in whole. However, the proposed lower-level collision recovery would need to be adjusted to consider the 3D dynamics for post-impact stabilization~\cite{liu2021toward}.

\paragraph*{\textbf{Use of Motion Primitives}} Besides the use of motion primitives (as herein) other methods are possible. For instance, direct control of the kinematic model~\eqref{state space model} of the specific robot considered herein, or use of fixed motion patterns (e.g., as in~\cite{nilles2021visibility, alam2017minimalist, lewis2013planning}) can be viable alternatives. However, use of motion primitives at the higher-level provides a unified way to make the proposed framework applicable to all the aforementioned types of robots and extendable from 2D to 3D, and hence it was preferred to over simpler approaches that would have worked specifically for the omni-directional robot we tested with herein but would be hard to scale to other types of robots. Furthermore, use of primitives allows for more flexibility which is critical to help determine where the robot should collide with the environment to help it redirect toward the globally-planned goal; this is achieved by directly using information on the velocity as per~\eqref{collsion cost}.

\paragraph*{\textbf{Choosing a Search- or Sampling-based Global Planner}} We showed that it is possible to derive collision-inclusive planning frameworks with the global planner being either search-based or sampling-based. Each has its own strengths and weaknesses, and as a matter of fact, our results are consistent with observations made in collision-avoidance methods. Consistent with collision avoidance, a user can choose which approach to select (search-based over sampling-based global planner) according to their application needs; our proposed framework can accommodate both. We highlight here that the sampling-based global planner can be further optimized by biasing search toward free space to increase computational efficiency (e.g.,~\cite{sandstrom2020topology}). Integration of the sampling-based planner into our overall real-time framework would require further adaptations of the collision-inclusive RRT* planner to make it online (faster nearest neighbor search, minimal cost path generation, and optimized rewiring methods). Similar to collision-avoidance online RRT* methods (e.g.,~\cite{hernandez2015online, karaman2011anytime}), a collision-inclusive anytime planning algorithm is required to extend the RRT* method for planning collision-inclusive trajectories online.

\subsection{Directions for Future Work}
The framework developed herein lays the basis toward a general method for collision-inclusive motion planning and control, and creates multiple opportunities for future research along these lines. These include extension to other robots and to systems with higher-order dynamics, evaluation of direct controllers against motion primitives (as well as different parameterizations of the latter), and integration of sampling-based planners into the overall framework.

Further, at its current form, our method does not consider the perception model of the robot in online planning; extension of the proposed algorithm to consider the perception problem based on the collision-inclusive method is another interesting direction of future research. Lastly, we have shown that it is possible to handle navigation in environments populated with isolated non-convex environments; however, study of navigation in more cluttered environments (e.g., maze-like) is a direction of research enabled by this work.

\section*{Acknowledgement}
The authors wish to thank Hanzhe Teng for his help implementing the mapping package used in this work.

\bibliographystyle{IEEEtran}
\bibliography{Citation}

\end{document}